\documentclass{article}
\usepackage{amsmath}
\usepackage{amsfonts}
\usepackage{amssymb}
\usepackage{amsthm}
\usepackage{xcolor}
\usepackage{microtype,amssymb,amsmath}
\usepackage{graphicx}
\usepackage{subcaption}
\usepackage{booktabs}
\usepackage{hyperref}

\usepackage[accepted]{icml2020}
\usepackage{cleveref}
\Crefformat{section}{Section~#2#1#3}
\Crefformat{proposition}{Proposition~#2#1#3}
\Crefformat{equation}{Eq.\,#2#1#3}
\Crefformat{definitions}{Definition~#2#1#3}
\icmltitlerunning{MCTS as regularized policy optimization}
\usepackage[]{todonotes}

\newcommand{\CommaBin}{\mathbin{\raisebox{0.5ex}{,}}}
\let\originalleft\left
\let\originalright\right
\renewcommand{\left}{\mathopen{}\mathclose\bgroup\originalleft}
\renewcommand{\right}{\aftergroup\egroup\originalright}

\renewcommand{\ast}{\star}

\newcommand{\E}{\mathbb{E}}

\newcommand{\cO}{\mathcal{O}}

\newcommand{\KL}{\mathrm{KL}}

\newcommand{\EEs}[2]{\E_{#1}\left[#2\right]}

\newcommand{\prth}[1]{\left( #1 \right)}
\newcommand{\brkt}[1]{\left[ #1 \right]}
\newcommand{\brc}[1]{\left\{ #1 \right\}}

\DeclareMathOperator*{\argmax}{arg\,max}
\DeclareMathOperator*{\argmin}{arg\,min}
\newlength{\minipagewidth}
\newcommand{\Real}{\mathbb{R}}

\newcommand{\eqdef}{\triangleq}

\newtheorem{predefinition}{Definition}
\newenvironment{definition}[1]
{
\begin{predefinition}[\emph{#1}]
}{
\end{predefinition}
}

\newtheorem{theorem}{Theorem}

\newtheorem{preobservation}{Observation}
\newenvironment{observation}[1]
{
\begin{preobservation}
}{
\end{preobservation}
}

\newtheorem{preproposition}{Proposition}
\newenvironment{proposition}[1]
{
\begin{preproposition}
}{
\end{preproposition}
}

\usepackage{mathtools}
\usepackage{bbm}

\newtheorem{lemma}{Lemma}

\newtheorem{remark}{Remark}

\renewcommand{\phi}{\varphi}
\renewcommand{\epsilon}{\varepsilon}

\newcommand{\A}{\mathcal{A}}

\newcommand{\X}{\mathcal{X}}

\newcommand{\pibar}{\bar{\pi}}
\newcommand{\pihat}{\hat{\pi}}
\newcommand{\algoact}{\textsc{Act}}
\newcommand{\algosearch}{\textsc{Search}}
\newcommand{\algolearn}{\textsc{Learn}}
\newcommand{\algoall}{\textsc{All}}

\Crefname{proposition}{Proposition}{Propositions}
\Crefname{prop}{Proposition}{Propositions}
\crefname{lemma}{Lemma}{Lemmas}

\begin{document}

\twocolumn[
\icmltitle{Monte-Carlo tree search as regularized policy optimization}



\icmlsetsymbol{equal}{*}

\begin{icmlauthorlist}
\icmlauthor{Jean-Bastien Grill}{equal,dmp}
\icmlauthor{Florent Altch\'e}{equal,dmp}
\icmlauthor{Yunhao Tang}{equal,dmp,col}
\icmlauthor{Thomas Hubert}{dm}
\icmlauthor{Michal Valko}{dmp}
\icmlauthor{Ioannis Antonoglou}{dm}
\icmlauthor{R\'emi Munos}{dmp}
\end{icmlauthorlist}

\icmlaffiliation{dm}{DeepMind, London, UK}
\icmlaffiliation{dmp}{DeepMind, Paris, FR}
\icmlaffiliation{col}{Columbia University, New York, USA}
\icmlcorrespondingauthor{Jean-Bastien Grill}{jbgrill@google.com}
\icmlkeywords{MCTS, policy optimization, planning}

\vskip 0.3in
]
\printAffiliationsAndNotice{\icmlEqualContribution} 
\begin{abstract}
The combination of Monte-Carlo tree search (MCTS) with deep reinforcement learning has led to significant advances in artificial intelligence. However, AlphaZero, the current state-of-the-art MCTS algorithm, still relies on handcrafted  heuristics that are only partially understood. In this paper, we show that AlphaZero's search heuristics, along with other common ones such as UCT, are an approximation to the solution of a specific regularized policy optimization problem. With this insight, we propose a variant of AlphaZero which uses the \textit{exact} solution to this policy optimization problem, and show experimentally that it reliably outperforms the original algorithm in multiple domains.
\end{abstract}
\section{Introduction}

Policy gradient is at the core of many state-of-the-art deep reinforcement learning (RL) algorithms. Among many successive improvements to the original algorithm~\citep{sutton2000policy}, \emph{regularized policy optimization} encompasses a large family of such techniques. Among them trust region policy optimization is a prominent example~\citep{schulman2015trust,schulman2017proximal,abdolmaleki2018maximum,song2019v}. These algorithmic enhancements have led to significant performance gains in various benchmark domains \citep{song2019v}.


As another successful RL framework, the AlphaZero family of algorithms~\citep{silver2016mastering,silver2017alphagozero,silver2017mastering,schrittwieser2019mastering} have obtained groundbreaking results on challenging domains by combining classical deep learning \citep{he2016deep} and RL \citep{williams1992simple} techniques with Monte-Carlo tree search~\citep{kocsis2006bandit}. To search efficiently, the MCTS action selection criteria takes inspiration from bandits~\citep{auer2002using}. Interestingly, AlphaZero employs an \textit{alternative} handcrafted heuristic to achieve super-human performance on board games~\citep{silver2016mastering}. 
Recent MCTS-based MuZero \citep{schrittwieser2019mastering} has also led to  state-of-the-art results in the Atari benchmarks~\citep{bellemare2013arcade}.


Our main contribution is connecting MCTS algorithms, in particular the highly-successful AlphaZero, with MPO, a state-of-the-art model-free policy-optimization algorithm~\citep{abdolmaleki2018maximum}. Specifically, we show that the \textit{empirical visit distribution} of actions in AlphaZero's search procedure approximates the solution of a regularized policy-optimization objective. With this insight, our second contribution a \textit{modified version of AlphaZero} that comes significant performance gains over the original algorithm, especially in cases where AlphaZero has been observed to fail, \emph{e.g.,} when per-search simulation budgets are low~\citep{hamrick2019combining}.

In \Cref{sec:background}, we briefly present MCTS with a focus on AlphaZero and provide a short summary of the model-free policy-optimization. In \Cref{sec:mcts_as_po}, we show that AlphaZero (and many other MCTS algorithms) computes approximate solutions to a family of regularized policy optimization problems. With this insight, \Cref{sec:algo} introduces a modified version of AlphaZero which leverages the benefits of the policy optimization formalism to improve upon the original algorithm.
Finally, \Cref{sec:experiments} shows that this modified algorithm outperforms AlphaZero on Atari games and continuous control tasks.


\section{Background}
\label{sec:background}
Consider a standard RL setting tied to a Markov decision process (MDP) with state space~$\X$ and action space $\A$. At a discrete round $t \geq 0$, the agent in state $x_t \in \X$ takes action $a_t \in \A$ given  a policy $a_t \sim \pi(\cdot|s_t)$, receives reward $r_t$, and transitions to a next state $x_{t+1} \sim p(\cdot|x_t,a_t)$. The RL problem consists in finding a policy which maximizes the discounted cumulative return $\mathbb{E}_\pi[\sum_{t\geq 0}\gamma^t r_t]$ for a discount factor $\gamma\in (0,1)$. To scale the method to large environments, we assume that the policy $\pi_\theta(a|x)$ is parameterized by a neural network $\theta$. 

\subsection{AlphaZero}
\label{sec:alphazero_muzero}
We focus on the AlphaZero family, comprised of AlphaGo~\cite{silver2016mastering}, AlphaGo Zero~\citep{silver2017alphagozero}, AlphaZero~\citep{silver2017mastering}, and MuZero~\citep{schrittwieser2019mastering}, which are among the most successful algorithms in combining model-free and model-based RL. Although they make different assumptions, all of these methods share the same underlying search algorithm, which we refer to as \textit{AlphaZero} for simplicity.

From a state~$x$, AlphaZero uses MCTS  (\citealp{browne2012survey}) to compute an improved policy $\pihat(\cdot|x)$ at the root of the search tree from the prior distribution predicted by a policy network $\pi_\theta(\cdot|x)$\footnote{We note here that terminologies such as \emph{prior} follow \citet{silver2017mastering} and do not relate to concepts in Bayesian statistics.}; see \Cref{eq:pihat} for the definition. This improved policy is then distilled back into~$\pi_\theta$ by updating~$\theta$ as 
$
    \theta \leftarrow \theta - \eta \nabla_\theta \EEs{x}{D( \pihat(\cdot|x),\pi_\theta(\cdot|x))}
$
for a certain divergence~$D$. In turn, the distilled parameterized policy~$\pi_\theta$  informs the next local search by predicting priors, further improving the local policy over successive iterations. Therefore, such an algorithmic procedure is a special case of generalized policy improvement~\citep{sutton_RLIntro_1998}.

One of the main differences between AlphaZero and previous MCTS algorithms such as UCT~\citep{kocsis2006bandit} is the introduction of a learned prior $\pi_\theta$ and value function  $v_\theta$. Additionally, AlphaZero's search procedure applies the following action selection heuristic,
\begin{align}
    \argmax_a \brkt{Q(x,a) + c \cdot \pi_\theta(a|x) \cdot \frac{\sqrt{\sum_b n(x,b)}}{1 + n(x,a)}}\CommaBin
    \label{eq:azselection}
\end{align}
where $c$ is a numerical constant,\footnote{\citet{schrittwieser2019mastering} uses a $c$ that has a slow-varying dependency on $\sum_b n(x,b)$, which we omit here for simplicity, as it was the case of~\citet{silver2017mastering}.} $n(x, a)$ is the number of times that action $a$ has been selected from state $x$ during search, and $Q(x, a)$ is an estimate of the Q-function for state-action pair $(x, a)$ computed from search statistics and using $v_\theta$ for bootstrapping. 

Intuitively, this selection criteria balances exploration and exploitation, by selecting the most promising actions (high Q-value $Q(x,a)$ and prior policy $\pi_\theta(a|x)$) or actions that have rarely been explored (small visit count $n(x,a)$). 
We denote by $N_\text{sim}$ the simulation budget, \textit{i.e.,} the search is run with $N_\text{sim}$ simulations.
A more detailed presentation of AlphaZero is in \Cref{app:alphazero-search}; for a full description of the algorithm, refer to~\citet{silver2017mastering}.

\newcommand{\transpose}{^\mathsf{\scriptscriptstyle T}}
\subsection{Policy optimization}
Policy optimization aims at finding a globally optimal policy~$\pi_\theta$, generally using iterative updates. Each iteration updates the current policy~$\pi_\theta$ by solving a local maximization problem of the form
\begin{align}
\pi_{\theta'} \triangleq \argmax_{\mathbf{y} \in \mathcal{S}}  \ {\mathcal{Q}\transpose_{\pi_\theta}} \mathbf{y} - \mathcal{R}\prth{\mathbf{y}, \pi_\theta},
\label{eq:po_opt}
\end{align}
where $\mathcal{Q}_{\pi_\theta}$ is an estimate of the Q-function,  $\mathcal{S}$ is the $|\mathcal{A}|$-dimensional simplex and $\mathcal{R} : \mathcal S^2 \rightarrow \Real$ a convex regularization term \cite{neu2017unified,grill2019planning,geist2019theory}. Intuitively, \Cref{eq:po_opt} updates $\pi_\theta$ to maximize the value $\mathcal {\mathcal{Q}\transpose_{\pi_\theta}}\mathbf{y}$ while constraining the update with a regularization term $\mathcal{R}(\mathbf{y},\pi_\theta)$. 

Without regularizations, \emph{i.e.,} $\mathcal{R}=0$, \Cref{eq:po_opt} reduces to policy iteration \citep{sutton_RLIntro_1998}. When $\pi_\theta$ is updated using a single gradient ascent step towards the solution of \Cref{eq:po_opt}, instead of using the solution directly, the above formulation reduces to (regularized) policy gradient \citep{sutton2000policy,levine2018reinforcement}.

Interestingly, the regularization term has been found to stabilize, and possibly to speed up the convergence of $\pi_\theta$. For instance, trust region policy search algorithms (TRPO, \citealp{schulman2015trust}; MPO \citealp{abdolmaleki2018maximum}; V-MPO, \citealp{song2019v}), set~$\mathcal{R}$ to be the KL-divergence between consecutive policies $\KL[\mathbf{y}, \pi_\theta]$; maximum entropy RL \cite{ziebart2010modeling, fox2015taming, o2016combining, haarnoja2017reinforcement} sets $\mathcal{R}$ to be the negative entropy of $\mathbf{y}$ to avoid collapsing to a deterministic policy. 


\section{MCTS as regularized policy optimization}
\label{sec:mcts_as_po}
In \Cref{sec:background}, we presented AlphaZero that relies on model-\emph{based} planning. We also presented policy optimization, a framework that has achieved good performance in model-\emph{free}\/ RL. In this section, we establish our main claim namely that AlphaZero's action selection criteria can be interpreted as approximating the solution to a regularized policy-optimization objective.

\subsection{Notation}
First, let us define the \textit{empirical visit distribution} $\pihat$ as
\begin{align}
    \pihat(a|x) \eqdef \frac{1 + n(x,a)}{|\mathcal{A}| + \sum_{b} n(x,b)}\cdot \label{eq:pihat}
\end{align} 
Note that in \Cref{eq:pihat}, we consider an extra visit per action compared to the acting policy and distillation target in the original definition~\citep{silver2016mastering}. This extra visit is introduced for convenience in the upcoming  analysis (to avoid divisions by zero) and does not change the generality of our results.

We also define the \textit{multiplier} $\lambda_N$ as
\begin{align}
    \lambda_N(x) \eqdef  c\cdot\frac{\sqrt{\sum_b n_b}}{|\mathcal{A}| + \sum_b n_b}\CommaBin \label{eq:lambda}
\end{align}
where the shorthand notation $n_a$ is used for $n(x, a)$, and $N(x) \eqdef \sum_b n_b$ denotes the number of visits to $x$ during search. With this notation, the action selection formula of \Cref{eq:azselection} can be written as selecting the action $a^\ast$ such that
\begin{align}
    a^\star(x) \triangleq \argmax_a \left[ Q(x,a) + \lambda_N \cdot \frac{\pi_\theta(a|x)}{\pihat(a|x)} \right]\cdot
    \label{eq:az_act_select_reformalized}
\end{align}
Note that in \Cref{eq:az_act_select_reformalized} and in the rest of the paper (unless otherwise specified), we use $Q$ to denote the \emph{search} Q-values, \emph{i.e.,} those estimated by the search algorithm as presented in \Cref{sec:alphazero_muzero}. For more compact notation, we use bold fonts to denote vector quantities, with the convention that $\frac{\mathbf u}{\mathbf v}[a] = \frac{\mathbf u[a]}{\mathbf v[a]}$ for two vectors $\mathbf u$ and $\mathbf v$ with the same dimension. Additionally, we omit the dependency of quantities on state $x$ when the context is clear. In particular, we use $\mathbf{q} \in \mathbb{R}^{|\mathcal{A}|}$ to denote the vector of search Q-function $Q(x,a)$ such that $\mathbf{q}_a = Q(x,a)$. With this notation, we can rewrite the action selection formula of \Cref{eq:az_act_select_reformalized} simply as\footnote{When the context is clear, we simplify for any $\mathbf{x} \in\mathbb{R}^{|\mathcal{A}|}$, that $\arg\max \ [\mathbf{x}] \triangleq \arg\max_a\  \{\mathbf{x}[a],a\in\mathcal{A} \}$.}
\begin{equation}a^\ast \triangleq \argmax \brkt{\mathbf q + \lambda_N \frac{\boldsymbol{\pi_\theta}}{\boldsymbol{\pihat}}}\cdot\label{eq:azselection_vector}\end{equation}

\subsection{A related regularized policy optimization problem}We now define $\pibar$ as the solution to a regularized policy optimization problem; we will see in the next subsection that the visit distribution $\pihat$ is a good approximation of~$\pibar$. 
\begin{definition}{$\boldsymbol{\pibar}$} Let $\boldsymbol{\pibar}$  be the solution to the following objective
\begin{align}
\boldsymbol{\pibar} \eqdef &\ \argmax_{\mathbf{y} \in \mathcal{S}}  \brkt{\mathbf{q}\transpose \mathbf{y} - \lambda_N \KL\brkt{\boldsymbol{\pi_\theta}, \mathbf{y}}},  
\end{align}
where $\mathcal{S}$ is the $|\mathcal{A}|-$dimensional simplex and $\KL$ is the KL-divergence.\footnote{We apply the definition $\KL[\mathbf{x},\mathbf{y}] \triangleq \sum_a \mathbf{x}[a] \log \frac{\mathbf{x}[a]}{\mathbf{y}[a]}\cdot$}
\label{def:pibar}
\end{definition}
We can see from \Cref{eq:po_opt} 
and \Cref{def:pibar} that $\boldsymbol{\pibar}$ is 
the solution to a policy optimization problem where $\mathcal{Q}$ is set to the search Q-values, and the regularization term $\mathcal{R}$ is a \emph{reversed} KL-divergence weighted by factor $\lambda_N$. 

In addition, note that $\pibar$ is as a smooth version of the $\argmax$ associated to the search Q-values $\mathbf{q}$. In fact, $\pibar$ can be computed as (\Cref{app:compute_pibar} gives a detailed derivation of $\pibar$)
\begin{align}
    \boldsymbol{\pibar} = \lambda_N  \frac{\boldsymbol{\pi_{\theta}}}{\alpha - \mathbf{q}}\CommaBin
    \label{eq:pibar}
\end{align}
where $\alpha\in\Real$ is such that $\pibar$ is a proper probability vector. 
This is slightly different from the softmax distribution obtained with $\KL[\mathbf y, \pi_\theta]$, which is written as
\begin{align}
    \nonumber
    \argmax_{\mathbf{y} \in \mathcal{S}}  \brkt{\mathbf{q}\transpose \mathbf{y} - \lambda_N \KL\brkt{\mathbf{y}, \boldsymbol{\pi_\theta}}} \propto \pi_\theta \exp\prth{\frac{\mathbf{q}}{\lambda_N}}\cdot
\end{align}
\paragraph{Remark} The factor $\lambda_N$ is a decreasing function of $N$. Asymptotically, $\lambda_N = \tilde O(1/\sqrt{N})$. Therefore, the influence of the regularization term decreases as the number of simulation increases, which makes $\pibar$ rely increasingly more on search Q-values $\mathbf{q}$ and less on the policy prior $\pi_\theta$. As we explain next, $\lambda_N$ follows the design choice of AlphaZero, and may be justified by 
a similar choice done in
bandits \citep{bubeck2012regret}.

\subsection{AlphaZero as policy optimization}
We now analyze the action selection formula of AlphaZero (\Cref{eq:azselection}). Interestingly, we show that this formula, which was \textit{handcrafted}\footnote{Nonetheless, this heuristic could be interpreted as loosely inspired by bandits \citep{rosin2011multi}, but was adapted to accommodate a prior term $\pi_\theta$.} independently of the policy optimization research, turns out to result in a distribution $\pihat$ that closely relates to the policy optimization solution $\pibar$.

The main formal claim of this section that AlphaZero's search policy $\pihat$ \textit{tracks} the exact solution $\pibar$ of the regularized policy optimization problem of \Cref{def:pibar}. We show that \Cref{prop:action_selection_diff} and \Cref{prop:pihat_leq_pibar} support this claim from two complementary perspectives.

First, with Proposition~\ref{prop:action_selection_diff}, we show that $\pihat$ approximately follows the gradient of the concave objective for which $\pibar$ is the optimum. 
\begin{proposition}{}
\label{prop:action_selection_diff}
For any action $a\in\mathcal{A}$, visit count $n\in\mathbb{R}^\mathcal{A}$, policy prior $\pi_\theta>0$ and Q-values $\mathbf{q}$, 
\begin{align}\label{eq:jb}
    a^\ast = \argmax_a\brkt{\frac{\partial}{\partial n_a}\prth{\mathbf{q}\transpose \boldsymbol{\pihat} - \lambda_N \KL\brkt{\mathbf{\boldsymbol{\pi_\theta}}, \boldsymbol{\pihat}}}},
\end{align}
with $a^\ast$ being the action realizing \Cref{eq:azselection} as defined in \Cref{eq:az_act_select_reformalized} and $\pihat=\prth{1+\mathbf{n}}/\prth{|\mathcal{A}| + \sum_b n_b}$ as defined in \Cref{eq:pihat}, is a function of the count vector extended to real values. 
\end{proposition}


The only thing that the search algorithm eventually influences through the tree search
is the visit count distribution.
If we could do an infinitesimally small update, then the greedy update 
maximizing \Cref{eq:pibar} would be in the direction of the partial derivative of \Cref{eq:jb}.
However, as we are \textit{restricted by a discrete update}, then increasing the visit count 
as in Proposition~\ref{prop:action_selection_diff} makes $\pihat$ track $\pibar$.
Below, we further characterize the selected action $a^\ast$ and 
 assume $\pi_\theta>0.$ 
\begin{proposition}{}
The action $a^\ast$ realizing \Cref{eq:azselection} is such that 
\begin{align}
\pihat(a^\ast|x) \leq \pibar(a^\ast|x).
\label{eq:smaller}
\end{align}
\label{prop:pihat_leq_pibar}
\end{proposition}
\vskip -1.5em
To acquire intuition from Proposition~\ref{prop:pihat_leq_pibar}, note that once $a^\ast$ is selected, its count $n_{a^\ast}$ increases and so does the total count~$N$. As a result, $\pihat(a^\ast)$ increases (in the order of $\cO\prth{1/N}$) and further approximates $\pibar(a^\ast)$. As such, Proposition~\ref{prop:pihat_leq_pibar} shows that the action selection formula encourages the shape of $\pihat$ to be close to that of $\pibar$, until in the limit the two distributions coincide. 


Note that \Cref{prop:action_selection_diff}
and \Cref{prop:pihat_leq_pibar} are a special case of a more general result that we formally prove in \Cref{sec:proof_action_selection_pibar}. In this particular case, the proof relies on noticing that
\begin{align}
\nonumber&\argmax_a \brkt{\mathbf{q}_a + c \cdot \pi_\theta(a) \cdot \frac{\sqrt{\sum_b \boldsymbol{n}_b}}{1 + \boldsymbol{n}_a}} \tag{\ref{eq:azselection}} \\ 
&= \ \argmax_a \brkt{\pi_\theta(a) \cdot \prth{\frac{1}{\pihat(a)} -  \frac{1}{\pibar(a)}}}\cdot
\label{eq:action_selection_pibar}
\end{align}
Then, since $\sum_a \pihat(a) = \sum_a \pibar(a)$ and $\pihat >0$ and $\pibar >0$, there exists at least one action for which $0 < \pihat(a) \leq \pibar(a)$, \emph{i.e.,} $1/\pihat(a) - 1/\pibar(a) \geq 0$. 


To state a formal statement on $\pihat$ approximating $\pibar$, in \Cref{sec:one_over_N_bound} we expand the conclusion under the assumption that $\pibar$ is a  constant. In this case we can derive a bound for the convergence rate of these two distributions as $N$ increases over the search, 
\begin{align}
    ||\boldsymbol{\pibar} - \boldsymbol{\pihat}||_{\infty} \le \frac{|\mathcal{A}| - 1}{|\mathcal{A}|+ N}\CommaBin
\end{align}
with ${\cal O}(1/N)$ matching the lowest possible approximation error (see \Cref{sec:one_over_N_bound}) among discrete distributions of the form $\prth{k_i/N}_i$ for $k_i\in\mathbb{N}$.

\subsection{Generalization to common MCTS algorithms}\label{sec:uct_act_pi_bar}
Besides AlphaZero, UCT~\cite{kocsis2006bandit} is another heuristic with a selection criteria inspired by UCB, defined as 
\begin{align}
    \argmax_a \brkt{\mathbf{q}_a + c \cdot \sqrt{\frac{\log\prth{\sum_b \boldsymbol{n}_b}}{1 + \boldsymbol{n}_a}}\;}\cdot
    \label{eq:uctselection_noprior}
\end{align} 

Contrary to AlphaZero, the standard UCT formula does not involve a prior policy. In this section, we consider a slightly modified version of UCT with a (learned) prior $\pi_\theta$, as defined in \Cref{eq:uctselection}. By setting the prior $\pi_\theta$ to the uniform distribution, we  recover the original UCT formula,
\begin{align}
    \argmax_a \brkt{\mathbf{q}_a + c \cdot \sqrt{\pi_\theta(a)\cdot\frac{\log\prth{\sum_b \boldsymbol{n}_b}}{1 + \boldsymbol{n}_a}}\;}\cdot
    \label{eq:uctselection}
\end{align}
Using the same reasoning as in Section 3.3, we now show that this modified UCT formula also tracks the solution to a regularized policy optimization problem, thus generalizing our result to commonly used MCTS algorithms.

First, we introduce $\pibar_\text{UCT}$, which is tracked by the UCT visit distribution, as:
\begin{align}
\pibar_\text{UCT} \eqdef \argmax_{\mathbf{y} \in \mathcal{S}}  \mathbf{q}^T \mathbf{y} - \lambda_N^\text{UCT} D\prth{\mathbf{y}, \boldsymbol{\pi_\theta}},
\label{def:pibar_uct}
\end{align}
where $\displaystyle D(\boldsymbol{x},\boldsymbol{y}) \eqdef 2 - 2\sum_i\sqrt{\boldsymbol{x}_i\cdot \boldsymbol{y}_i}$ is an $f$-divergence\footnote{In particular $D(x,y) \ge 0, D(x,y) = 0 \implies x=y$ and $D(x,y)$ is jointly convex in $x$ and $y$ \cite{csiszar1964informationstheoretische, liese2006divergences}.}  
\begin{align*}
 \text{and\quad} \lambda_N^\text{UCT}(x) \eqdef c\cdot\sqrt{\frac{\log \sum_b \boldsymbol{n}_b}{|A| + \sum_b \boldsymbol{n}_b}}\cdot
\end{align*}

Similar to AlphaZero, $\lambda_N^\text{UCT}$ behaves\footnote{We ignore logarithmic terms.} as $\tilde O\prth{1/\sqrt{N}}$ and therefore the regularization gets weaker as $N$ increases. We can also derive tracking properties between $\pibar_\text{UCT}$ and the UCT empirical visit distribution $\pihat_\text{UCT}$ as we did for AlphaZero in the previous section, with Proposition~\ref{prop:uct}; as in the previous section, this is a special case of the general result with any $f$-divergence in \Cref{sec:proof_action_selection_pibar}.

\begin{proposition}{}
\label{prop:uct} We have that
\begin{align}
\nonumber&\ \argmax_a \brkt{\mathbf q_a  + c \cdot \sqrt{\pi_\theta(a) \cdot \frac{\log\prth{\sum_b \boldsymbol{n}_b}}{1 + \boldsymbol{n}_a}}\;} \\ 
&\nonumber=\ \argmax_a \brkt{\sqrt{\pi_\theta(a)} \cdot \prth{\frac{1}{\sqrt{\pihat(a)}} -  \frac{1}{\sqrt{\pibar_\text{UCT}(a)}}}}
\end{align}
and 
\begin{align}
    a^\ast_\text{UCT} = \argmax_a\brkt{\frac{\partial}{\partial \boldsymbol{n}_a}\prth{\mathbf{q}\transpose \boldsymbol{\pihat} - \lambda_N^\text{UCT} D\brkt{\mathbf{\boldsymbol{\pi_\theta}}, \boldsymbol{\pihat}}}}.
\end{align}
\end{proposition}

To sum up, similar to the previous section, we show that UCT's search policy $\pihat_\text{UCT}$ tracks the exact solution $\pibar_\text{UCT}$ of the regularized policy optimization problem of \Cref{def:pibar_uct}.


\section{Algorithmic benefits}
\label{sec:algo}
In \Cref{sec:mcts_as_po}, we introduced a distribution $\pibar$ as the solution to a regularized policy optimization problem. We then showed that AlphaZero, along with general MCTS algorithms, select actions such that the empirical visit distribution $\pihat$ actively approximates $\pibar$. Building on this insight, below we argue that $\pibar$ is preferable to $\pihat$, and we propose three  complementary algorithmic changes to AlphaZero.


\subsection{Advantages of using $\pibar$ over $\pihat$}
\label{sec:mcts_as_po-motivation}
MCTS algorithms produce Q-values as a by-product of the search procedure. However, MCTS does not directly use search Q-values to compute the policy, but instead  uses the visit distribution $\pihat$ (search Q-values implicitly influence $\pihat$ by guiding the search). We postulate that this degrades the performance especially at low simulation budgets $N_\text{sim}$ for several reasons:
\begin{enumerate}
    \item When a  promising new (high-value) leaf is discovered, many additional simulations might be needed before this information is reflected in $\pihat$; since $\pibar$ is  directly computed from Q-values, this information is updated instantly.
    \item By definition (\Cref{eq:pihat}), $\pihat$ is the ratio of two integers and has limited expressiveness when $N_\text{sim}$ is low, which might lead to a sub-optimal policy; $\pibar$ does not have this constraint.
    \item The prior $\pi_\theta$ is trained against the target $\pihat$, but the latter is only improved for actions that have been sampled at least once during search. Due to the deterministic action selection (\Cref{eq:azselection}), this may be problematic for certain actions that would require a large simulation budget to be sampled even once. 
\end{enumerate}



The above downsides cause MCTS to be highly sensitive to simulation budgets $N_\text{sim}$. When $N_\text{sim}$ is high relative to the branching factor of the tree search, i.e., number of actions, MCTS algorithms such as AlphaZero perform well. However, this performance drops significantly when $N_\text{sim}$ is low as showed by \citet{hamrick2019combining}; see also \emph{e.g.}, Figure~3.D. by \citet{schrittwieser2019mastering}.

\begin{figure}[t]\centering
\includegraphics[width=0.9\linewidth]{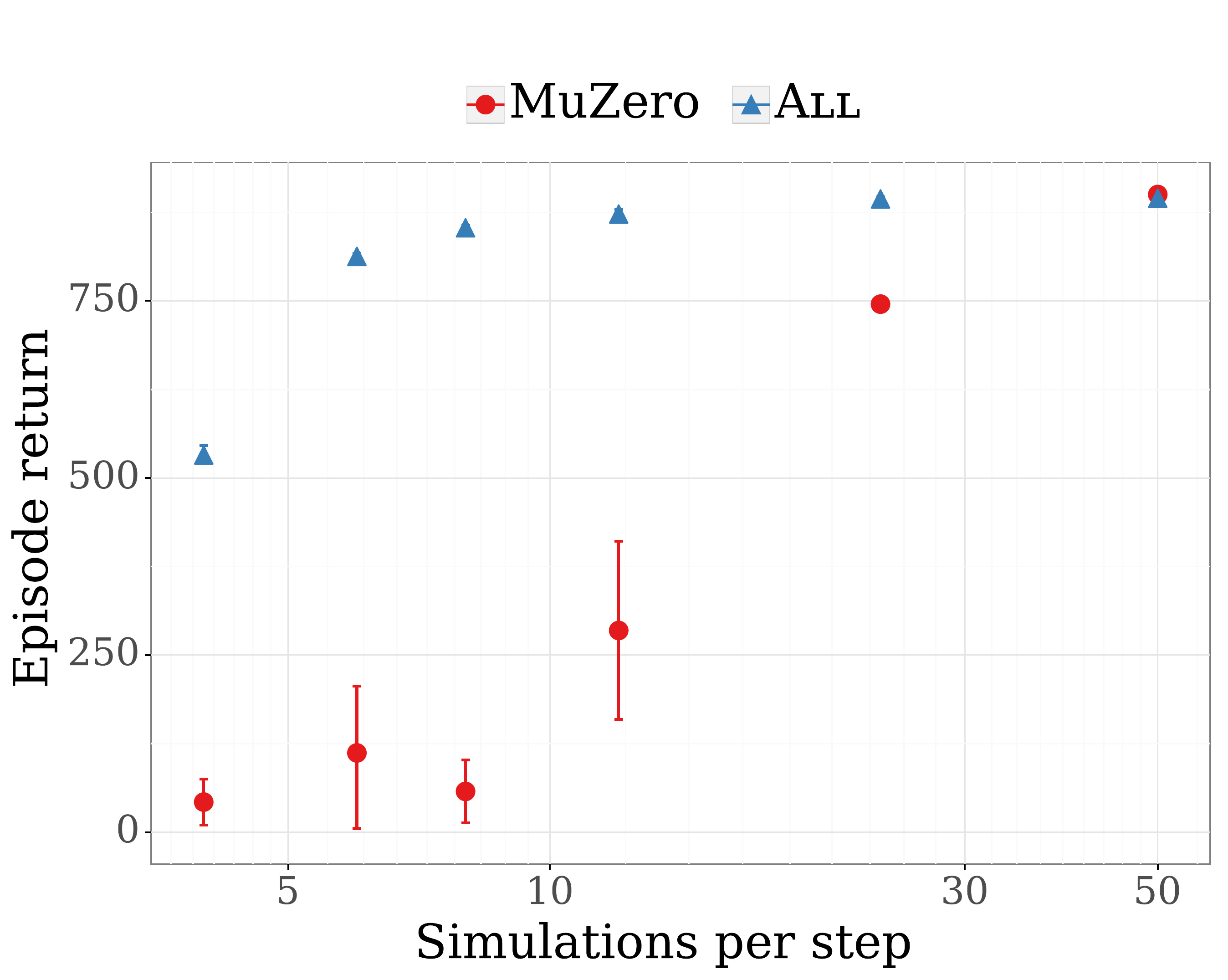}\caption{\label{fig:mz_vs_pz_cheetah_asympt}Comparison of the  score (median score over 3 seeds) of MuZero (red: using $\pihat$) and \algoall{} (blue:  using $\pibar$) after 100k learner steps as a function of $N_\text{sim}$ on Cheetah Run of the Control Suite.}\vspace*{-.3cm}
\end{figure}

We illustrate the effect of simulation budgets in \Cref{fig:mz_vs_pz_cheetah_asympt}, where $x$-axis shows the budgets $N_\text{sim}$ and $y$-axis shows the episodic performance of algorithms applying $\pihat$ vs.\,$\pibar$; see the details of these algorithms in the following sections. We see that~$\pihat$ is highly sensitive to simulation budgets while~$\pibar$ performs consistently well across all budget values.



\subsection{Proposed improvements to AlphaZero}
We have pointed out potential issues due to $\pihat$. We now detail how to use $\pibar$ as a replacement to resolve such issues.\footnote{Recall that we have identified three issues. Each algorithmic variant below helps in addressing issue~1 and~2. Furthermore, the \algolearn{} variant helps address issue 3.} 
\Cref{app:compute_pibar} shows how to compute $\pibar$ in practice.
\paragraph{\algoact{}: acting with $\pibar$} 
AlphaZero acts in the real environment by sampling actions according to  $a\sim\pihat(\cdot|x_\text{root})$. Instead, we propose to to sample actions sampling according to $a\sim\pibar(\cdot|x_\text{root})$. We label this variant as \algoact{}.

\paragraph{\algosearch{}: searching with $\pibar$}
During search, we propose to stochastically sample actions according to $\pibar$ instead of the deterministic action selection rule of \Cref{eq:azselection}. At each node $x$ in the tree,  $\pibar(\cdot)$ is computed with Q-values and total visit counts at the node based on \Cref{def:pibar}.
We label this variant as \algosearch{}.

\paragraph{\algolearn{}: learning with $\pibar$}
AlphaZero computes locally improved policy with tree search and distills such improved policy into $\pi_\theta$. We propose to use $\pibar$ as the target policy in place of $\pihat$ to train our prior policy. As a result, the parameters are updated as  
\begin{align}    \label{eq:learnupdate} \theta \leftarrow \theta - \eta \nabla_\theta \mathbb{E}_{x_\text{root}}\Big[\mathbb{KL}\brkt{\pibar(\cdot|x_\text{root}), \pi_\theta(\cdot|x_\text{root})}\Big],
\end{align}
where $x_\text{root}$ is sampled from a prioritized replay buffer as in AlphaZero. We label this variant as \algolearn{}.


\paragraph{\algoall{}: combining them all} We refer to the combination of these three independent variants as \algoall{}. \Cref{app:implementation} provides additional implementation details. 

\paragraph{Remark} Note that AlphaZero entangles search and learning, which is not desirable. For example, when the action selection formula changes, this impacts not only intermediate search results but also the root visit distribution $\pihat(\cdot|x_\text{root})$, which is also the learning target for $\pi_\theta$. However, the \algolearn{} variant partially disentangles these components. Indeed, the new learning target is $\pibar(\cdot|x_\text{root})$ which is computed from search Q-values, rendering it less sensitive to e.g., the action selection formula.

\subsection{Connections between AlphaZero and model-free policy optimization.}
\label{sec:connection_az_po}
Next, we make the explicit link between proposed algorithmic variants and existing policy optimization algorithms. First, we provide two complementary interpretations.

\paragraph{\algolearn{} as policy optimization} For this interpretation, we treat \algosearch{} as a blackbox, \emph{i.e.,} a subroutine that takes a root node $x$ and returns statistics such as search Q-values.

Recall that policy optimization (\Cref{eq:po_opt}) maximizes the objective $\approx \mathcal{Q}_{\pi_\theta}\transpose \mathbf{y}$ with the local policy $\mathbf{y}$. There are many model-free methods for the estimation of $\mathcal{Q}_{\pi_\theta}$, ranging from Monte-Carlo estimates  of cumulative returns $\mathcal{Q}^{\pi_\theta}\approx \sum_{t\geq 0} \gamma^t r_t$ \citep{schulman2015trust,schulman2017proximal} to using predictions from a Q-value critic $\mathcal{Q}_{\pi_\theta} \approx \mathbf{q}_\theta$ trained with off-policy samples \citep{abdolmaleki2018maximum,song2019v}. When solving $\pibar$ for the update (\Cref{eq:learnupdate}), we can interpret \algolearn{} as a policy optimization algorithm using tree search to estimate $\mathcal{Q}_{\pi_\theta}$. Indeed, \algolearn{} could be interpreted as building a Q-function\footnote{During search, because child nodes have fewer simulations than the root, the Q-function estimate at the root slightly under-estimates the acting policy Q-function.} critic $\mathbf{q}_\theta$ with a tree-structured inductive bias. However, this inductive bias is not built-in a network architecture
  \citep{silver2017predictron,farquhar2017treeqn,oh2017value,guez2018learning}, but constructed online by an algorithm, i.e., MCTS. Next, \algolearn{} computes the locally optimal policy $\pibar$ to the regularized policy optimization objective and distills $\pibar$ into $\pi_\theta$. This is exactly the approach taken by MPO  \cite{abdolmaleki2018maximum}.


\paragraph{\algosearch{} as policy optimization} We now unpack the algorithmic procedure of the tree search, and show that it can also be interpreted as policy optimization.

During the forward simulation phase of \algosearch{}, the action at each node $x$ is selected by sampling $a \sim \pibar (\cdot|x)$. As a result, the full imaginary trajectory is generated consistently according to policy $\pibar$. During backward updates, each encountered node $x$ receives a backup value from its child node, which is an exact estimate of $Q^{\pibar}(x,a)$. Finally, the local policy $\pibar(\cdot|x)$ is updated by solving the constrained optimization problem of \Cref{def:pibar}, leading to an improved policy over previous $\pibar(\cdot|x)$. Overall, with $N_\text{sim}$ simulated trajectories, \algosearch{} optimizes the root policy $\pibar(\cdot|x_\text{root})$ and root search Q-values, by carrying out $N_\text{sim}$ sequences of MPO-style updates across the entire tree.\footnote{Note that there are several differences from typical model-free implementations of policy optimization: most notably, unlike a fully-parameterized policy, the tree search policy is tabular at each node. This also entails that the MPO-style distillation is exact.} A highly related approach is to update local policies via policy gradients \citep{anthony2019policy}.

By combining the above two interpretations, we see that the \algoall{} variant is very similar to a full policy optimization algorithm. Specifically, on a high level, \algoall{} carries out MPO updates with search Q-values. These search Q-values are also themselves obtained via MPO-style updates within the tree search. This
paves the way to our major revelation stated next.
\begin{observation}

\algoall{} can be interpreted as regularized policy optimization. Further, since $\pihat$ approximates $\pibar$, AlphaZero and other MCTS algorithms can be interpreted as approximate regularized policy optimization. \vspace*{-.3cm}
\end{observation}




\begin{figure}[t]\centering
\includegraphics[width=\linewidth]{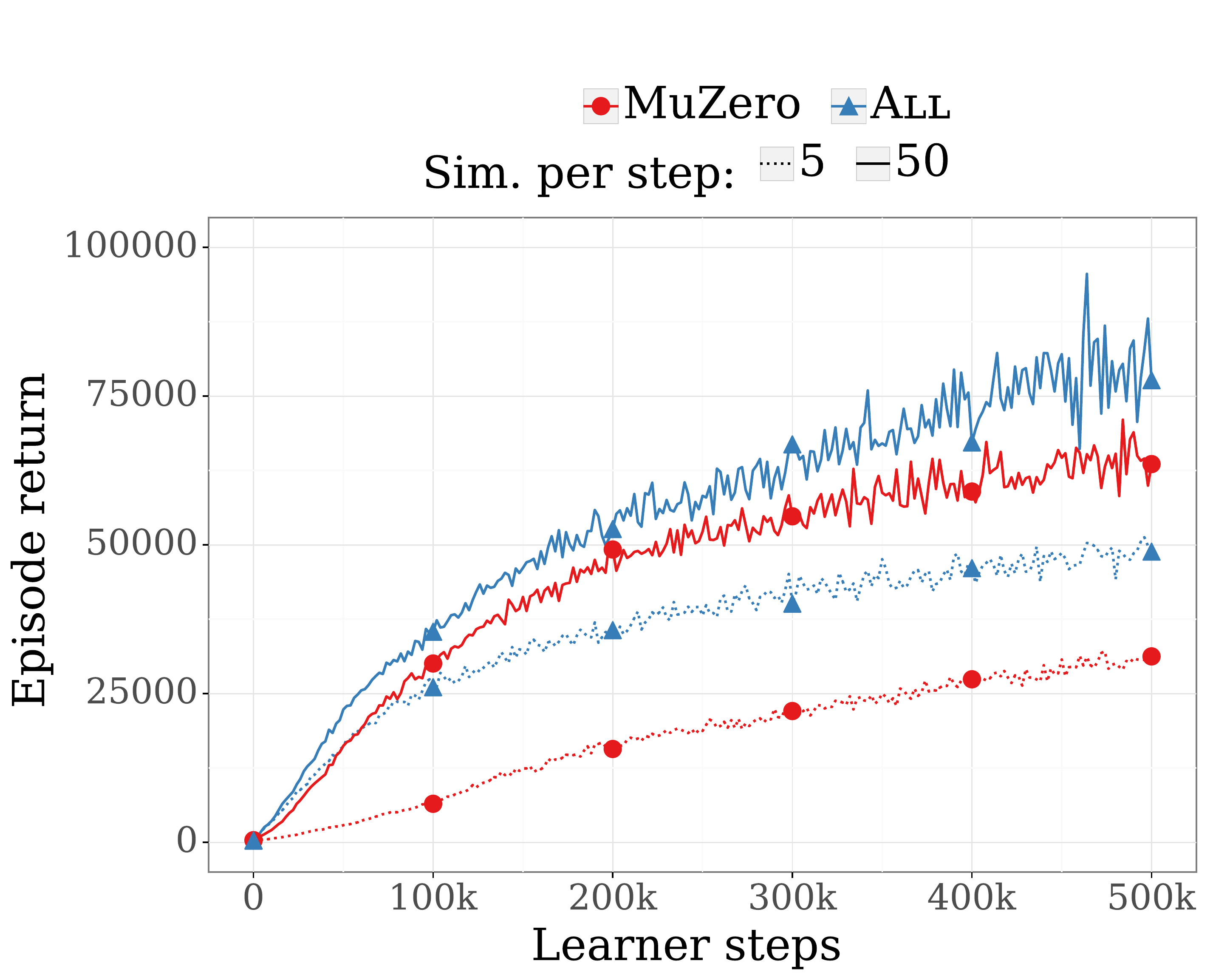}
\caption{\label{fig:mz_vs_pz_5_50_sims}Comparison of median scores of  MuZero (red) and \algoall{} (blue) at $N_\text{sim}=5$ (dotted line) and $N_\text{sim}=50$ (solid line) simulations per step on Ms Pacman (Atari). Averaged across 8 seeds.} \vspace*{-0.3cm}
\end{figure}

\section{Experiments}
\label{sec:experiments}


In this section, we aim to address several questions: \textbf{(1)} How sensitive are state-of-the-art hybrid algorithms such as AlphaZero to low simulation budgets and can the \algoall{} variant provide a more robust alternative? \textbf{(2)} What changes among \algoact{}, \algosearch{}, and \algolearn{} are most critical in this variant performance? \textbf{(3)} How does the performance of the \algoall{} variant compare with AlphaZero in environments with large branching factors?

\paragraph{Baseline algorithm} Throughout the experiments, we take MuZero \citep{schrittwieser2019mastering} as the baseline algorithm. As a variant of AlphaZero, MuZero applies tree search in learned models instead of real environments, which makes it applicable to a wider range of problems. Since MuZero shares the same search procedure as AlphaGo, AlphaGo Zero, and AlphaZero, we expect the performance gains to be transferable to these algorithms. Note that the results below were obtained with a scaled-down version of MuZero, which is described in \Cref{app:baby_zero}. 

\paragraph{Hyper-parameters} The hyper-parameters of the algorithms are tuned to achieve the maximum possible performance for baseline MuZero on the Ms Pacman level of the Atari suite~\citep{bellemare2013arcade}, and are identical in all experiments with the exception of the number of simulations per step $N_\text{sim}$.\footnote{The number of actors is scaled linearly with $N_\text{sim}$ to maintain the same total number of generated frames per second.} In particular, no further tuning was required for the \algolearn{}, \algosearch{}, \algoact{}, and \algoall{} variants, as was expected from the theoretical considerations of \Cref{sec:mcts_as_po}.


\begin{figure}[t]\centering
\begin{subfigure}{0.48\linewidth}
\includegraphics[width=\linewidth]{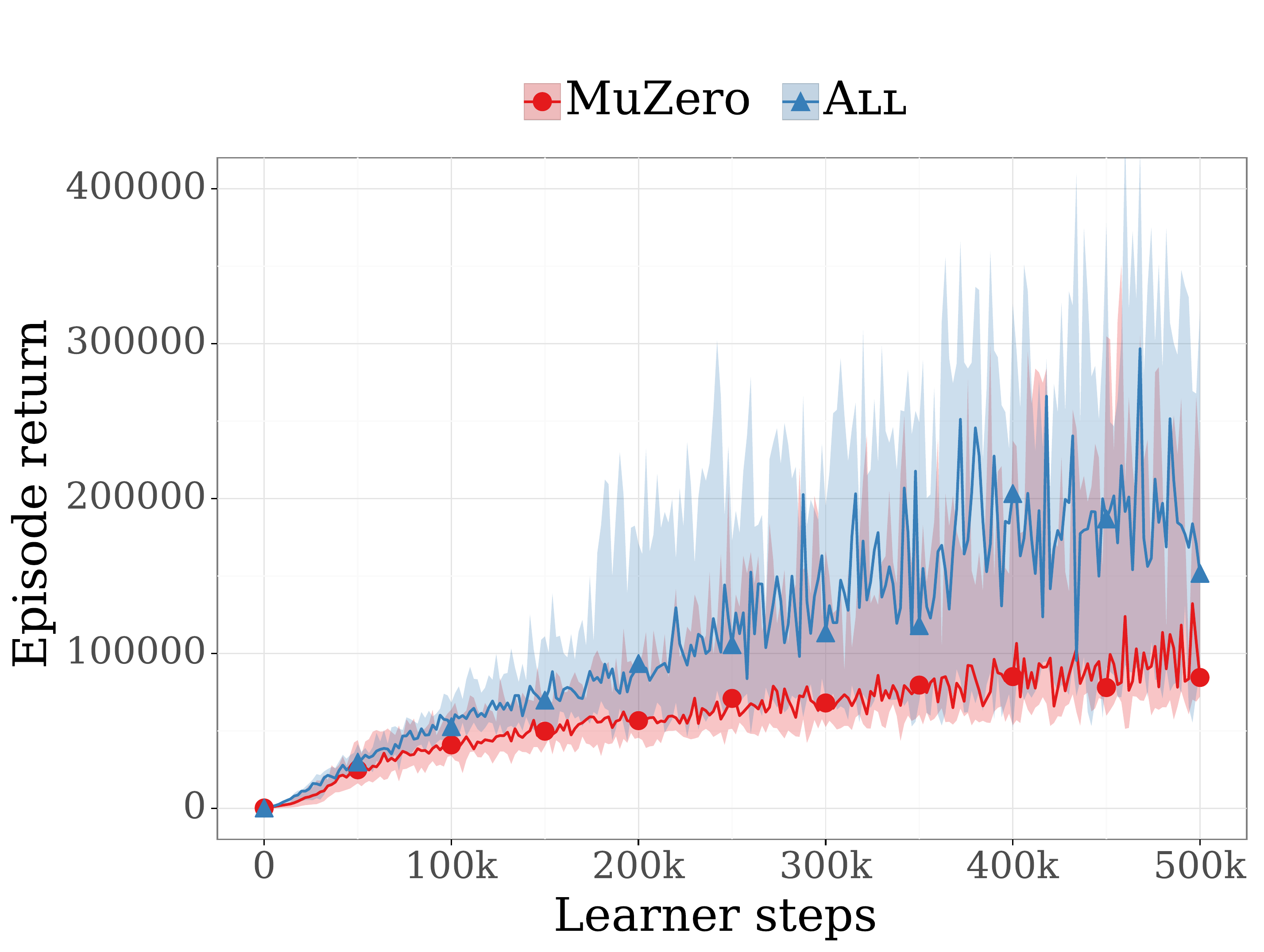}\caption{Alien}
\end{subfigure}
\begin{subfigure}{0.48\linewidth}
\includegraphics[width=\linewidth]{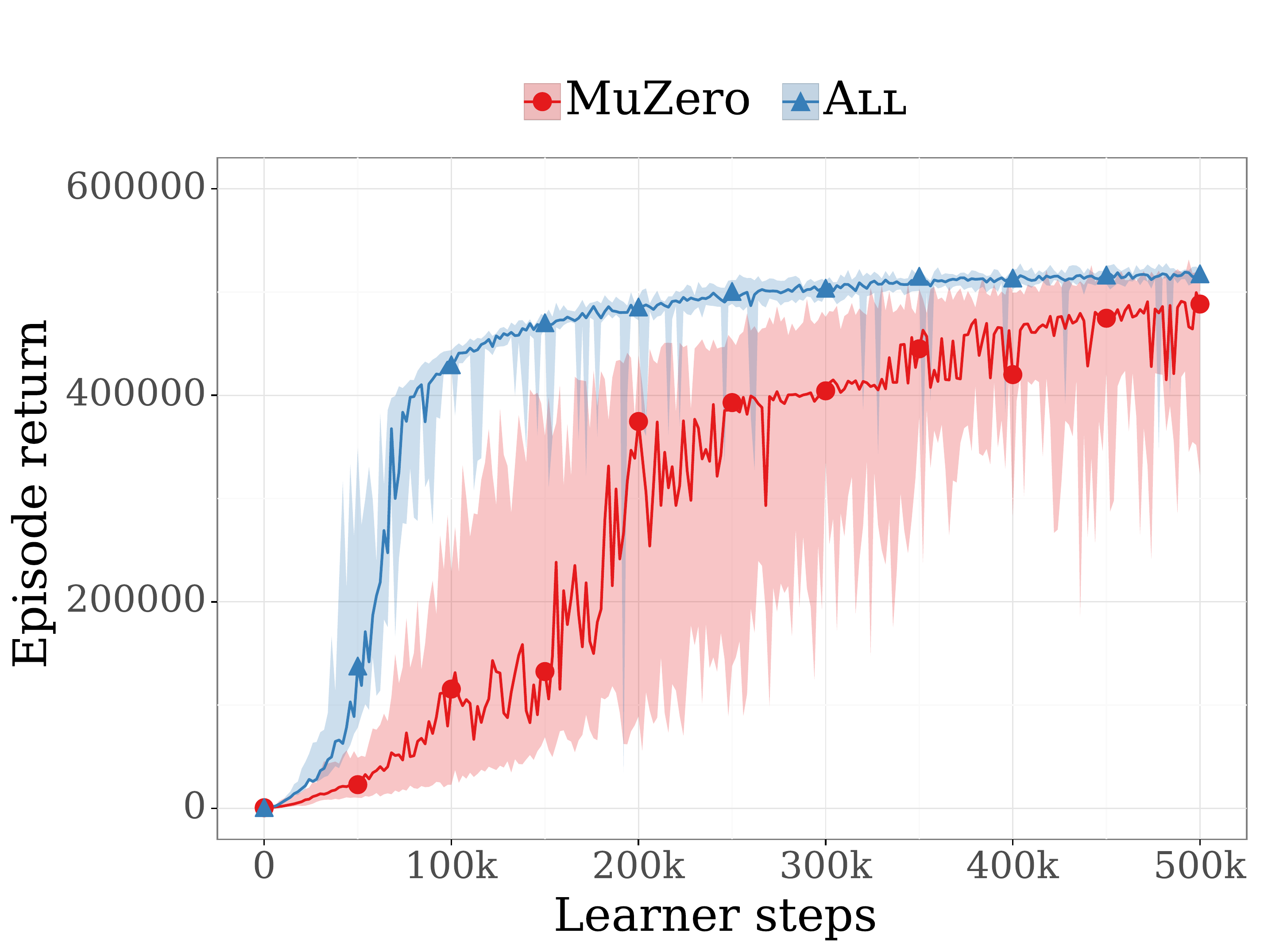}\caption{Asteroids}
\end{subfigure}
\begin{subfigure}{0.48\linewidth}
\includegraphics[width=\linewidth]{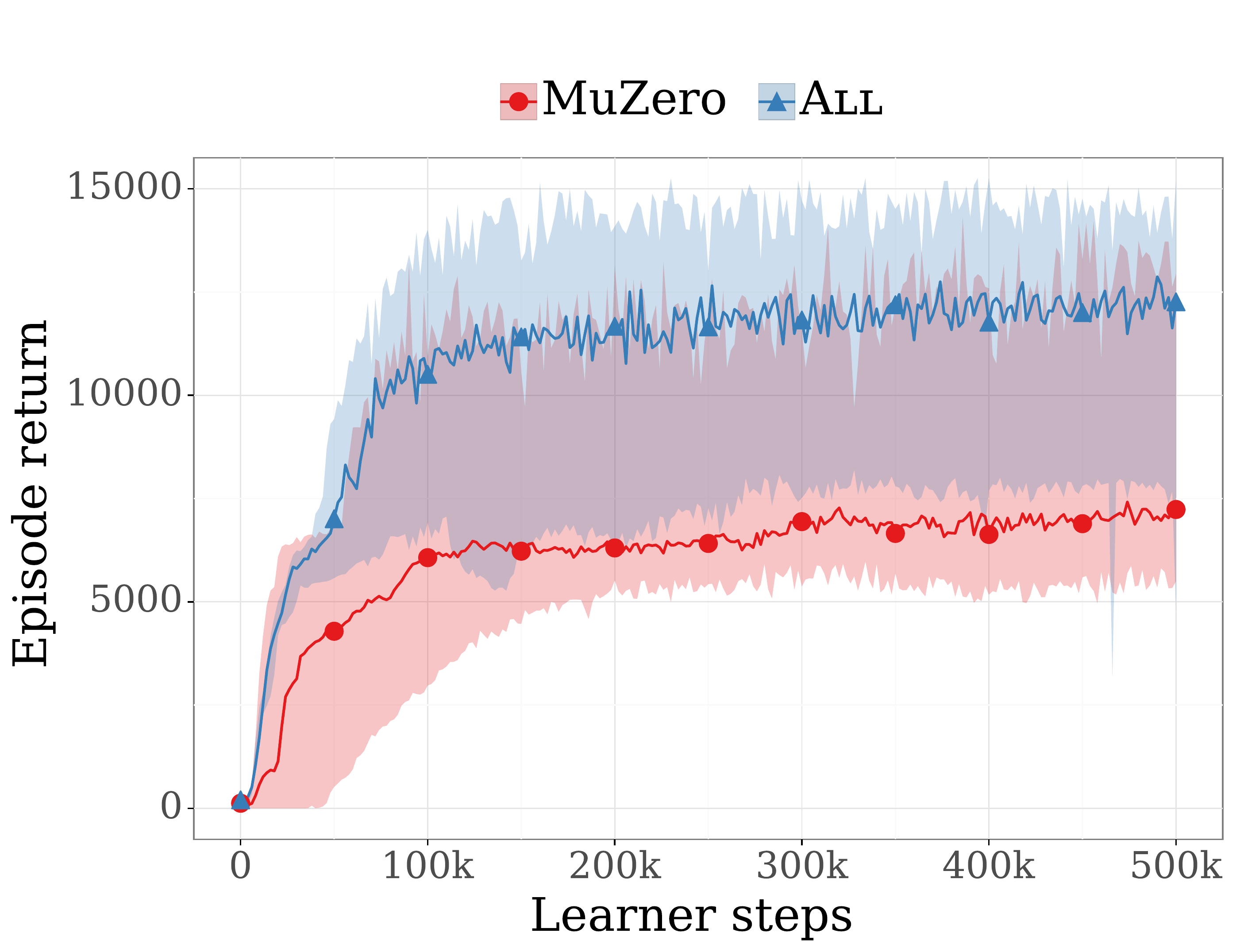}\caption{Gravitar}
\end{subfigure}
\begin{subfigure}{0.48\linewidth}
\includegraphics[width=\linewidth]{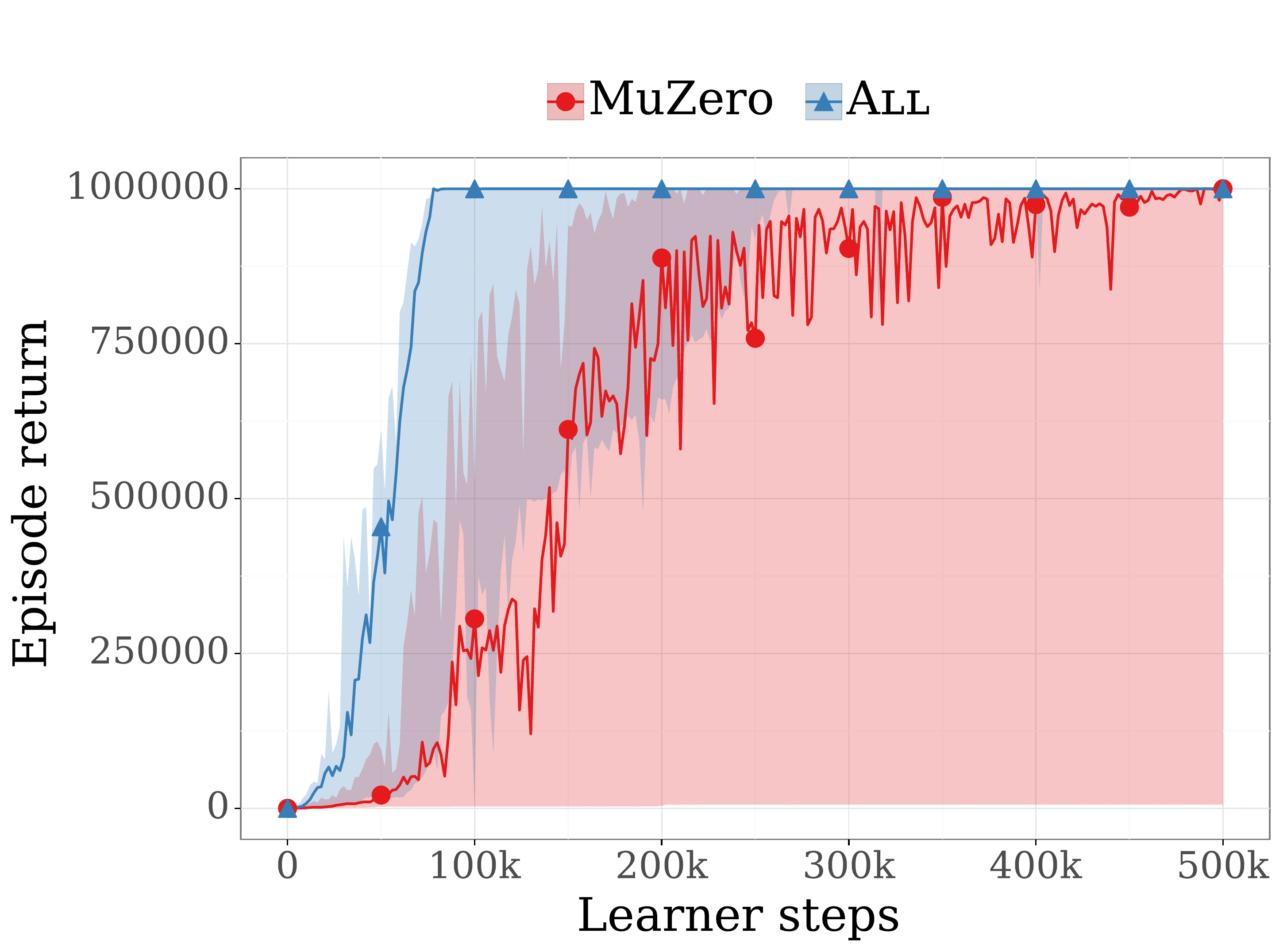}\caption{Seaquest}
\end{subfigure}
\caption{\label{fig:sixgames_intext}Comparison of median score (solid lines) over 6 seeds of MuZero and \algoall{} on four Atari games with $N_\text{sim}=50$. The shaded areas correspond to the range of the best and worst seed. \algoall{} (blue) performs consistently better than MuZero (red). }
\end{figure}

\subsection{Search with low simulation budgets}\label{sec:mz_vs_pz_ms_pacman}
Since AlphaZero solely relies on the $\pihat$ for training targets, it may misbehave when simulation budgets $N_\text{sim}$ are low. On the other hand, our new algorithmic variants might perform better in this regime. To confirm these hypotheses, we compare the performance of MuZero and the \algoall{} variant on the Ms Pacman level of the Atari suite at different levels of simulation budgets.

\paragraph{Result} In \Cref{fig:mz_vs_pz_5_50_sims}, we  compare the episodic return of \algoall{} vs.\,MuZero averaged over 8 seeds, with a simulation budget $N_\text{sim} = 5$ and $N_\text{sim} = 50$ for an action set of size $|\mathcal A| \leq 18$; thus, we consider that $N_\text{sim}=5$ and $N_\text{sim}=50$ respectively correspond to a low and high simulation budgets relative to the number of actions. We make several observations: \textbf{(1)} At a relatively high simulation budget, $N_\text{sim} = 50$, same as~\citet{schrittwieser2019mastering}, both MuZero and \algoall{} exhibit reasonably close levels of performance; though \algoall{} obtains marginally better performance than MuZero; \textbf{(2)} At low simulation budget, $N_\text{sim} = 5$, though both algorithms suffer in performance relative to high budgets, \algoall{} significantly outperforms MuZero both in terms of learning speed and asymptotic performance; \textbf{(3)} \Cref{fig:mz_vs_pz_ms_pacman} in \Cref{app:mz_vs_pz_ms_pacman} shows that this behavior is consistently observed at intermediate simulation budgets, with the two algorithms starting to reach comparable levels of performance when $N_\text{sim} \geq 24$ simulations. These observations confirm the intuitions from \Cref{sec:mcts_as_po}. \textbf{(4)} We provide results on a subset of Atari games in \Cref{fig:sixgames_intext}, which show that the performance gains due to $\pibar$ over $\pihat$ are also observed in other levels than Ms Pacman; see \Cref{app:mz_vs_pz_cc} for results on additional levels. This subset of levels are selected based on the experiment setup in Figure~S1 of  \citet{schrittwieser2019mastering}. Importantly, note that the performance gains of \algoall{} are consistently significant across selected levels, even at a higher simulation budget of $N_\text{sim}=50$.

\begin{figure}\centering
\begin{subfigure}{0.94\columnwidth}
\includegraphics[width=\linewidth]{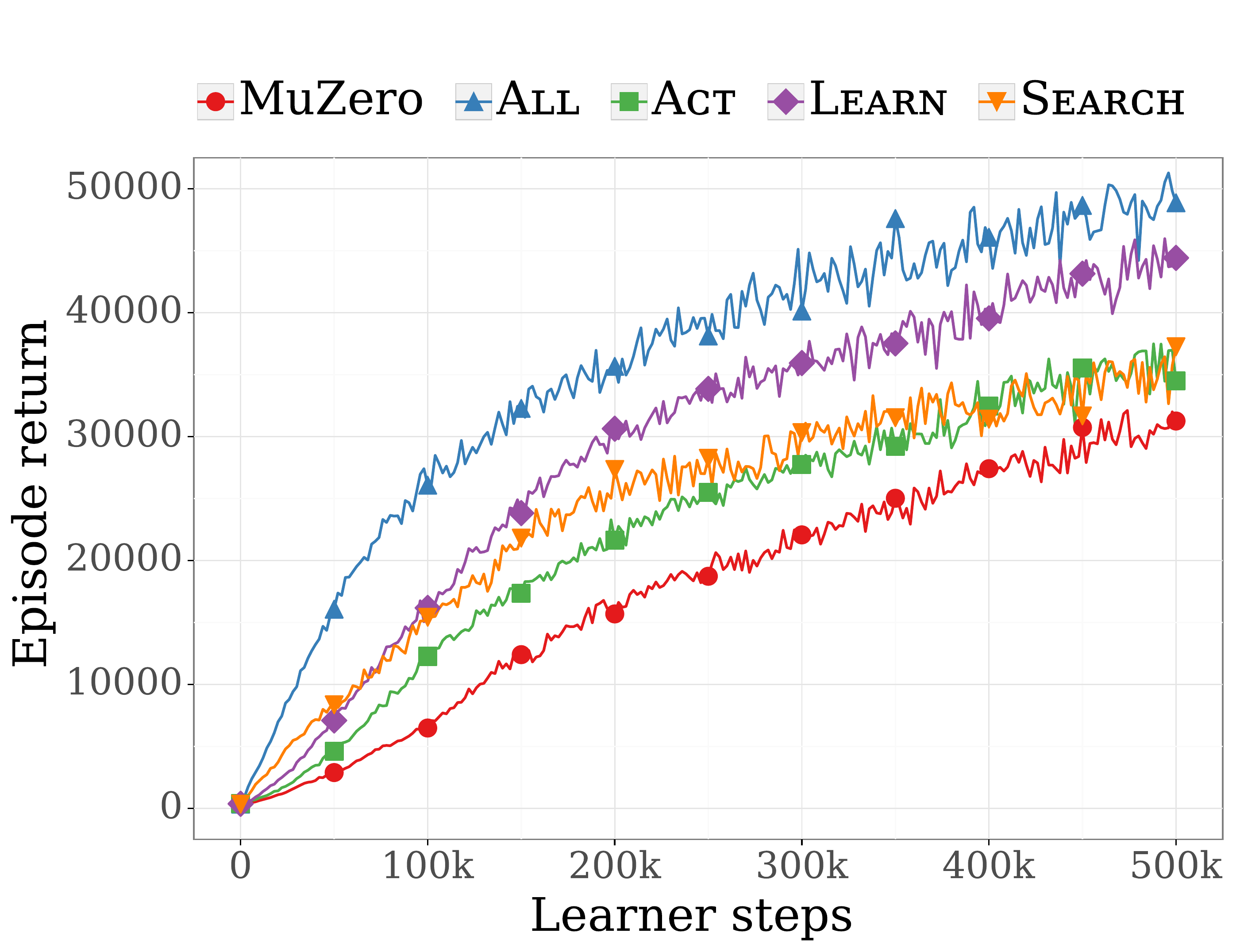}\caption{\label{fig:pz_ablation_5_sims}5 simulations}
\end{subfigure} 
\begin{subfigure}{0.94\columnwidth}
\includegraphics[width=\linewidth]{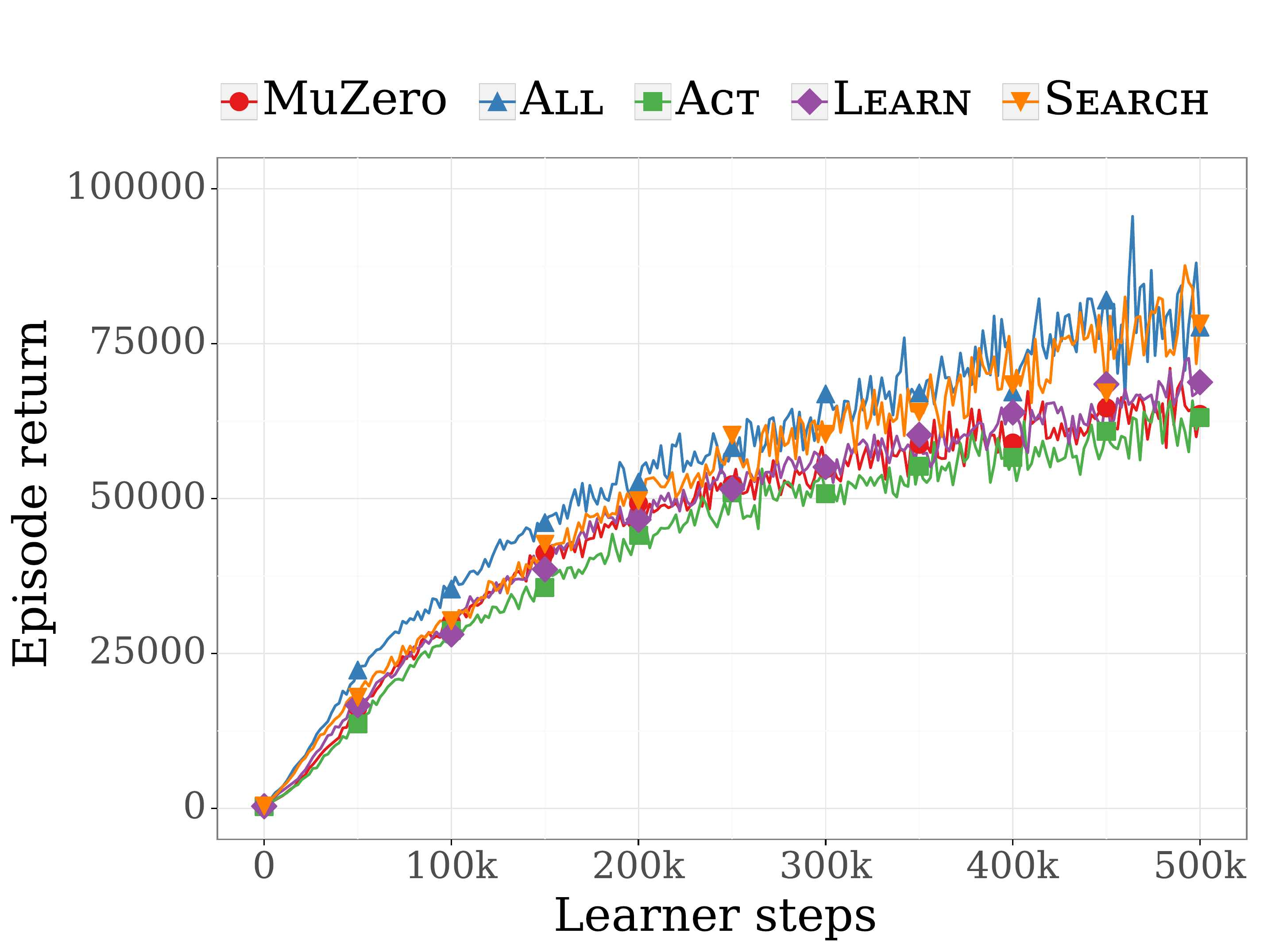} \caption{\label{fig:pz_ablation_50_sims} 50 simulations}
\end{subfigure}
\caption{\label{fig:pz_ablation}Ablation study at 5 and 50 simulations per step on Ms Pacman (Atari); average across 8 seeds.}
\label{fig:ablation}
\end{figure}

\subsection{Ablation study}
To better understand which component of the \algoall{} contributes the most to the performance gains, \Cref{fig:pz_ablation} presents the results of an ablation study where we compare individual component \algolearn{}, \algosearch{}, or \algoact{}. 

\paragraph{Result} The comparison is shown in  \Cref{fig:pz_ablation}, we make several observations: \textbf{(1)} At $N_\text{sim}=5$ (\Cref{fig:pz_ablation_5_sims}), the main improvement comes from using the policy optimization solution $\pibar$ as the learning target (\algolearn{} variant); using~$\pibar$ during search or acting leads to an additional marginal improvement; \textbf{(2)} Interestingly, we observe a different behavior at $N_\text{sim}=50$ (\Cref{fig:pz_ablation_50_sims}). In this case, using $\pibar$ for learning or acting does not lead to a noticeable improvement. However, the superior performance of \algoall{} is mostly due to sampling according to~$\pibar$ during search (\algosearch{}). 

The improved performance when using $\pibar$ as the learning target (\algolearn{}) illustrates the theoretical considerations of \Cref{sec:mcts_as_po}: at low simulation budgets, the discretization noise in $\pihat$ makes it a worse training target than $\pibar$, but this advantage vanishes when the number of simulations per step increases. As predicted by the theoretical results of \Cref{sec:mcts_as_po}, learning and acting using $\pibar$ and $\pihat$ becomes equivalent when the simulation budget increases.

On the other hand, we see a slight but  significant improvement when sampling the next node according to~$\pibar$ during search (\algosearch{}) regardless of the simulation budget. This could be explained by the fact that even at high simulations budget, the \algosearch{} modification also affect deeper node that have less simulations. 

\subsection{Search with large action space --  continuous control}\label{sec:mz_vs_pz_cc}
The previous results confirm the intuitions presented in \Cref{sec:mcts_as_po,sec:algo}; namely, the \algoall{} variation  greatly improves performance at low simulation budgets, and obtain marginally higher performance at high simulation budgets. Since simulation budgets are relative to the number of action, these improvements are  critical in tasks with a high number of actions, where MuZero might require a prohibitively high simulation budgets; prior work \citep{dulac2015deep,metz2017discrete,van2020q} has already identified continuous control tasks as an interesting testbed. 

\paragraph{Benchmarks} We select high-dimensional environments from DeepMind Control Suite~\citep{tassa2018control}. The observations are images and action space $\mathcal{A}=[-1,1]^m$ with~$m$ dimensions. We apply an action discretization method similar to that of \citet{tang2019discretizing}. In short, for a continuous action space $m$ dimensions, each dimension is discretized into $K$ evenly spaced atomic actions. With proper parameterization of the policy network (see, \emph{e.g.}, \Cref{app:continuous_action_discretization}), we can reduce the effective branching factor to $Km \ll K^m$, though this still results in a much larger action space than Atari benchmarks. In
\Cref{app:mz_vs_pz_cc}, we provide additional descriptions of the tasks.

\begin{figure}\centering\vspace*{-.3cm}
\includegraphics[width=0.95\linewidth]{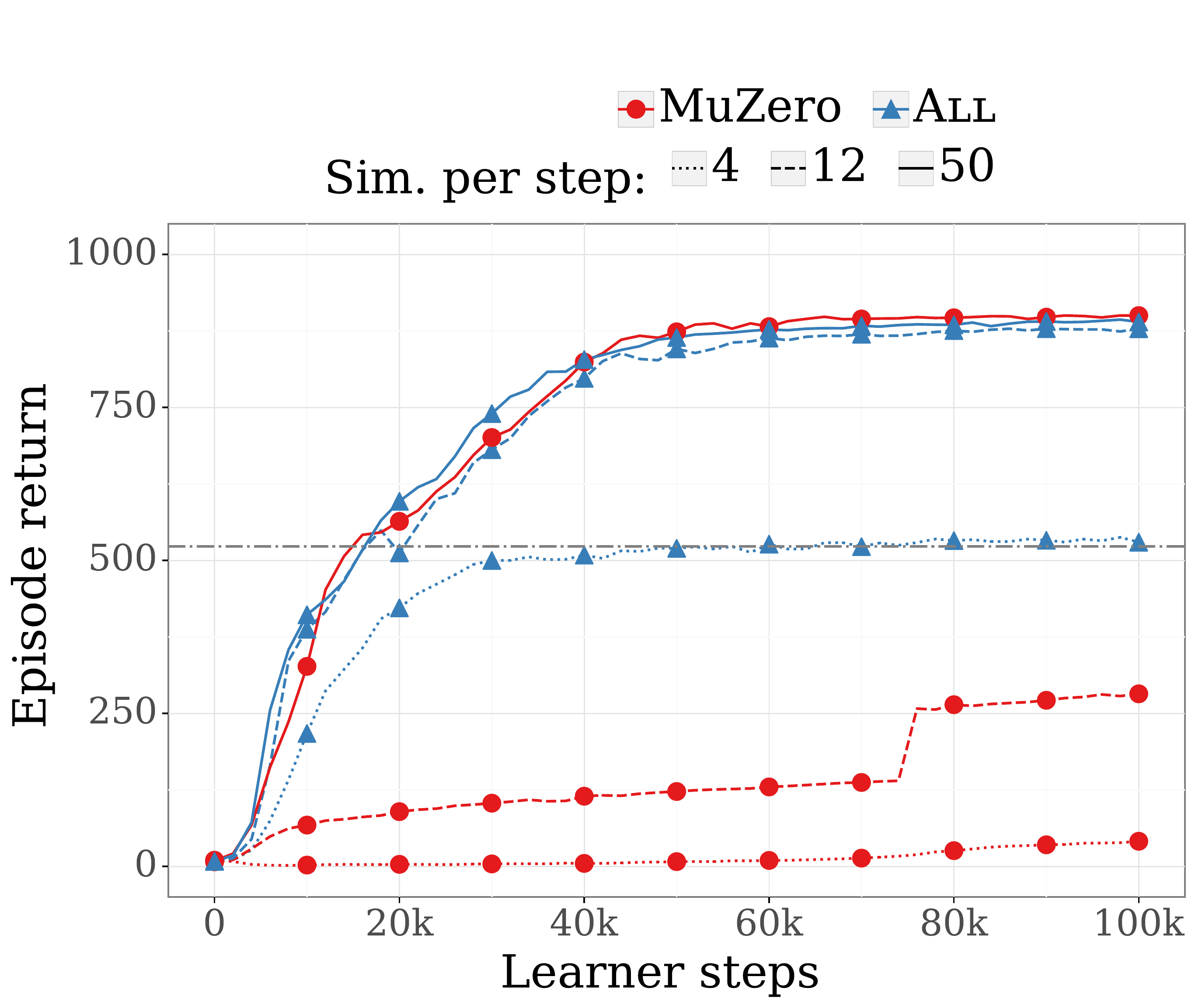}
\caption{\label{fig:mz_vs_pz_cc}Comparison of the median score over 3 seeds of MuZero (red) and \algoall{} (blue) at 4 (dotted) and 50 (solid line) simulations per step on Cheetah Run (Control Suite). }
\end{figure}

\paragraph{Result} In \Cref{fig:mz_vs_pz_cc}, we compare MuZero with the \algoall{} variant on the CheetahRun environment of the DeepMind Control Suite~\citep{tassa2018control}. We evaluate the performance at low ($N_\text{sim} = 4$), medium ($N_\text{sim} = 12$) and ``high'' ($N_\text{sim} = 50$) simulation budgets, for an effective action space of size 30 ($m=6$, $K=5$). The horizontal line corresponds to the performance of model-free D4PG also trained on pixel observations~\citep{barth-maron2018distributional}, as reported in~\citep{tassa2018control}. \Cref{app:mz_vs_pz_cc} provides experimental results on additional tasks. We again observe that \algoall{} outperforms the original MuZero at low simulation budgets and still achieves faster convergence to the same asymptotic performance with more simulations. \Cref{fig:mz_vs_pz_cheetah_asympt} compares the asymptotic performance of MuZero and \algoall{} as a function of the simulation budget at 100k learner steps.

%

\paragraph{Conclusion}
In this paper, 
we showed that the action selection formula used in MCTS algorithms, most notably AlphaZero,  approximates the solution to a regularized policy optimization problem formulated with search Q-values. 
From this theoretical insight, we proposed variations of the original AlphaZero algorithm by explicitly using the exact policy optimization solution instead of the approximation. We show experimentally that these variants achieve much higher performance at low simulation budget, while also providing statistically significant improvements when this budget increases. 

Our analysis on the behavior of model-based algorithms (i.e., MCTS) has made explicit connections  to model-free algorithms. We hope that this sheds light on new ways of combining both paradigms and opens doors to future ideas and improvements.

\paragraph{Acknowledgements} 
The authors would like to thank 
Alaa Saade, 
Bernardo Avila Pires, 
Bilal Piot, 
Corentin Tallec, 
Daniel Guo, 
David Silver,
Eugene Tarassov, 
Florian Strub,  
Jessica Hamrick, 
Julian Schrittwieser, 
Katrina McKinney, 
Mohammad Gheshlaghi Azar, 
Nathalie Beauguerlange, 
Pierre M\'enard, 
Shantanu Thakoor, 
Th\'eophane Weber, 
Thomas Mesnard, 
Toby Pohlen 
and the DeepMind team. 

%

\clearpage
\bibliographystyle{apalike}
\bibliography{refs}


\appendix
\onecolumn
\section{Details on search for AlphaZero}\label{app:alphazero-search}

Below we briefly present details of the search procedure for AlphaZero. Please refer to the original work \citep{silver2017mastering} for more comprehensive explanations.

As explained in the main text, the search procedure starts with a MDP state $x_0$, which is used as the root node of the tree. The rest of this tree is progressively built as more simulations are generated. In addition to Q-function $Q(x,a)$, prior $\pi_\theta(x,a)$ and visit counts $n(x,a)$, each node also maintains a reward $R(x,a) = r(x,a)$ and value $V(x)$ estimate.

In each simulation, the search consists of several parts: \textbf{Selection}, \textbf{Expansion} and \textbf{Backup}, as below.

\paragraph{Selection.} From the root node $x_0$, the search traverses the tree using the action selection formula of \Cref{eq:azselection} until a leaf node $x_l$ is reached. 

\paragraph{Expansion.} After a leaf node $x_l$ is reached, the search selects an action from the leaf node, generates the corresponding child node $x_c$ and appends it to the tree $\mathcal T$. The statistics for the new node are then initialized to $Q(x_c,a) = \min_{x\in \mathcal{T}, a'\in \mathcal{A}} Q(x,a')$ (pessimistic initialization), $n(x,a) = 0$ for $\forall a\in \mathcal{A}$.

\paragraph{Back-up.} The back-up consists of updating statistics of nodes encountered during the forward traversal. Statistics that need updating include the Q-function $Q(x,a)$, count $n(x,a)$ and value $V(x)$. The newly expanded node $n_c$ updates its value $V(x)$ to be either the Monte-Carlo estimation from random rollouts (e.g. board games) or a prediction of the value network (e.g. Atari games). For the other nodes encountered during the forward traversal, all other statistics are updated as follows:
\begin{align}
    V(x) &\leftarrow (V(x)\cdot \sum_b n(x,b) + (R(x,a) + \gamma V(\text{child}(x,a)) / (1 + \sum_b n(x,b)) \nonumber \\
    Q(x,a) &\leftarrow R(x,a) + \gamma V(\text{child}(x,a)), \nonumber \\
    n(x,a) &\leftarrow n(x,a) + 1, \nonumber
\end{align}
where $\text{child}(x,a)$ refers to the child node obtained by taking action $a$ from node $x$. 

Note that, in order to make search parameters agnostic to the scale of the numerical rewards (and, therefore, values), Q-function statistics $Q(x,a)$ are always normalized by statistics in the search tree before applying the action selection formula; in practice, \Cref{eq:azselection} uses the normalized $Q_z(x,a)$ defined as:
\begin{align}
    Q_z(x,a) = \frac{Q(x,a) - \min_{x\in \mathcal{T}, a\in \mathcal{A}} Q(x,a)}{\max_{x\in \mathcal{T}, a\in \mathcal{A}} Q(x,a) - \min_{x\in \mathcal{T}, a\in \mathcal{A}} Q(x,a)}.
\end{align}

\section{Implementation details}\label{app:implementation}

\subsection{Agent}\label{app:baby_zero}
For ease of implementation and availability of computational resources, the experimental results from \Cref{sec:experiments} were obtained with a scaled-down version of MuZero~\citep{schrittwieser2019mastering}. In particular, our implementation uses smaller networks compared to the architecture described in Appendix F of~\citep{schrittwieser2019mastering}: we use only 5 residual blocks with 128 hidden layers for the dynamics function, and the residual blocks in the representation functions have half the number of channels. Furthermore, we use a stack of only 4 past observations instead of 32. Additionally, some algorithmic refinements (such as those described in Appendix H of~\citep{schrittwieser2019mastering}) have not been implemented in the version that we use in this paper.

Our experimental results have been obtained using either 4 or 8 Tesla v100 GPUs for learning (compared to 8 third-generation Google Cloud TPUs~\citep{tpu} in the original MuZero paper, which are approximately equivalent to 64 v100 GPUs). Each learner GPU receives data from a separated, prioritized experience replay buffer~\citep{horgan2018distributed} storing the last $500000$ transitions. Each of these buffers is filled by 512 dedicated CPU actors\footnote{For 50 simulations per step; this number is scaled linearly as $12 + 10 \cdot N_\text{sim}$ to maintain a constant total number of frames per second when varying $N_\text{sim}$.}, each running a different environment instance. Finally, each actor receives updated parameters from the learner every 500 learner steps (corresponding to approximately 4 minutes of wall-clock time); because episodes can potentially last several minutes of wall-clock time, weights updating will usually occur within the duration of an episode. The total score at the end of an episode is associated to the version of the weights that were used to select the final action in the episode.

Hyperparameters choice generally follows those of \citep{schrittwieser2019mastering}, with the exception that we use the Adam optimizer with a constant learning rate of $0.001$.





\subsection{Details on discretizing continuous action space}\label{app:continuous_action_discretization}
AlphaZero \citep{silver2017mastering} is designed for discrete action spaces. When applying this algorithm to continuous control, we use the method described in~\citep{tang2019discretizing} to discretize the action space. Although the idea is simple, discretizing continuous action space has proved empirically efficient \citep{andrychowicz2020learning,tang2019discretizing}. We present the details below for completeness.

\paragraph{Discretizing the action space} We consider a continuous action space $\mathcal{A} = [-1,1]^m$ with $m$ dimensions. Each dimension is discretized into $K=5$ bins; specifically, the continuous action along each dimension is replaced by $K$ atomic categorical actions, evenly spaced between $[-1,1]$. This leads to a total of $K^m$ actions, which grows exponentially fast (e.g. $m=6$ leads to about $10^4$ joint actions). To avoid the curse of dimensionality, we assume that the parameterized policy can be factorized as $\pi_\theta(\mathbf{a}|x) = \Pi_{i=1}^m \pi_\theta^{(i)}(a_i|x)$, where $\pi_\theta^{(i)}(a_i|x)$ is the marginal distribution for dimension $i$, $a_i \in \{1,2...K\}$ is the discrete action along dimension $i$ and $\mathbf{a} = [a_1,a_2...a_m]$ is the joint action.

\paragraph{Modification to the search procedure} Though it is convenient to assume a factorized form of the parameterized policy \citep{andrychowicz2020learning,tang2019discretizing}, it is not as straightforward to apply the same factorization assumption to the Q-function $Q(x,\mathbf{a})$. A most naive way of applying the search procedure is to maintain a Q-table of size $K^m$ with one entry for each joint action, which may not be tractable in practice. Instead, we maintain $m$ separate Q-tables each with $K$ entries $Q_i(x,a_i)$. We also maintain $m$ count tables $n(x,a_i)$ with $K$ entries for each dimension.

To make the presentation clear, we detail on how the search is applied. At each node of the search tree, we maintain $m$ tables each with $K$ entries as introduced above. The three core components of the tree search are modified as follows.
\begin{itemize}
    \item \textbf{Selection.} During forward action selection, the algorithm needs to select an action $\mathbf{a}$ at node $x$. This joint action $\mathbf{a}$ has all its components $a_i$ selected independently, using the action selection formula applied to each dimension. To select action at dimension $i$, we need the Q-table $Q_i(x,a_i)$, the prior $\pi_\theta^{(i)}(a_i|x)$ and count $n(x,a_i)$ for dimension $i$.
    \item \textbf{Expansion.} The expansion part does not change.
    \item \textbf{Back-up.} During the value back-up, we update Q-tables of each dimension independently. At a node $x$, given the downstream reward $R(x,a)$ and child value $V(\text{child}(x,\mathbf{a}))$, we generate the target update for each Q-table and count table as $Q(x,a_i) \leftarrow R(x,a) + \gamma V(\text{child}(x,\mathbf{a}))$ and $n(x,a_i)\leftarrow n(x,a_i) + 1$.
\end{itemize}

The $m$ small Q-tables can be interpreted as maintaining the marginalized values of the joint Q-table. Indeed, let us denote by $Q(x,\mathbf{a})$ the joint Q-table with $K^m$ entries. At dimension $i$, the Q-table $Q(x,a_i)$ increments its values purely based on the choice of $a_i$, regardless of actions in other dimension $a_j,j\neq i$. This implies that the Q-table $Q(x,a_i)$ marginalizes the joint Q-table $Q(x,\mathbf{a})$ via the visit count distribution.

\paragraph{Details on the learning} At the end of the tree search, a distribution target $\mathbf{\hat{\pi}}$ or $\mathbf{\bar{\pi}}$ is computed from the root node. In the discretized case, each component of the target distribution is computed independently. For example, $\mathbf{\hat{\pi}}_i$ is computed from $N(x_0,a_i)$. The target distribution derived from constrained optimization $\mathbf{\bar{\pi}}_i$ is also computed independently across dimensions, from $Q(x_0,a_i)$ and $N(x_0,a_i)$. In general, let $\mathbf{\pi}_\text{target}(\cdot|x)$ be the target distribution and $\mathbf{\pi}^{(i)}_\text{target}(\cdot|x)$ its marginal for dimension $i$. Due to the factorized assumption on the policy distribution, the update can be carried out independently for each dimension. Indeed,
$\KL[\mathbf{\pi}_\text{target}(\cdot|x), \pi_\theta(\cdot|x)] = \sum_{i=1}^m \KL[\mathbf{\pi}^{(i)}_\text{target}(\cdot|x),\mathbf{\pi}^{(i)}_\theta(\cdot|x)]$, sums over dimensions.

\subsection{Practical computation of $\pibar$}\label{app:compute_pibar}

The vector $\pibar$ is defined as the solution to a multi-dimensional optimization problem; however, we show that it can be computed easily by dichotomic search. We first restate the definition of $\pibar$,
\begin{align}
\pibar \eqdef \argmax_{\mathbf{y} \in \mathcal{S}}  \brkt{\mathbf{q}^T \mathbf{y} - \lambda_N \KL\brkt{\pi_\theta, \mathbf{y}}}. \tag{\ref{eq:pibar}}
\end{align}

Let us define
\begin{align}
    \forall a\in\mathcal{A}\quad\pi_\alpha[a] \eqdef \lambda_N \frac{\pi_\theta[a]}{\alpha - \mathbf{q}[a]}\quad\text{and}\quad\alpha^\ast \eqdef \max\brc{\alpha \in \Real \;\text{s.t}\; \sum_b \pi_\alpha[b] = 1}\cdot
\end{align}
\begin{proposition}{}\label{prop:compute_pibar}
\begin{align}
    (i) &\hspace{1em} \pi_{\alpha^\ast} = \pibar\\
    (ii) &\hspace{1em} \alpha^\ast \ge \alpha_\text{min} \eqdef \max_{b\in\mathcal{A}} \prth{q[b] + \lambda_N\cdot\pi_\theta[b]}\\
    (iii) &\hspace{1em}\alpha^\ast \le \alpha_\text{max} \eqdef \max_{b\in\mathcal{A}} q[b] + \lambda_N\\
\end{align}
\end{proposition}

As $\sum_b \pibar_\alpha[b]$ is strictly decreasing on $\alpha \in\prth{\alpha_\text{min}, \alpha_\text{max}}$, \Cref{prop:compute_pibar} guarantees that $\pibar$ can be computed easily using dichotomic search over $\prth{\alpha_\text{min}, \alpha_\text{max}}$.

\paragraph{Proof of (i). }
\begin{proof}
The proof start the same as the one of \Cref{lemma:argmax} of \Cref{sec:proof_action_selection_pibar} setting $f(x) = -\log(x)$ to get
\begin{align}
\exists \alpha \quad q + \lambda_N \cdot \frac{\pi_\theta}{\pibar} = \alpha \mathbbm{1},
\end{align}
with $\mathbbm{1}$ being the the vector such that $\forall a\;\mathbbm{1}_a = 1$. Therefore there exists $\alpha\in\Real$ such that 
\begin{align}
    \pibar = \frac{\lambda_N\cdot\pi_\theta}{\alpha - q}
\end{align}
Then $\alpha$ is set such that $\sum_b \pibar_b = 1$ and $\forall b\;\pibar_b\ge 0$. 
\end{proof}

\paragraph{Proof of (ii). }
\begin{proof}
\begin{align}
    \forall a\quad 1 \ge \pibar[a] = \frac{\lambda_N\cdot\pi_\theta[a]}{\alpha - q[a]} \implies \forall a\quad \alpha \ge q[a] + \lambda_N\cdot\pi_\theta[a]
\end{align}
\end{proof}

\paragraph{Proof of (iii). }
\begin{proof}
\begin{align}
    \sum_b \pi_{\alpha_\text{max}}[b] &= \sum_b \frac{\lambda_N\cdot\pi_\theta[b]}{\max_{c\in\mathcal{A}} q[c] + \lambda_N - q[b]}
    \le \sum_b \frac{\lambda_N\cdot\pi_\theta[b]}{\lambda_N} = 1
\end{align}
We combine this with the fact that $\sum_b \pi_{\alpha}[b]$ is a decreasing function of $\alpha$ for any $\alpha > \max_b q[b]$, and $\sum_b \pi_{\alpha^\ast}[b] = 1$.  
\end{proof}

\section{Additional experimental results}
\subsection{Complements to \Cref{sec:mz_vs_pz_ms_pacman}}\label{app:mz_vs_pz_ms_pacman}
\begin{figure}
\begin{subfigure}{0.32\columnwidth}
\includegraphics[width=\linewidth]{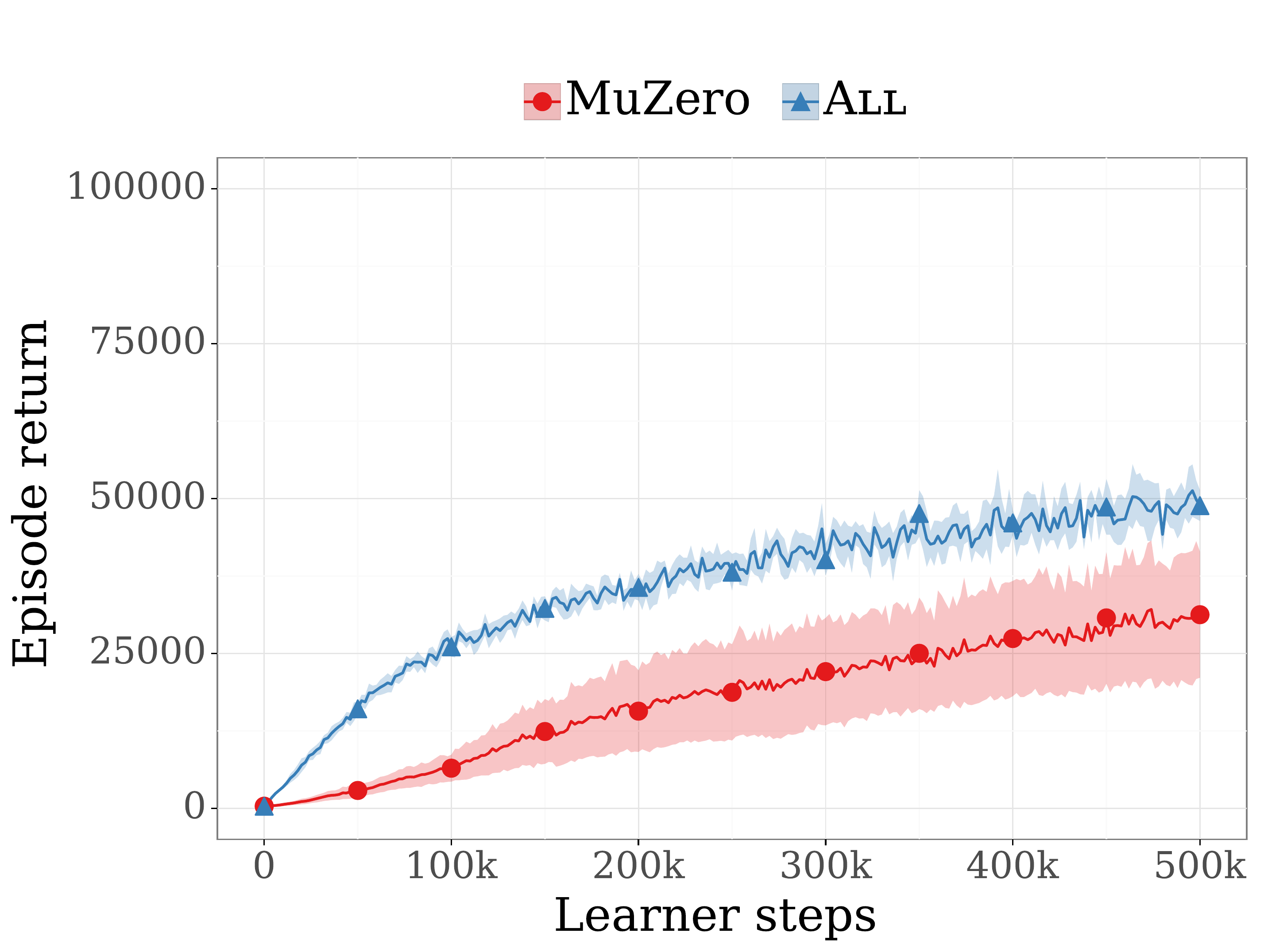}\caption{5 simulations}
\end{subfigure}
\begin{subfigure}{0.32\columnwidth}
\includegraphics[width=\linewidth]{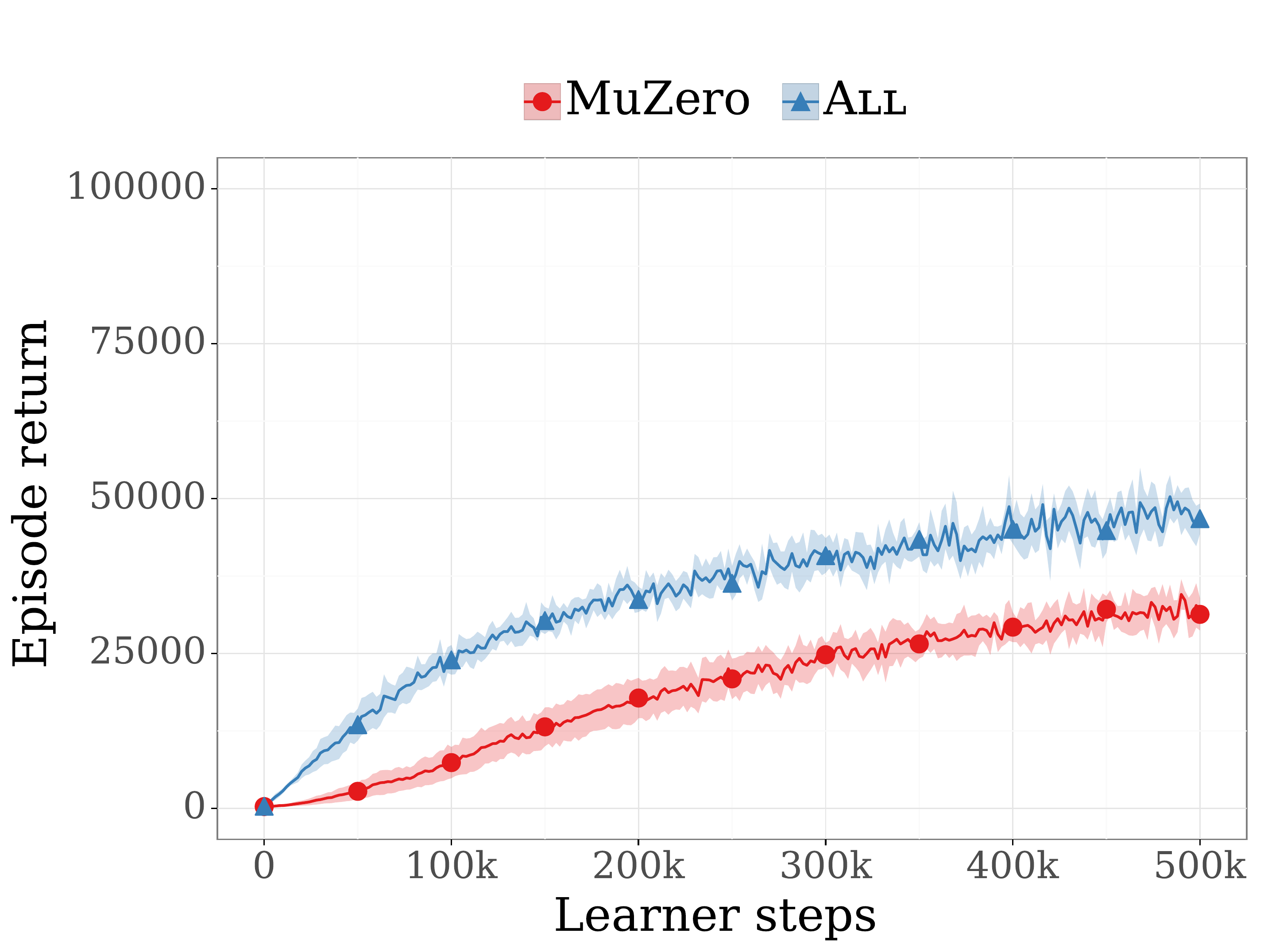}\caption{6 simulations}
\end{subfigure}
\begin{subfigure}{0.32\columnwidth}
\includegraphics[width=\linewidth]{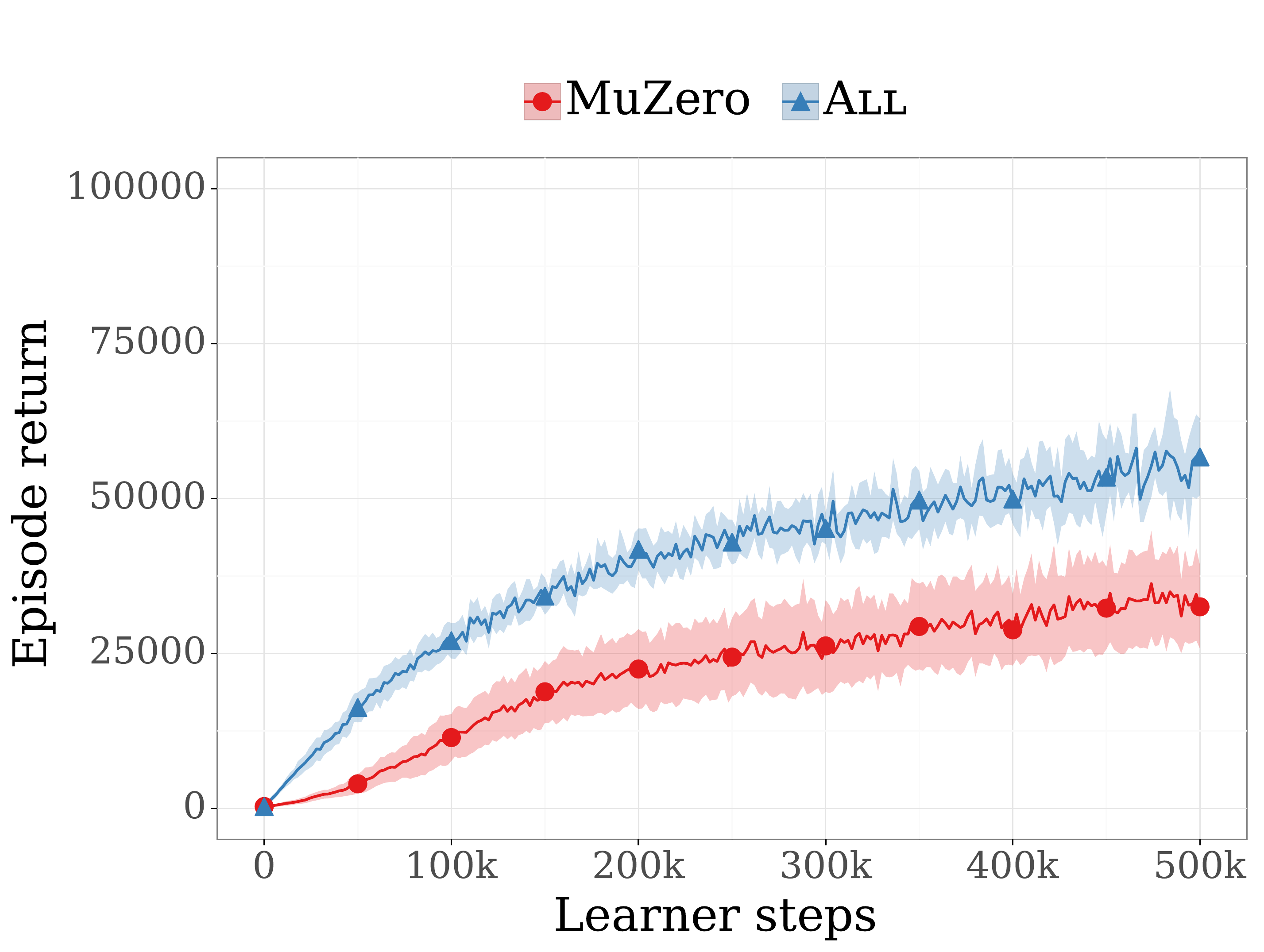}\caption{8 simulations}
\end{subfigure}

\begin{subfigure}{0.32\columnwidth}
\includegraphics[width=\linewidth]{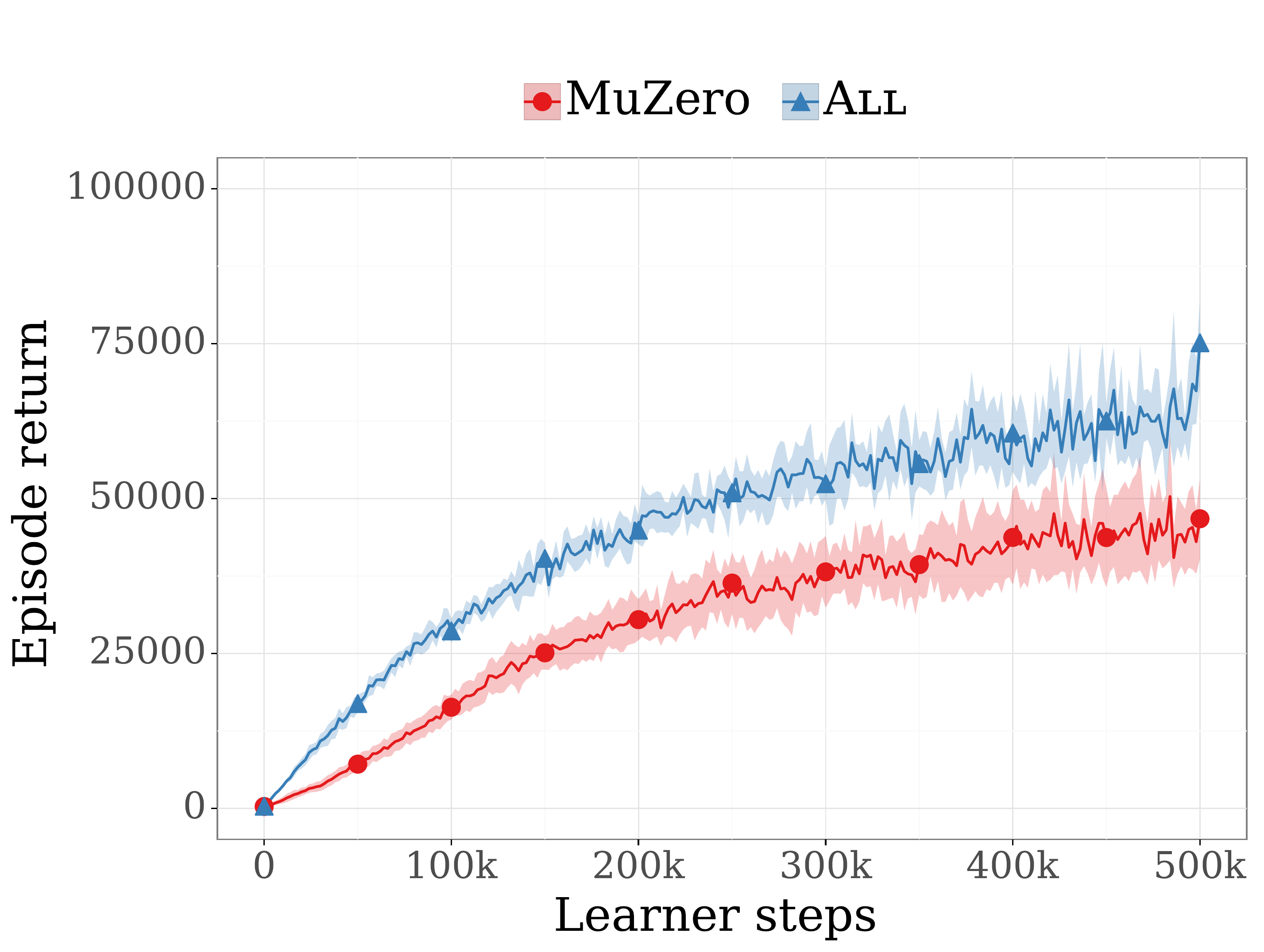} \caption{12 simulations}
\end{subfigure}
\begin{subfigure}{0.32\columnwidth}
\includegraphics[width=\linewidth]{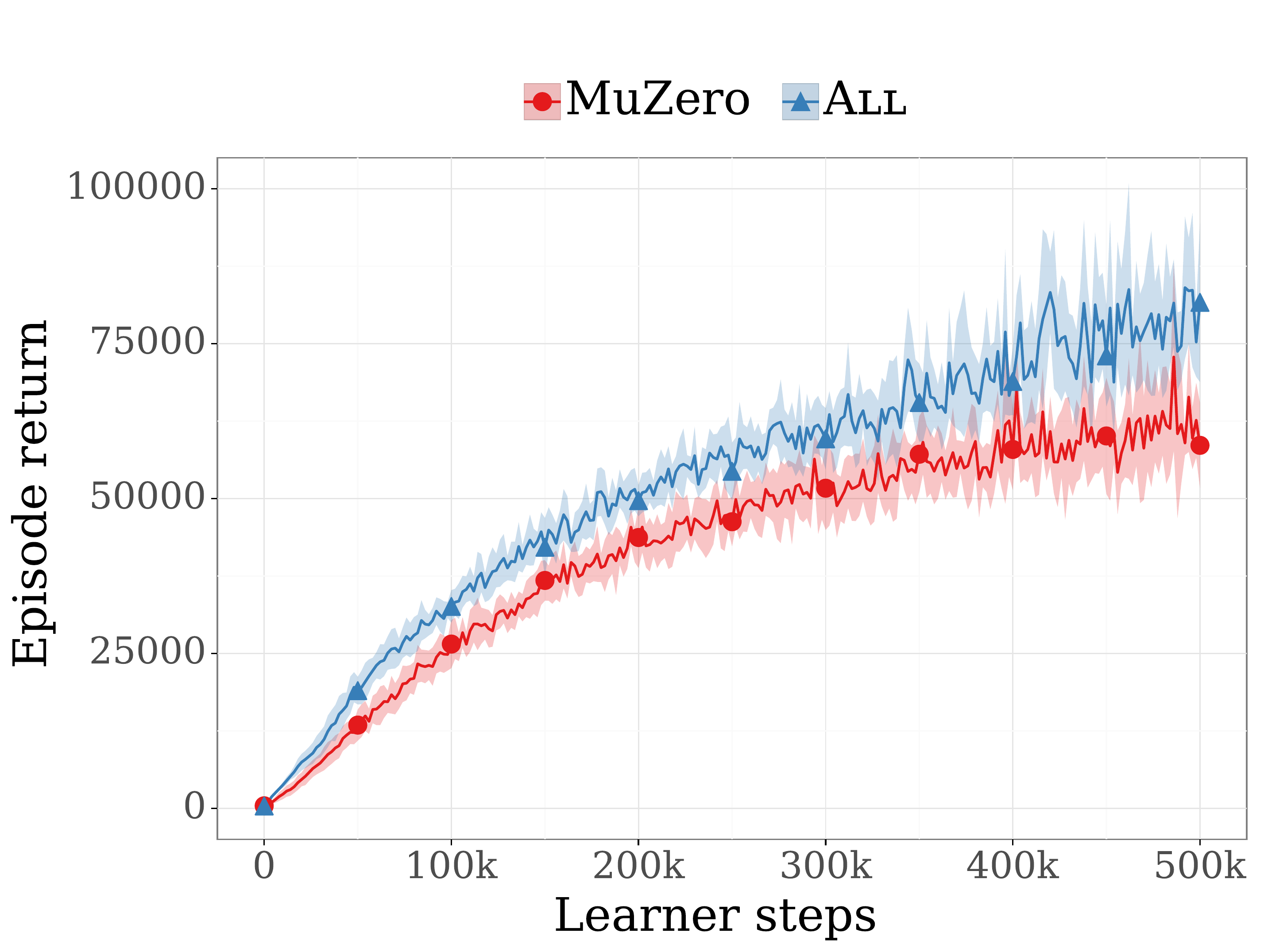}\caption{24 simulations}
\end{subfigure}
\begin{subfigure}{0.32\columnwidth}
\includegraphics[width=\linewidth]{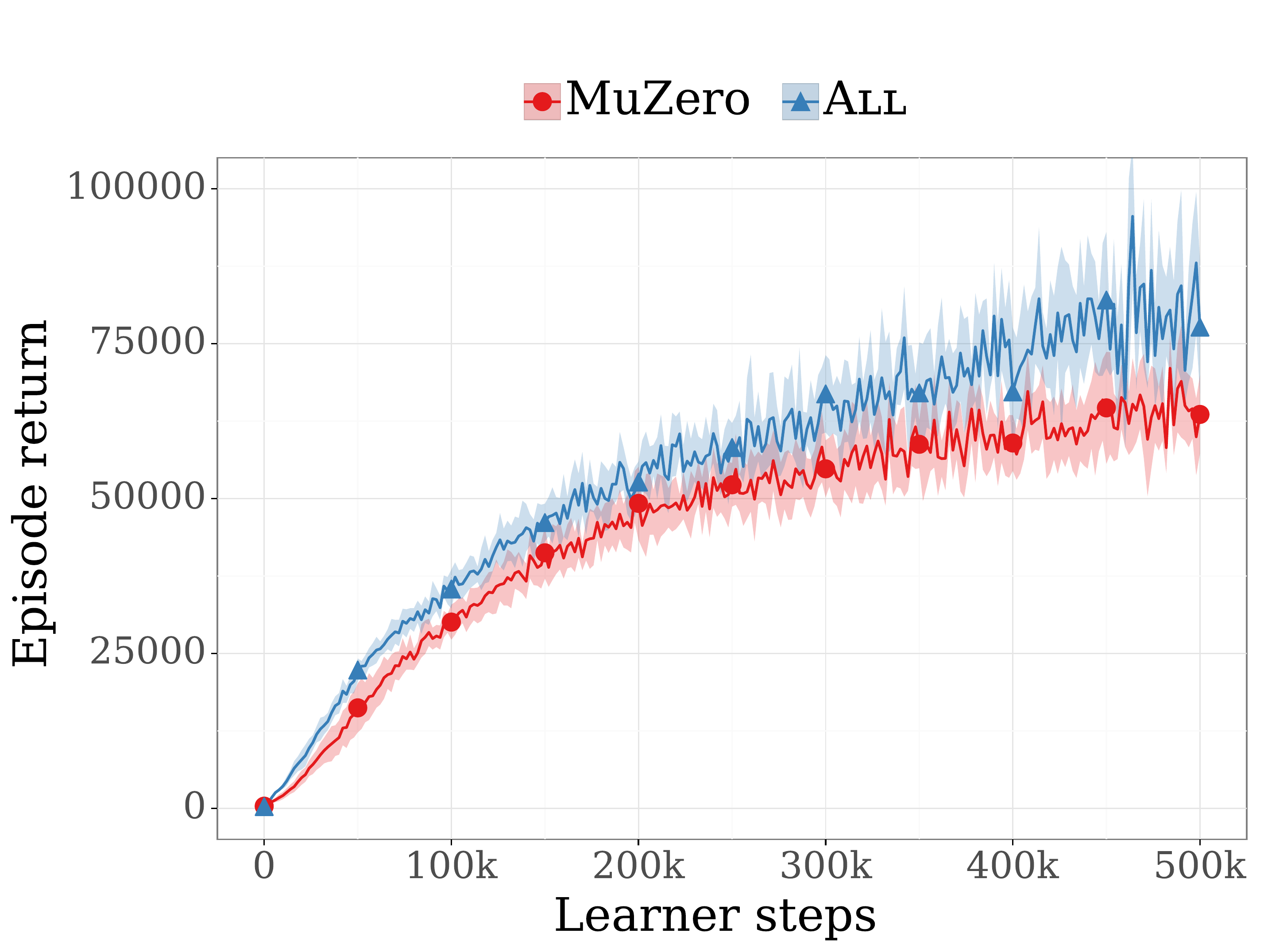}\caption{50 simulations}
\end{subfigure}
\caption{\label{fig:mz_vs_pz_ms_pacman}Dispersion between seeds at different number of simulations per step on Ms Pacman.}
\end{figure}

\Cref{fig:mz_vs_pz_ms_pacman} presents a comparison of the score obtained by our MuZero implementation and the proposed \algoall{} variant at different simulation budgets on the Ms. Pacman level; the results from \Cref{fig:mz_vs_pz_5_50_sims} are also included fore completeness. In this experiment, we used 8 seeds with 8 GPUs per seed and a batch size of 256 per GPU. We use the same set of hyper-parameters for MuZero and \algoall{}; these parameters were tuned on MuZero. The solid line corresponds to the average score (solid line) and the 95\% confidence interval (shaded area) over the 8 seeds, averaged for each seed over buckets of 2000 learner steps without additional smoothing. Interestingly, we observe that \algoall{} provides improved performance at low simulation budgets while also reducing the dispersion between seeds.

\begin{figure}
\begin{subfigure}{0.32\columnwidth}
\includegraphics[width=\linewidth]{plots/sixgames/sixgames_alien_50_sims.pdf}\caption{Alien}
\end{subfigure}
\begin{subfigure}{0.32\columnwidth}
\includegraphics[width=\linewidth]{plots/sixgames/sixgames_asteroids_50_sims.pdf}\caption{Asteroids}
\end{subfigure}
\begin{subfigure}{0.32\columnwidth}
\includegraphics[width=\linewidth]{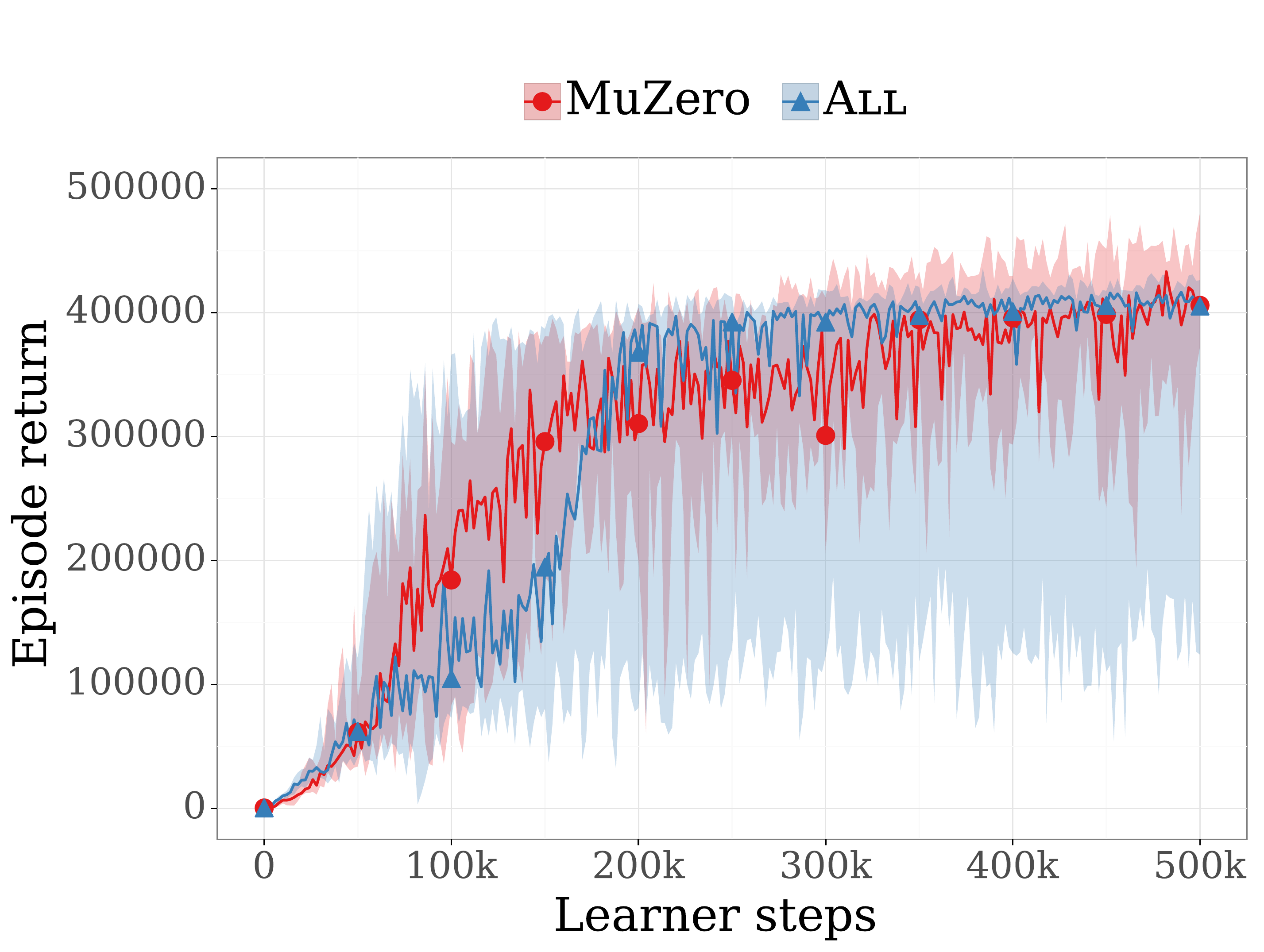}\caption{Beam rider}
\end{subfigure}

\begin{subfigure}{0.32\columnwidth}
\includegraphics[width=\linewidth]{plots/sixgames/sixgames_gravitar_50_sims.pdf}\caption{Gravitar}
\end{subfigure}
\begin{subfigure}{0.32\columnwidth}
\includegraphics[width=\linewidth]{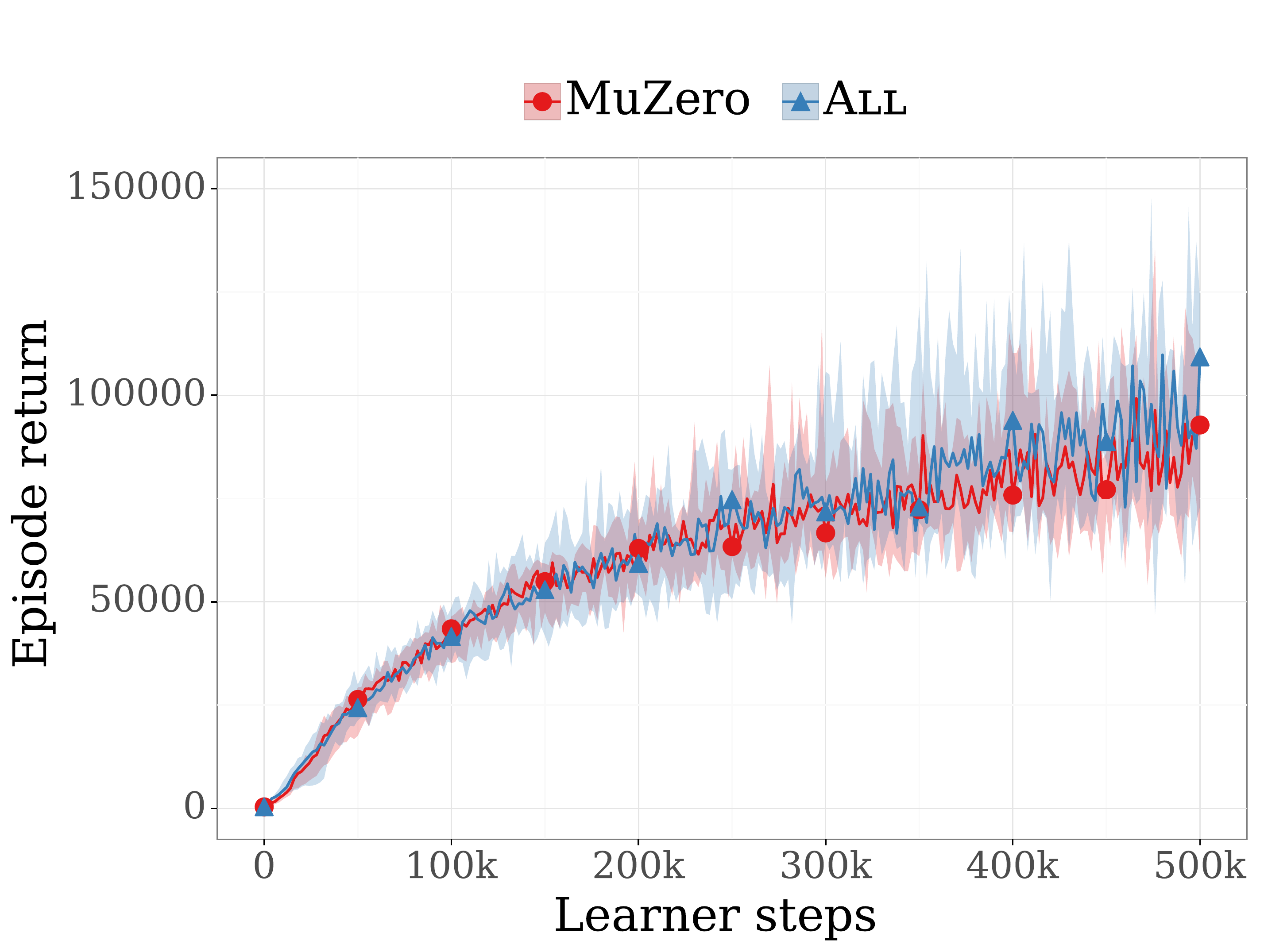}\caption{Ms Pacman}
\end{subfigure}
\begin{subfigure}{0.32\columnwidth}
\includegraphics[width=\linewidth]{plots/sixgames/sixgames_seaquest_50_sims.pdf}\caption{Seaquest}
\end{subfigure}
\caption{\label{fig:sixgames}Comparison of median score over 6 seeds of MuZero and \algoall{} on six Atari games with 50 simulations per step. The shaded area correspond the the best and worst seeds. }
\end{figure}

\Cref{fig:sixgames} presents a comparison of the score obtained by our MuZero implementation and the proposed \algoall{} variant on six Atari games, using 6 seeds per game and a batch size of 512 per GPU and 8 GPUs; we use the same set of hyper-parameters as in the other experiments. Because the distribution of scores across seeds is skewed towards higher values, we represent dispersion between seeds using the min-max interval over the 6 seeds (shaded area) instead of using the standard deviation; the solid line represents the median score over the seeds.

\subsection{Complements to \Cref{sec:mz_vs_pz_cc}}\label{app:mz_vs_pz_cc}

\paragraph{Details on the environments} The DeepMind Control Suite environments \citep{tassa2018control} are control tasks with continuous action space $\mathcal{A} = [-1,1]^m$. These tasks all involve simulated robotic systems and the reward functions are designed so as to guide the system for accomplish e.g. locomotion tasks. Typically, these robotic systems have relatively low-dimensional sensory recordings which summarize the environment states. To make the tasks more challenging, for observations, we take the third-person camera of the robotic system and use the image recordings as observations to the RL agent. These images are of dimension $64 \times 64 \times 3$.

\begin{figure}\centering
\begin{subfigure}{0.32\columnwidth}
\includegraphics[width=\linewidth]{plots/control/mz_vs_po_cheetah_run_asymptotic.pdf}\caption{Cheetah Run}
\end{subfigure}
\begin{subfigure}{0.32\columnwidth}
\includegraphics[width=\linewidth]{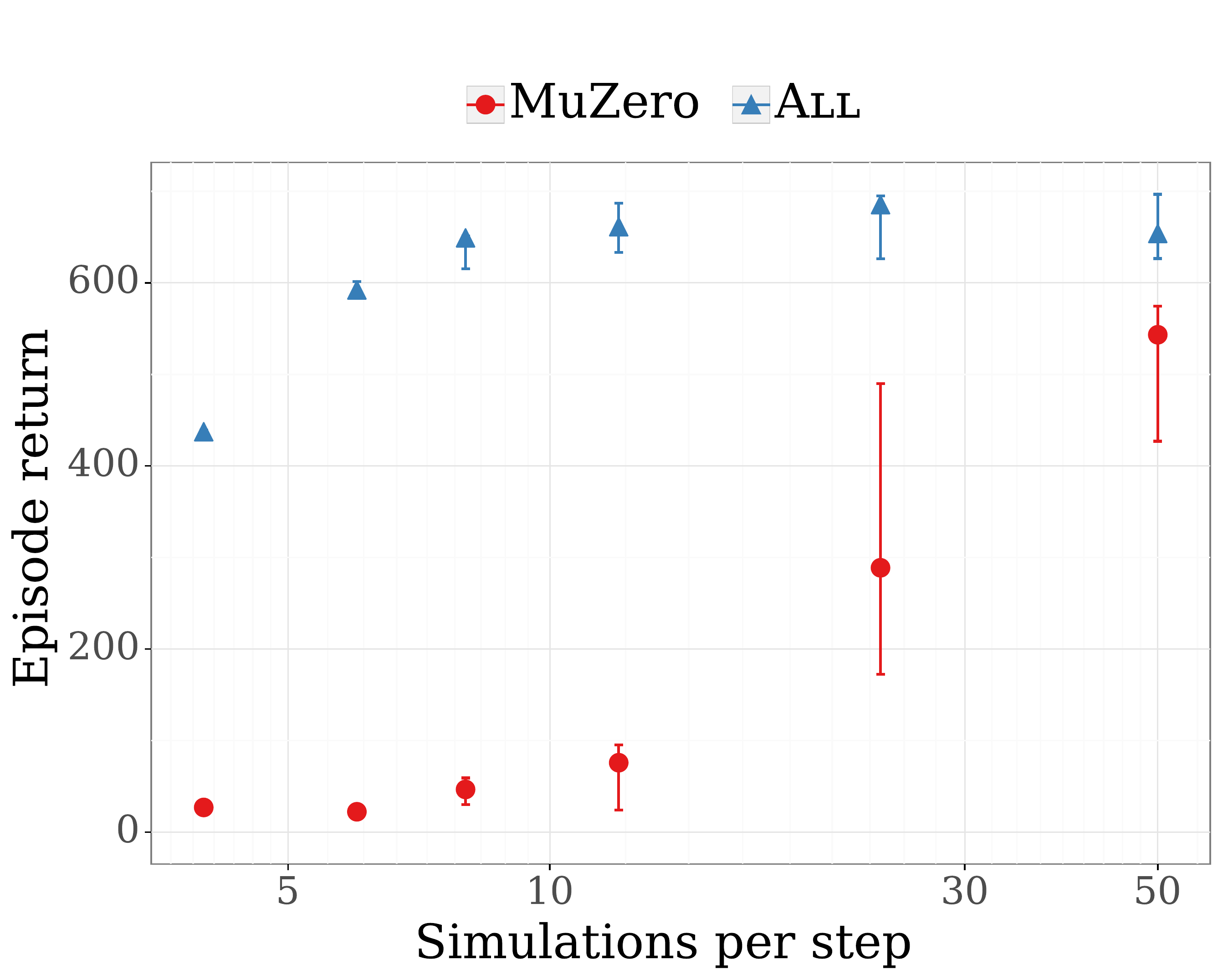}\caption{Walker Run}
\end{subfigure}

\begin{subfigure}{0.32\columnwidth}
\includegraphics[width=\linewidth]{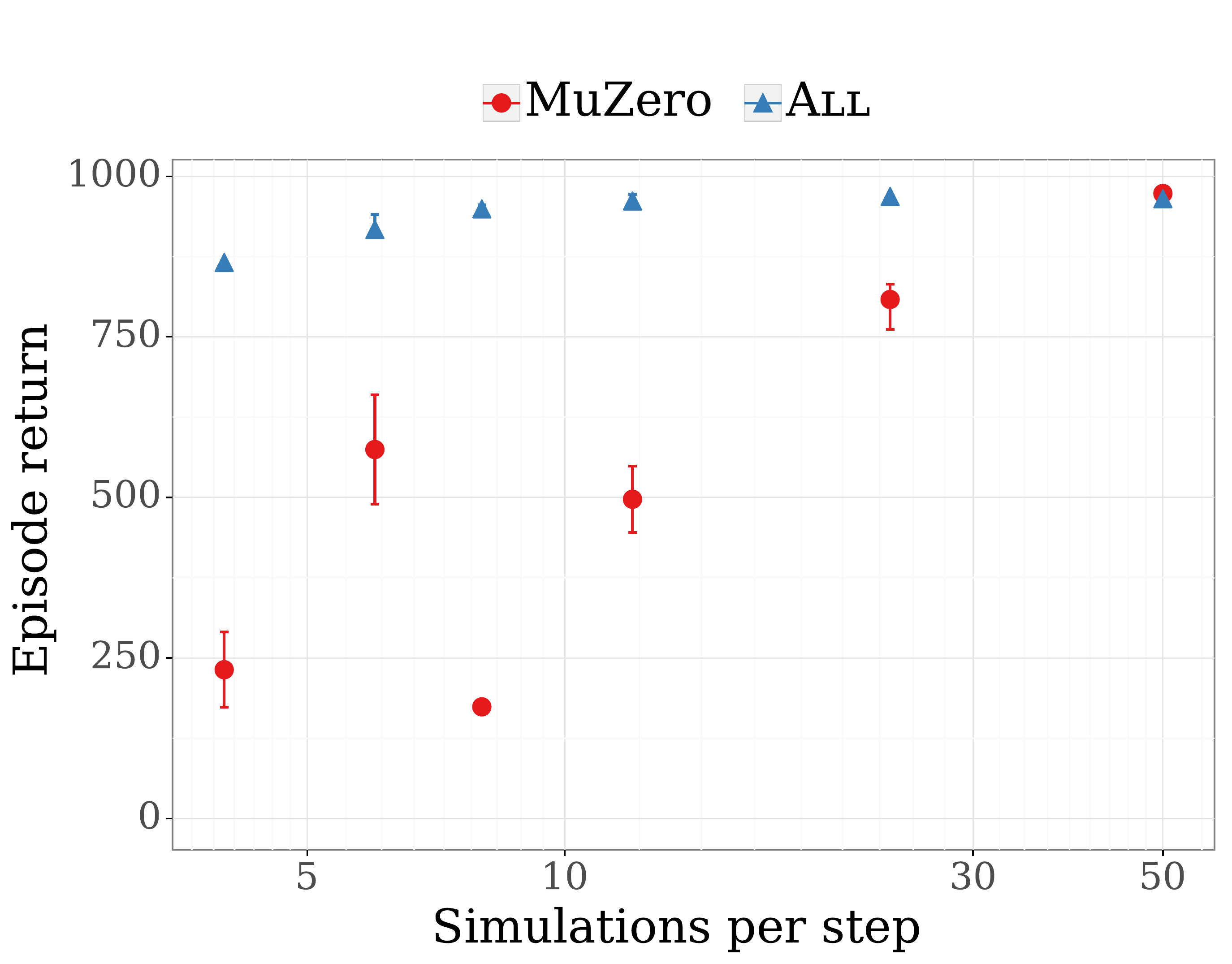}\caption{Walker Stand}
\end{subfigure}
\begin{subfigure}{0.32\columnwidth}
\includegraphics[width=\linewidth]{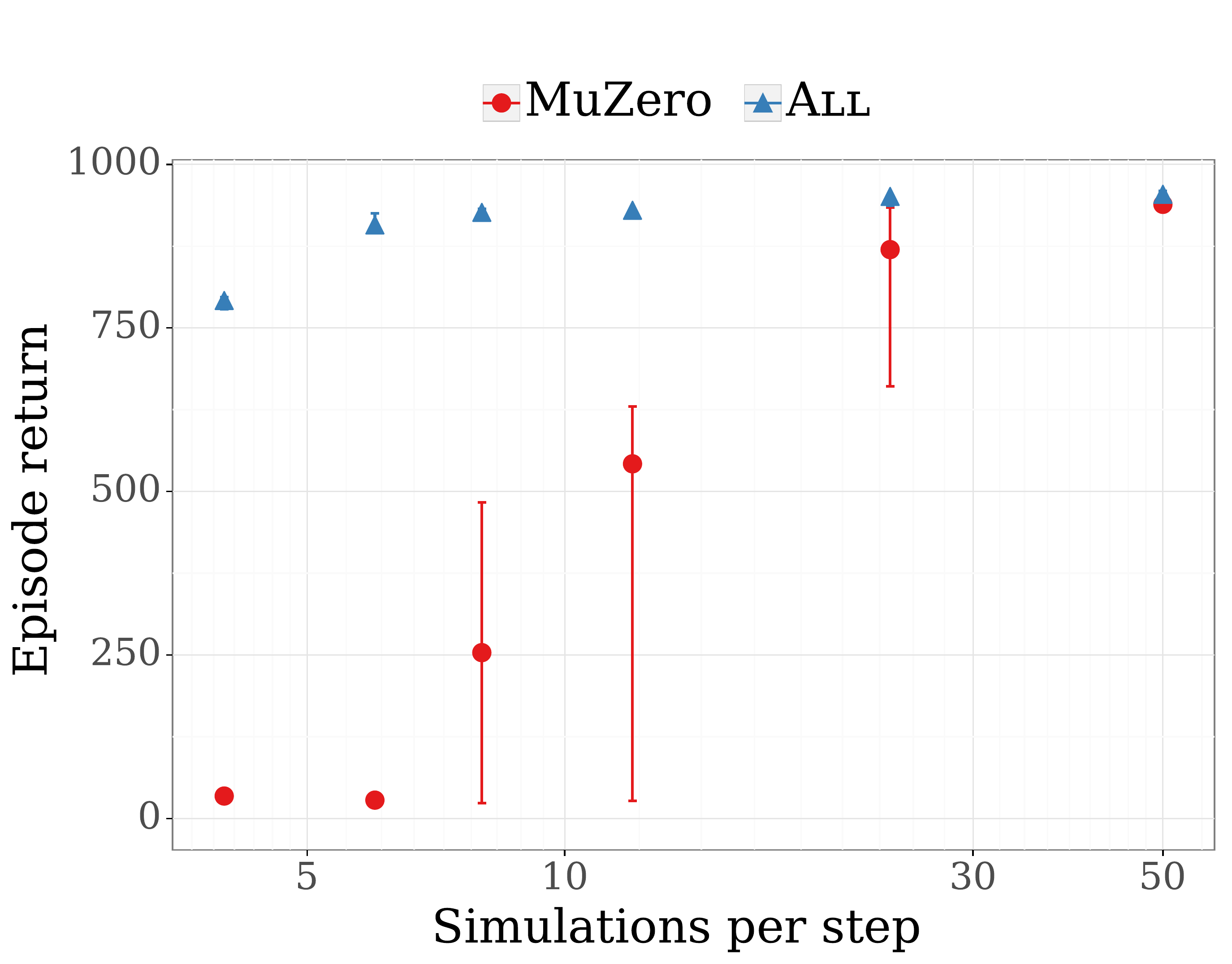}\caption{Walker Walk}
\end{subfigure}
\caption{\label{fig:mz_vs_pz_with_nsim}Score of MuZero and \algoall{} on Continuous control tasks after 100k learner steps as a function of the number of simulations $N_\text{sim}$.}
\end{figure}

\begin{figure}
\begin{subfigure}{0.32\columnwidth}
\includegraphics[width=\linewidth]{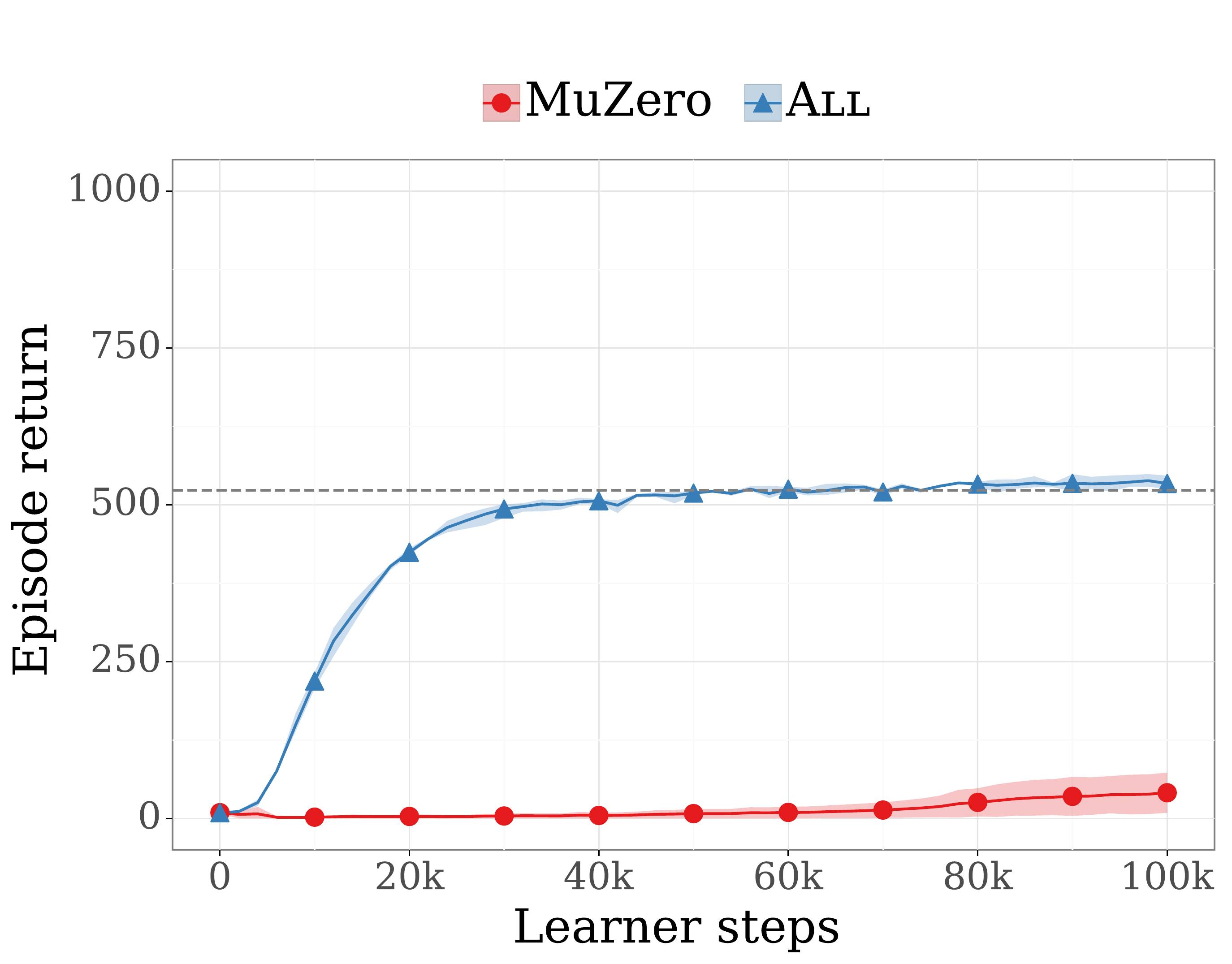}\caption{4 simulations}
\end{subfigure}
\begin{subfigure}{0.32\columnwidth}
\includegraphics[width=\linewidth]{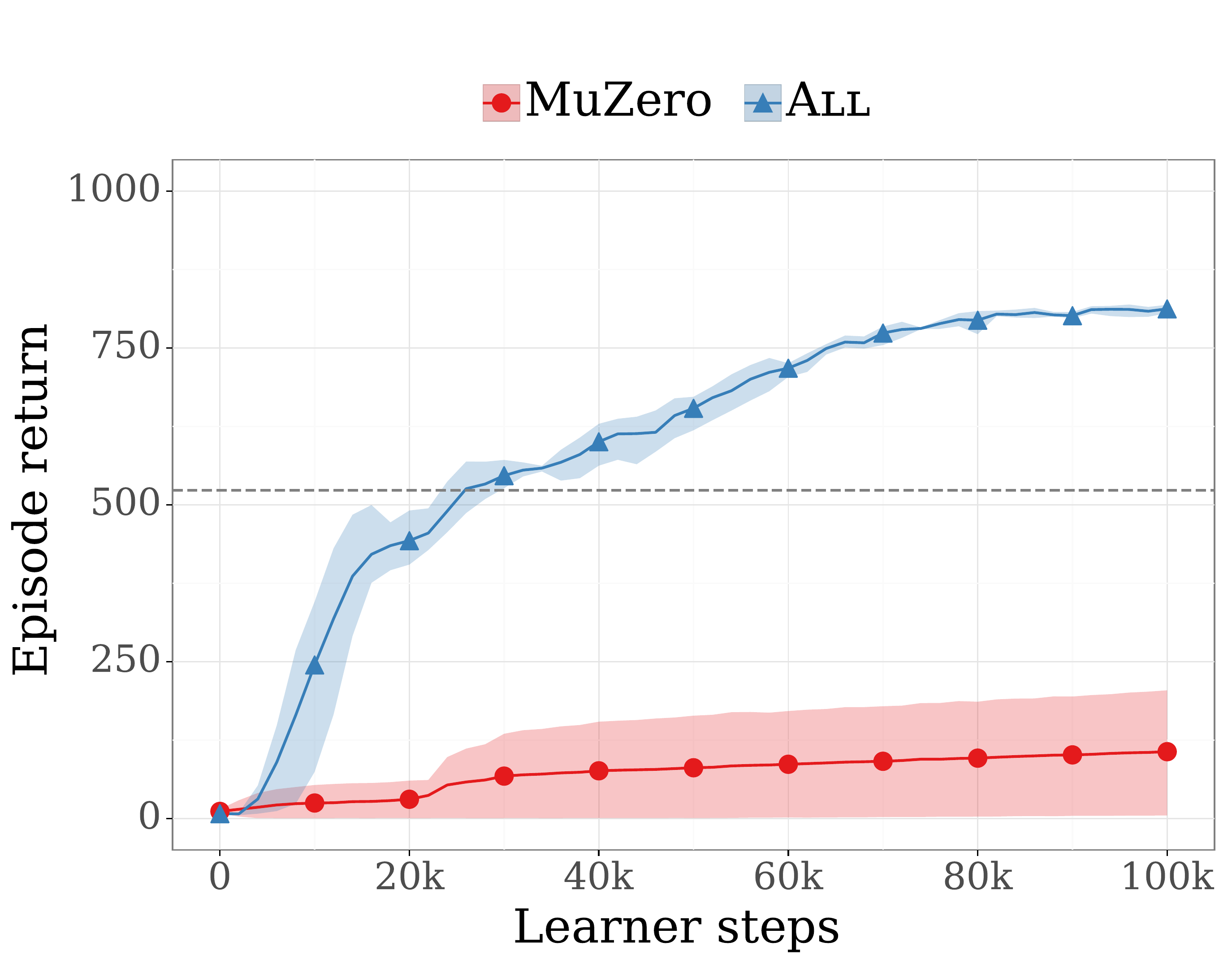}\caption{6 simulations}
\end{subfigure}
\begin{subfigure}{0.32\columnwidth}
\includegraphics[width=\linewidth]{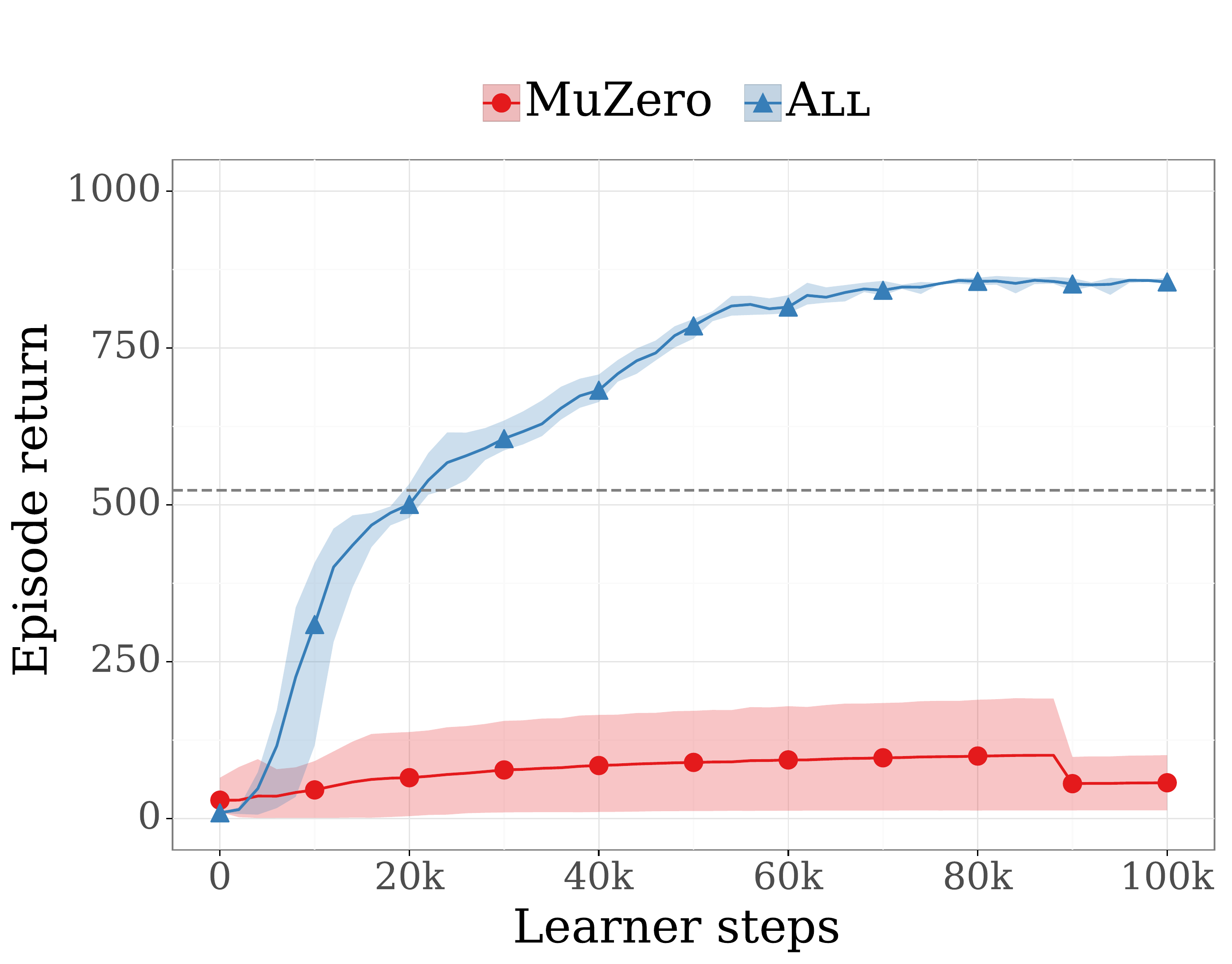}\caption{8 simulations}
\end{subfigure}

\begin{subfigure}{0.32\columnwidth}
\includegraphics[width=\linewidth]{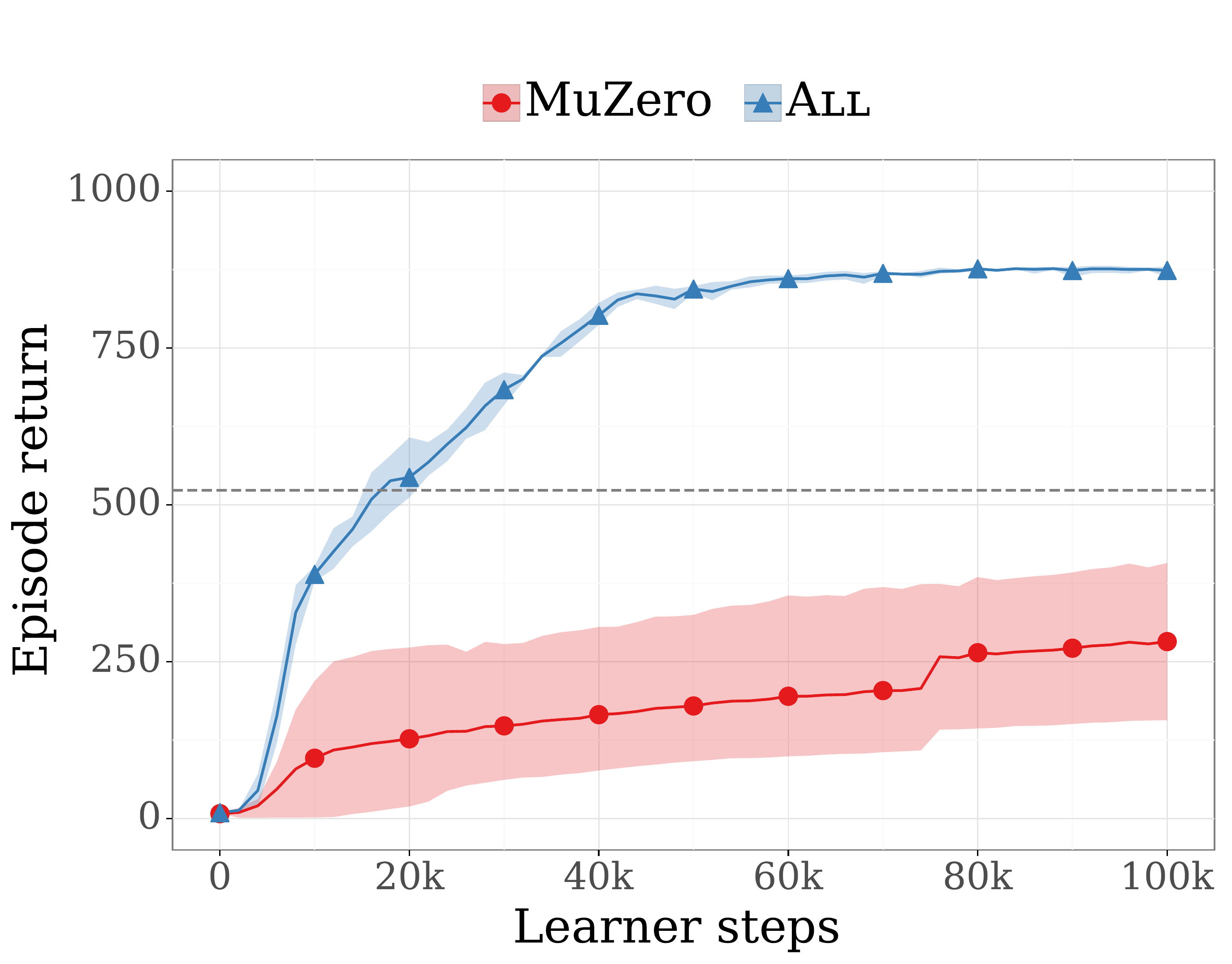}\caption{12 simulations}
\end{subfigure}
\begin{subfigure}{0.32\columnwidth}
\includegraphics[width=\linewidth]{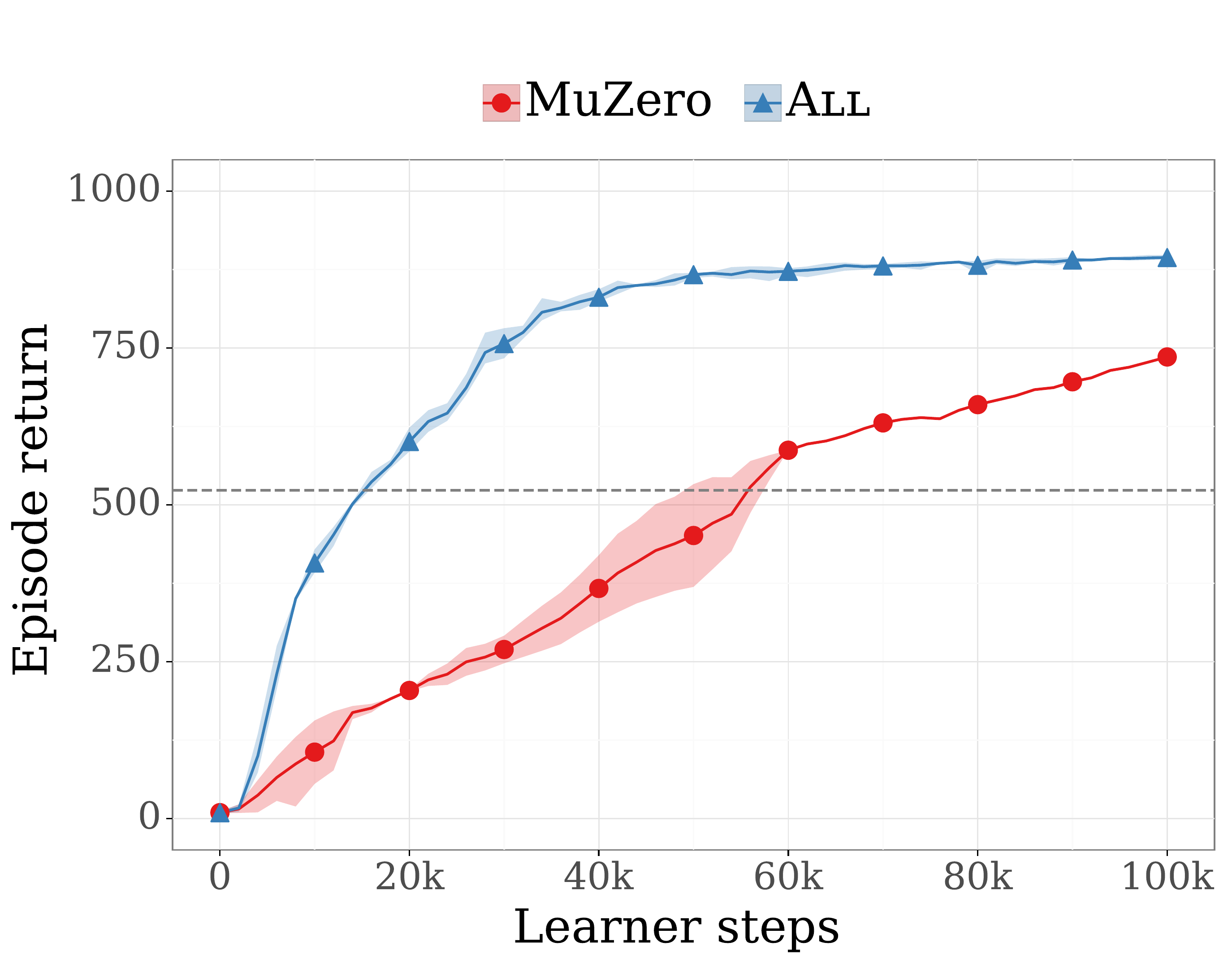}\caption{24 simulations}
\end{subfigure}
\begin{subfigure}{0.32\columnwidth}
\includegraphics[width=\linewidth]{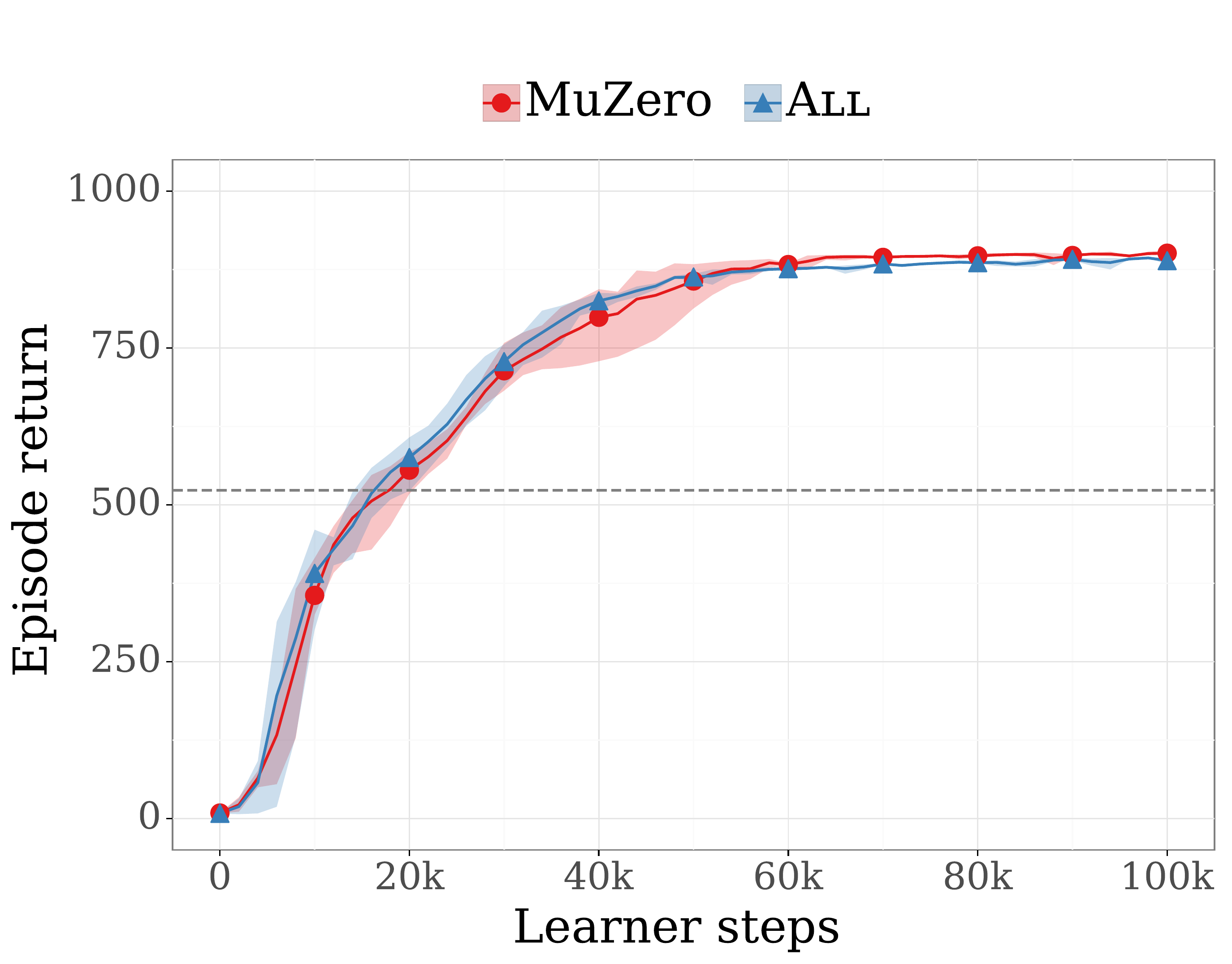}\caption{50 simulations}
\end{subfigure}
\caption{\label{fig:mz_vs_pz_cc_cheetah}Comparison of MuZero and \algoall{} on Cheetah Run.}
\end{figure}

\begin{figure}
\begin{subfigure}{0.32\columnwidth}
\includegraphics[width=\linewidth]{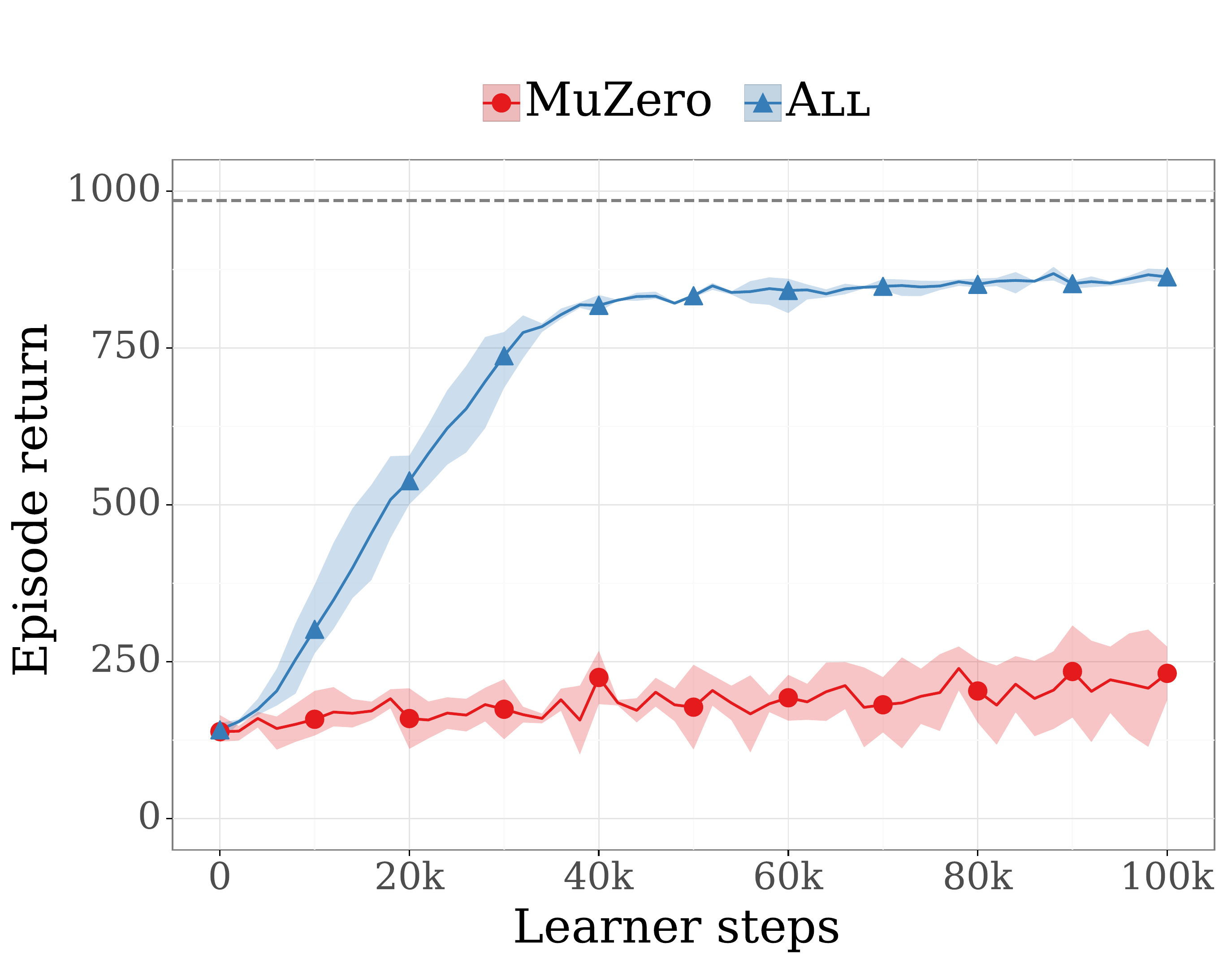}\caption{4 simulations}
\end{subfigure}
\begin{subfigure}{0.32\columnwidth}
\includegraphics[width=\linewidth]{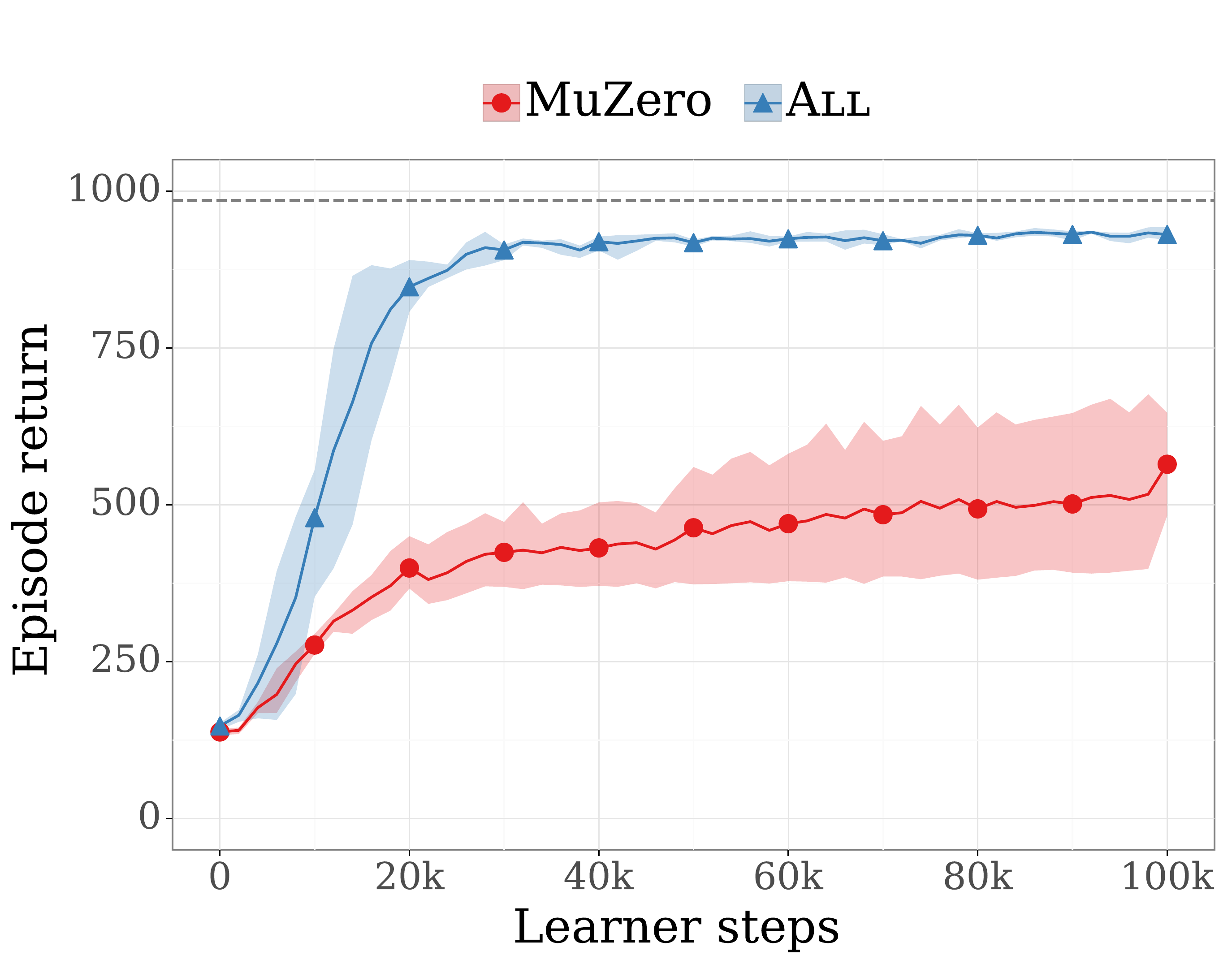}\caption{6 simulations}
\end{subfigure}
\begin{subfigure}{0.32\columnwidth}
\includegraphics[width=\linewidth]{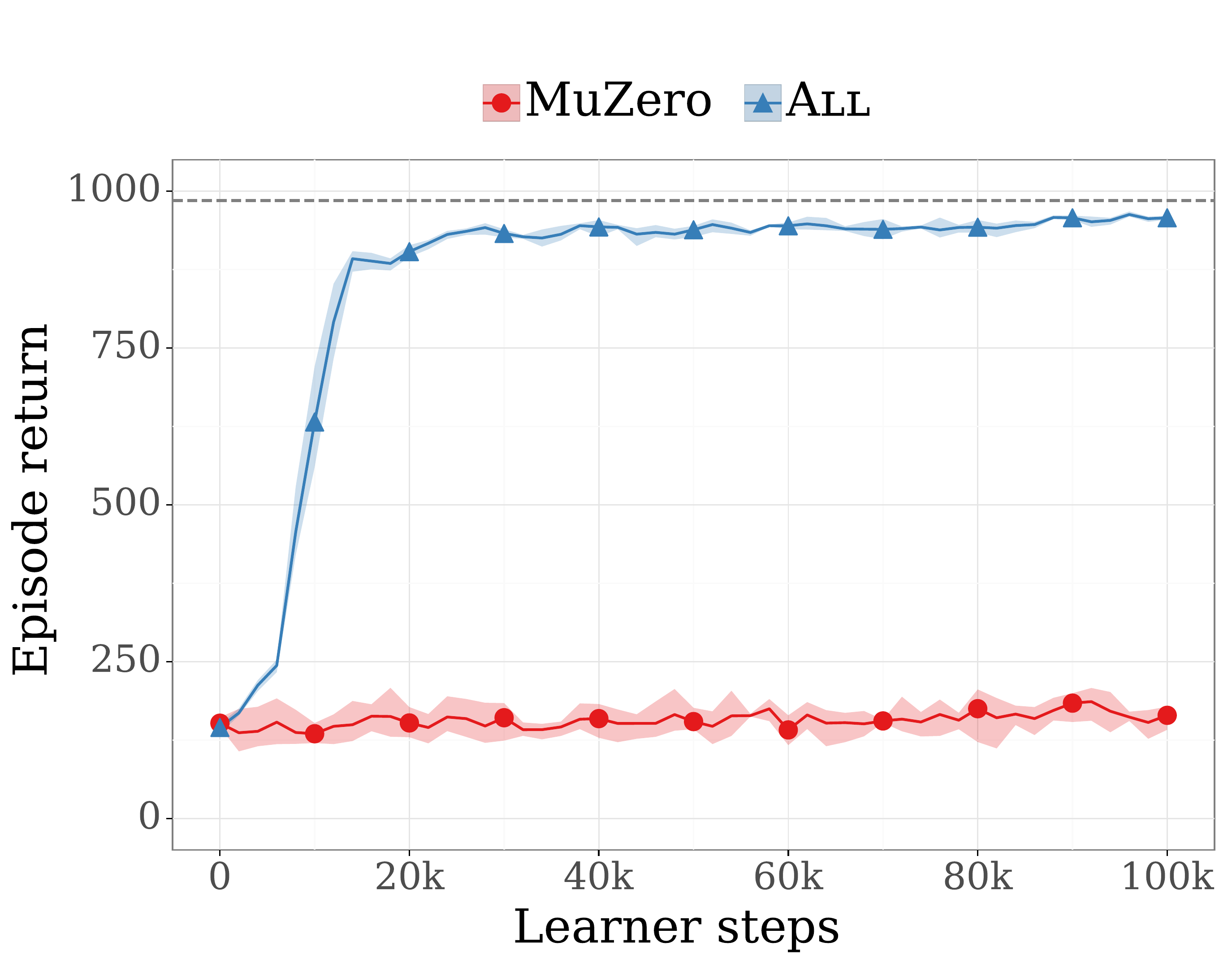}\caption{8 simulations}
\end{subfigure}

\begin{subfigure}{0.32\columnwidth}
\includegraphics[width=\linewidth]{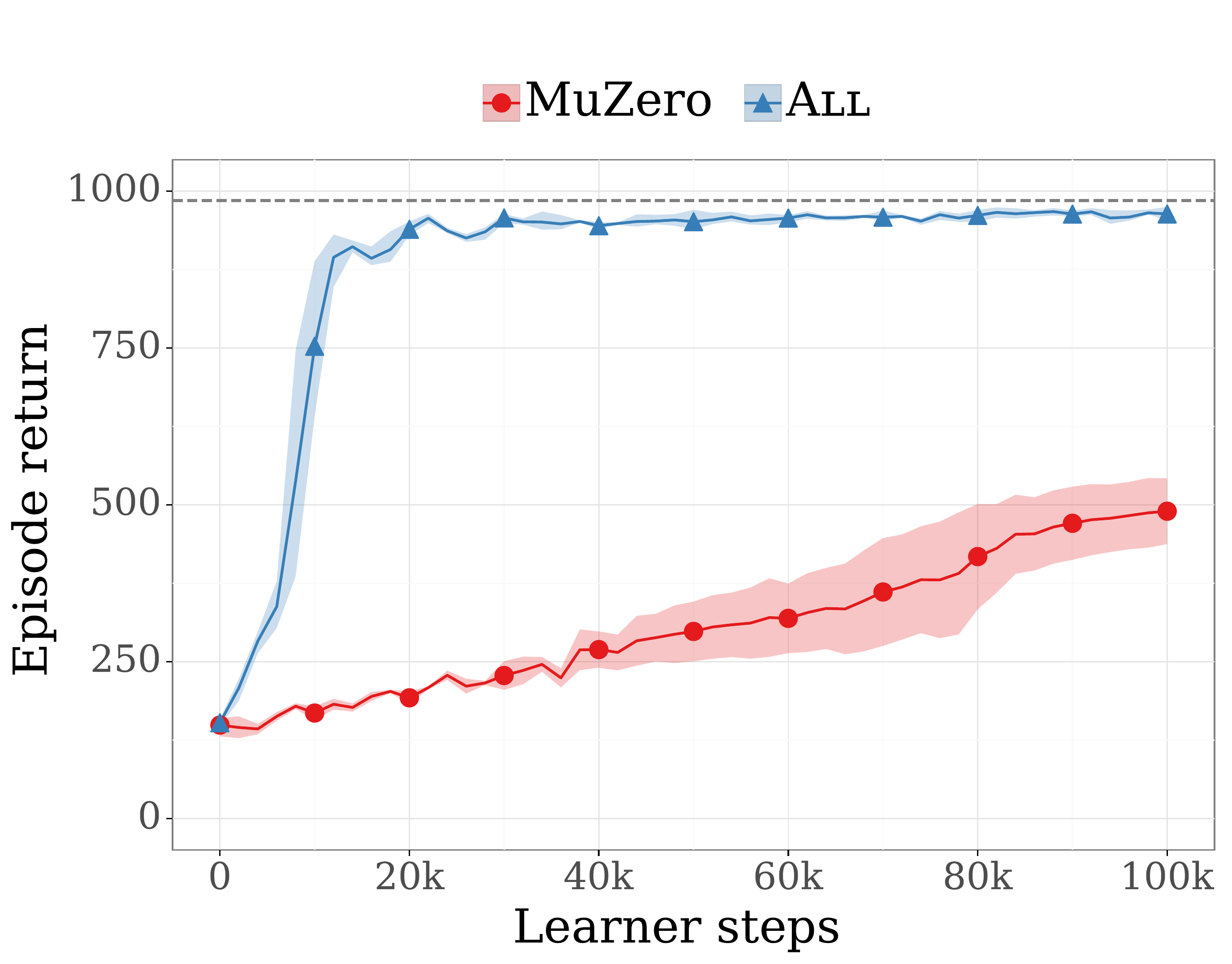}\caption{12 simulations}
\end{subfigure}
\begin{subfigure}{0.32\columnwidth}
\includegraphics[width=\linewidth]{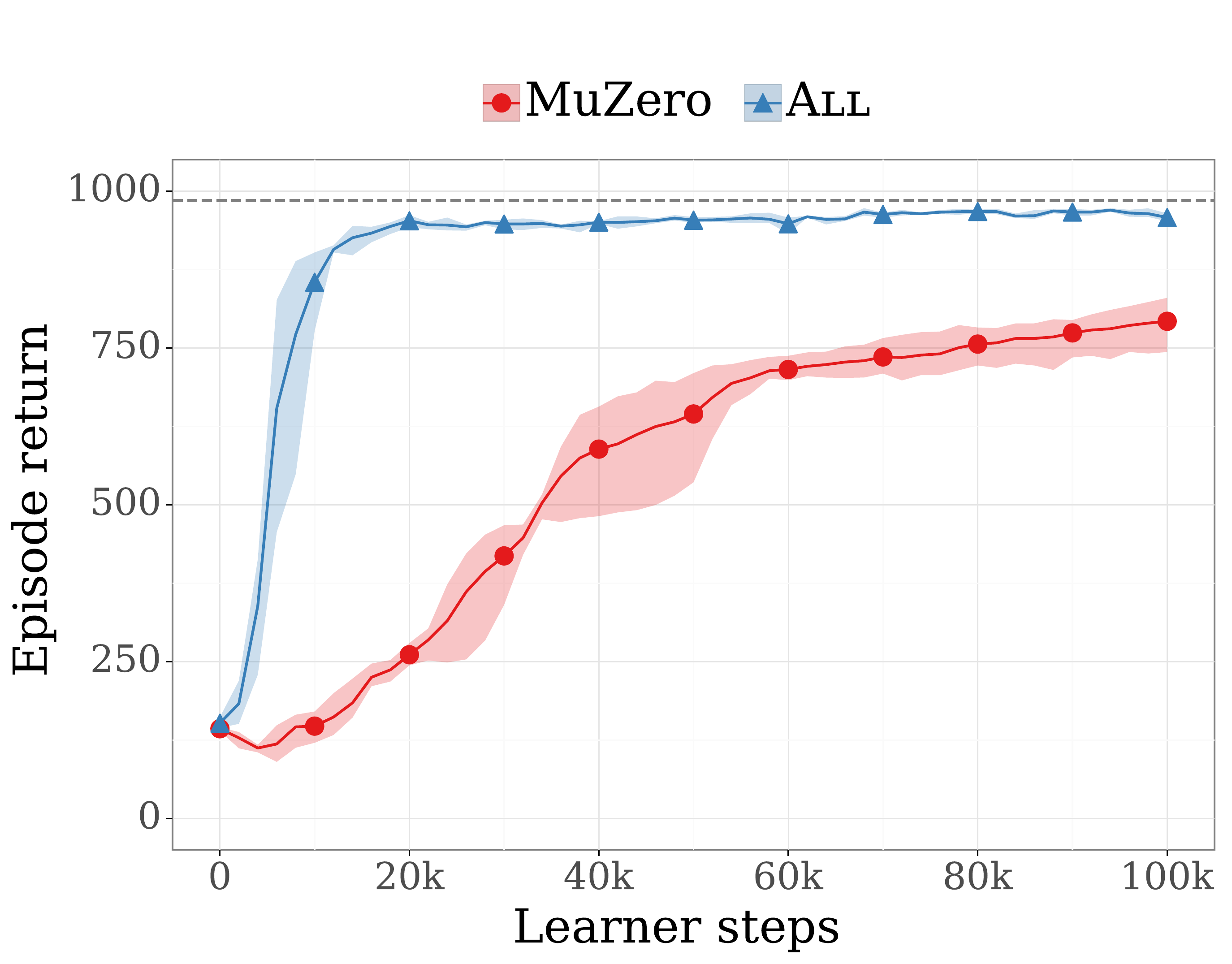}\caption{24 simulations}
\end{subfigure}
\begin{subfigure}{0.32\columnwidth}
\includegraphics[width=\linewidth]{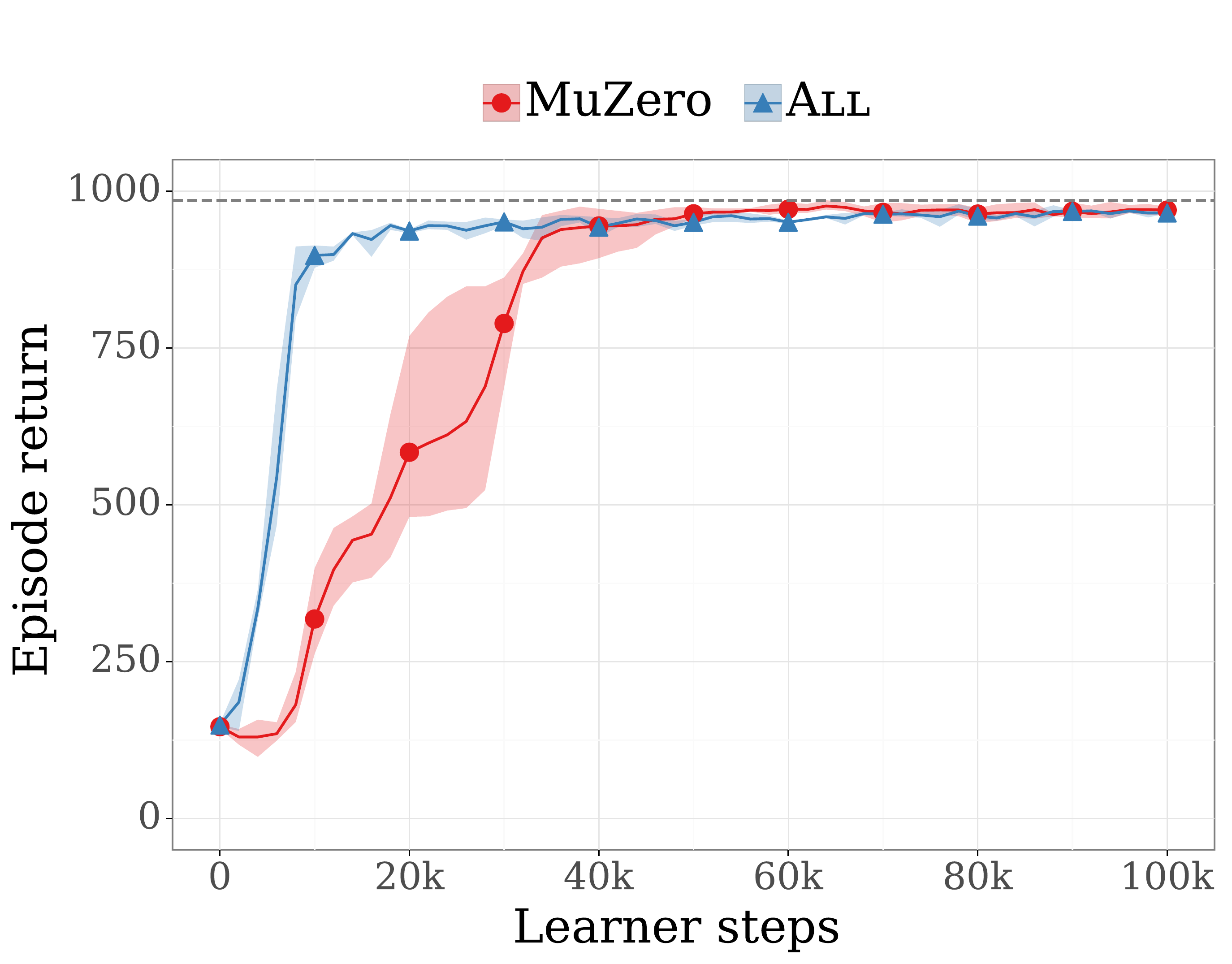}\caption{50 simulations}
\end{subfigure}
\caption{\label{fig:mz_vs_pz_cc_ws}Comparison of MuZero and \algoall{} on Walker Stand.}
\end{figure}

\begin{figure}
\begin{subfigure}{0.32\columnwidth}
\includegraphics[width=\linewidth]{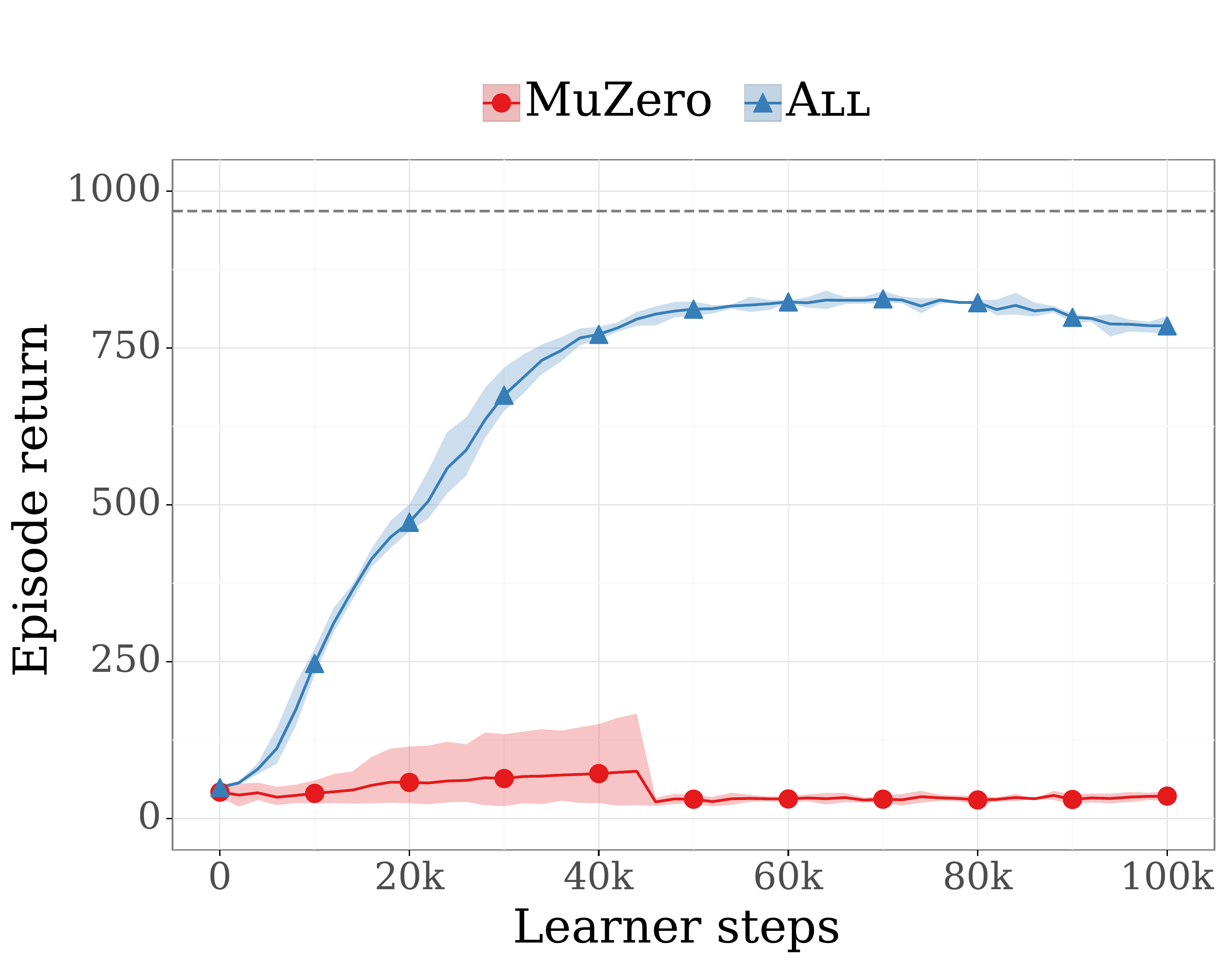}\caption{4 simulations}
\end{subfigure}
\begin{subfigure}{0.32\columnwidth}
\includegraphics[width=\linewidth]{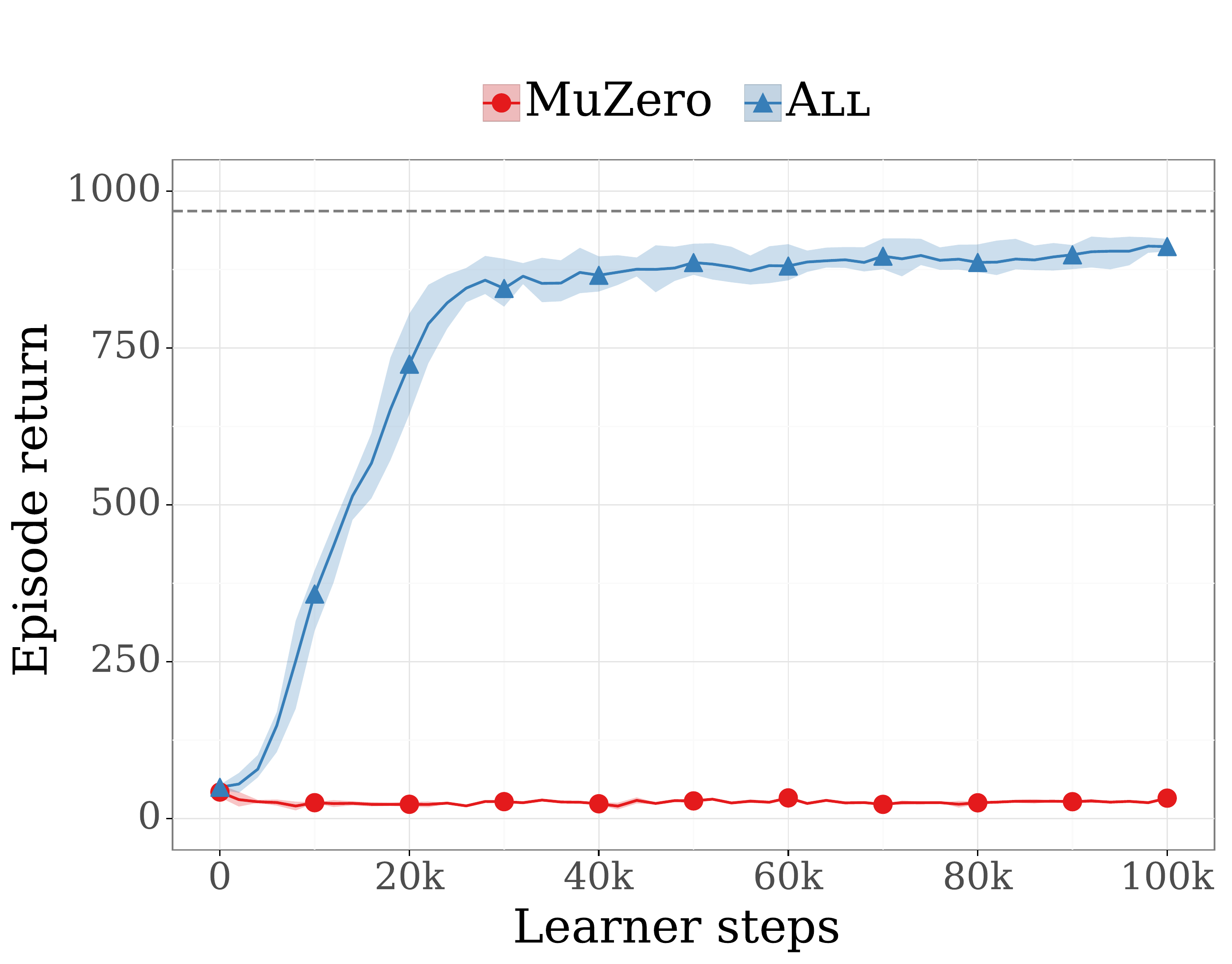}\caption{6 simulations}
\end{subfigure}
\begin{subfigure}{0.32\columnwidth}
\includegraphics[width=\linewidth]{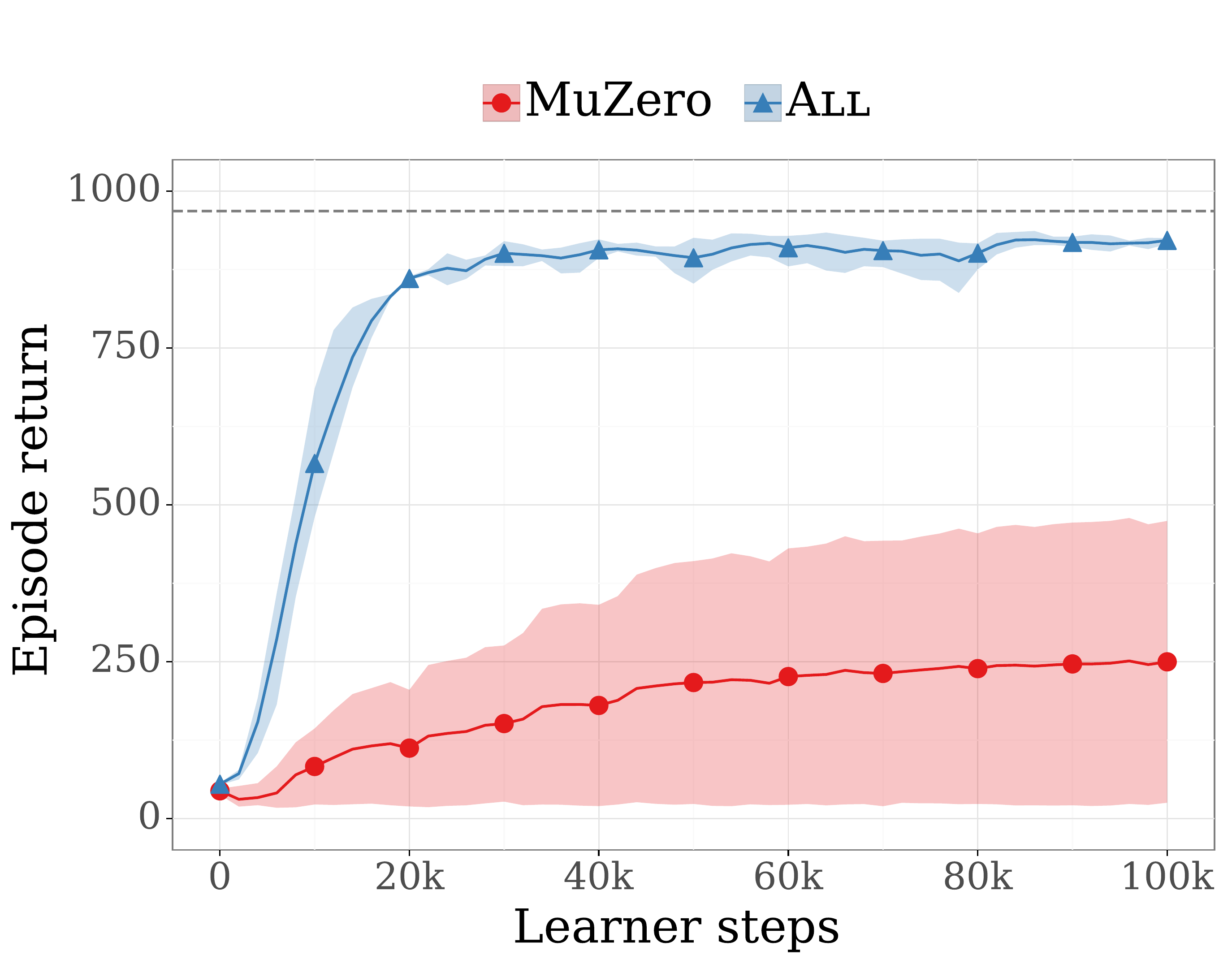}\caption{8 simulations}
\end{subfigure}

\begin{subfigure}{0.32\columnwidth}
\includegraphics[width=\linewidth]{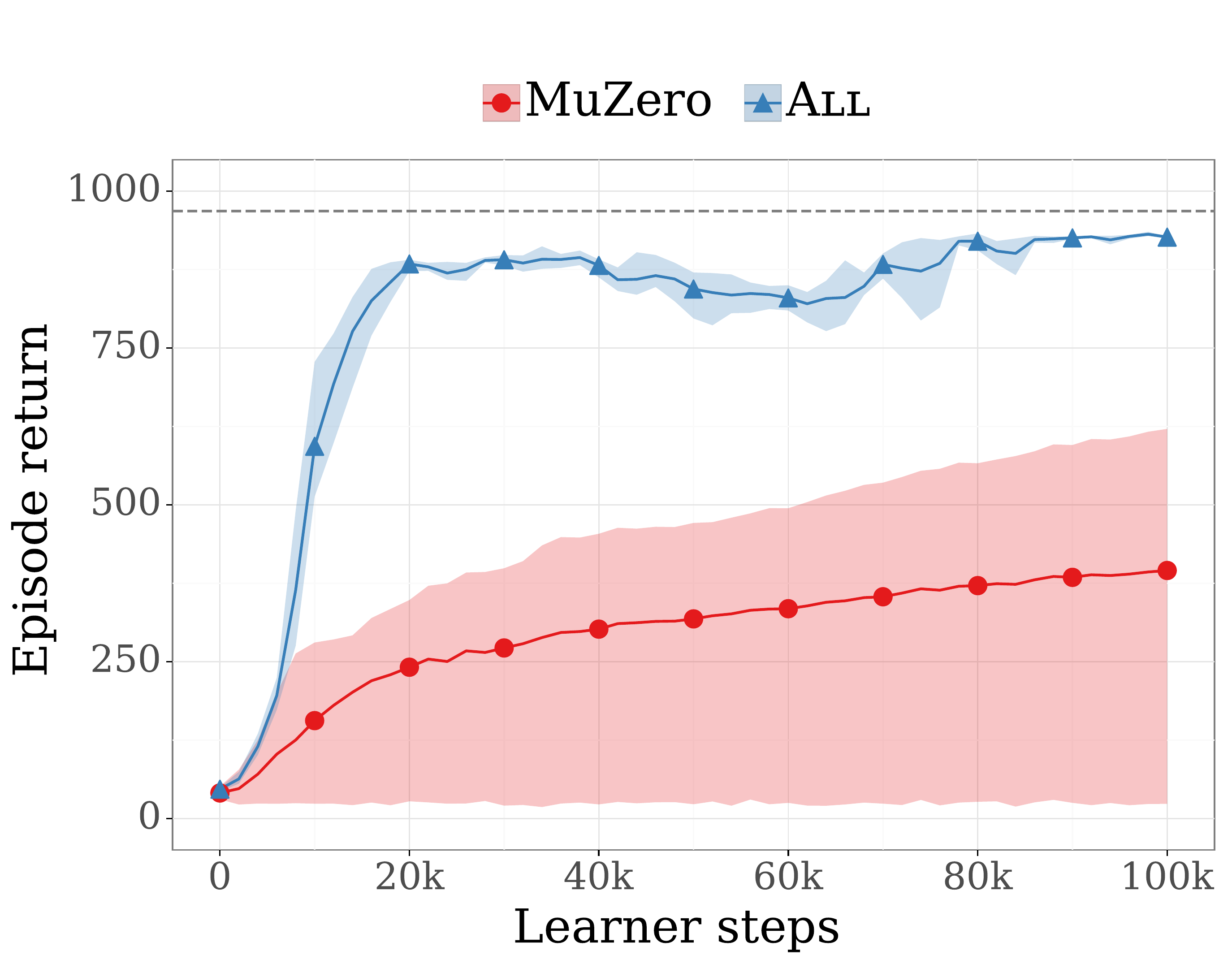}\caption{12 simulations}
\end{subfigure}
\begin{subfigure}{0.32\columnwidth}
\includegraphics[width=\linewidth]{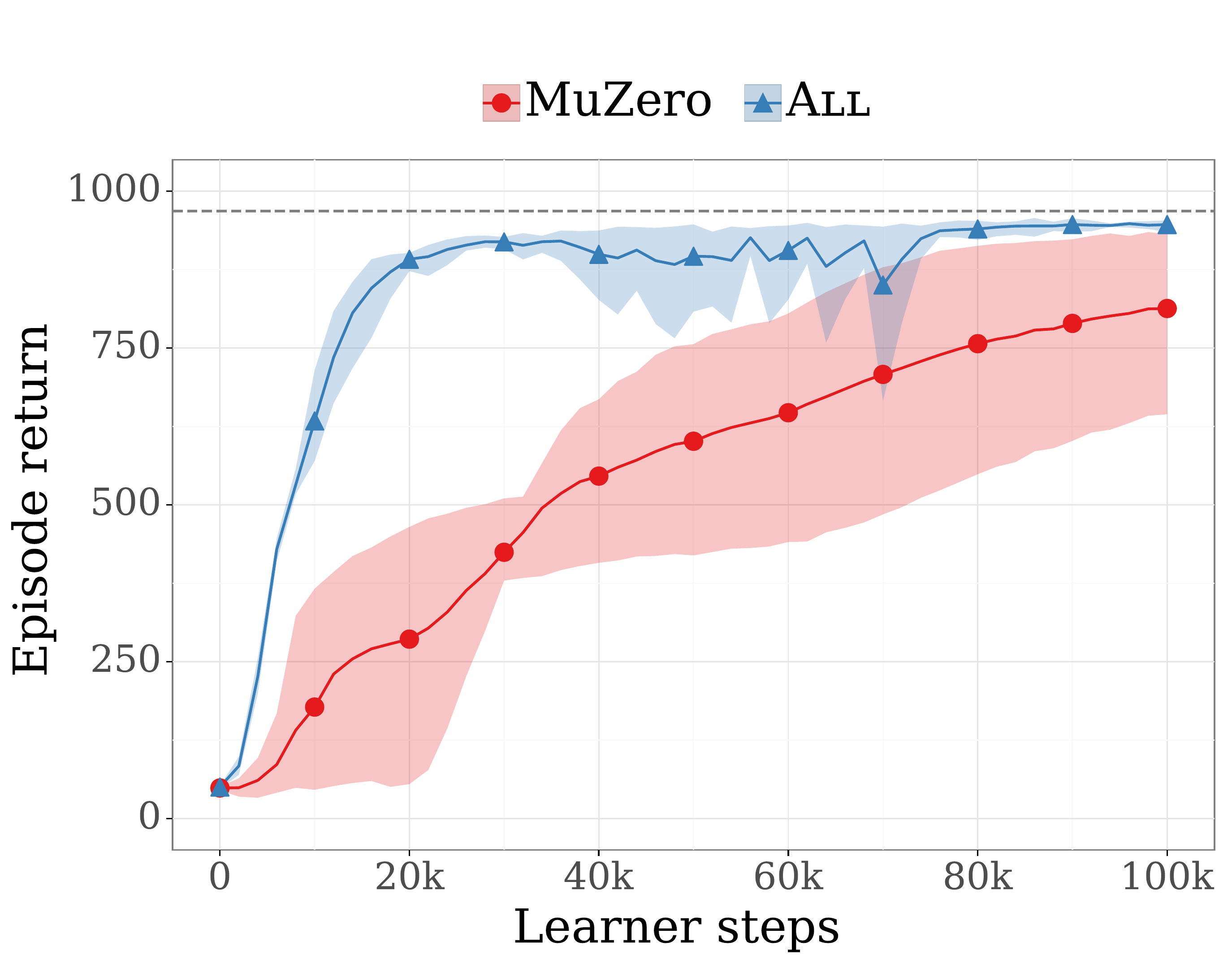}\caption{24 simulations}
\end{subfigure}
\begin{subfigure}{0.32\columnwidth}
\includegraphics[width=\linewidth]{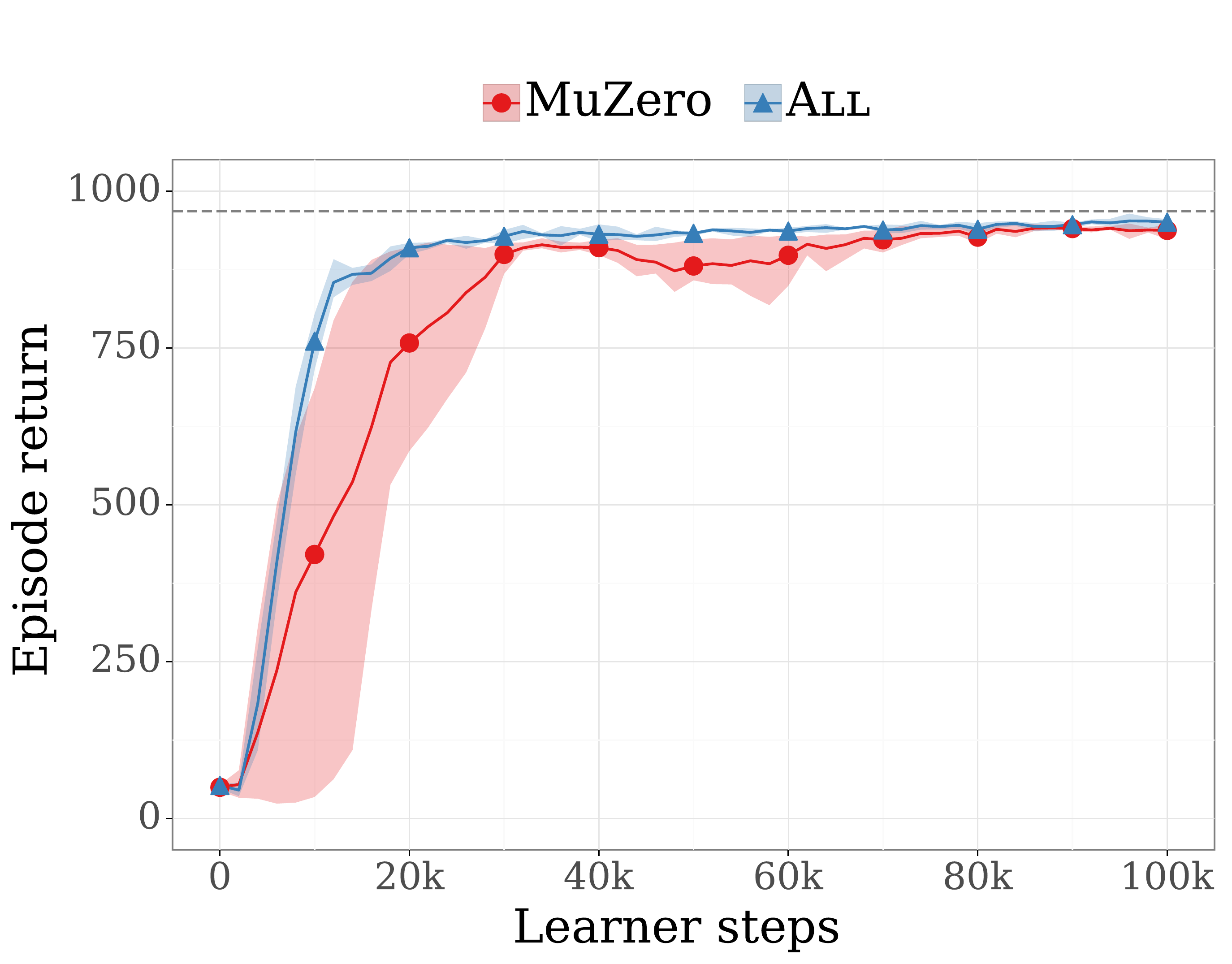}\caption{50 simulations}
\end{subfigure}
\caption{\label{fig:mz_vs_pz_cc_ww}Comparison of MuZero and \algoall{} on Walker Walk.}
\end{figure}

\begin{figure}
\begin{subfigure}{0.32\columnwidth}
\includegraphics[width=\linewidth]{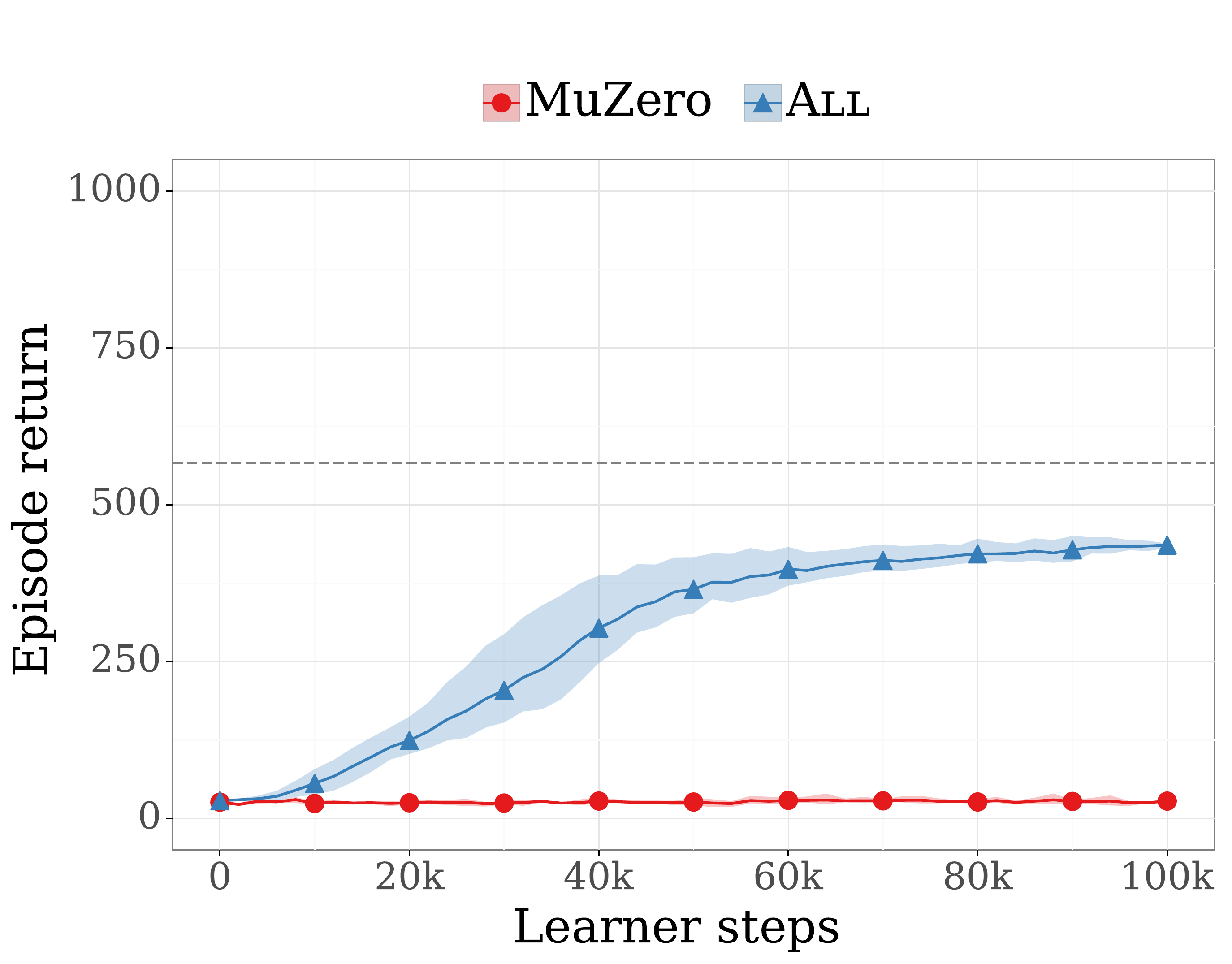}\caption{4 simulations}
\end{subfigure}
\begin{subfigure}{0.32\columnwidth}
\includegraphics[width=\linewidth]{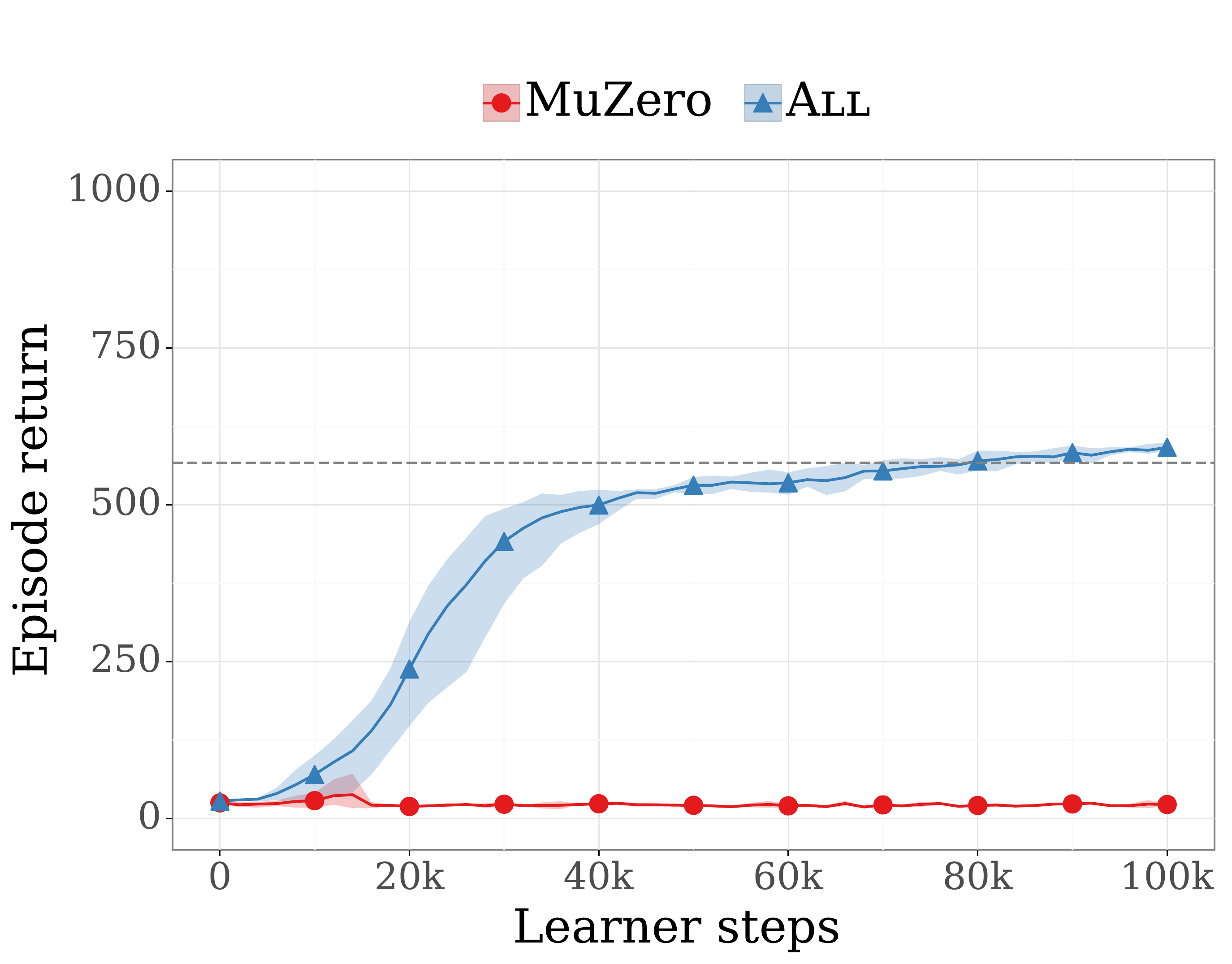}\caption{6 simulations}
\end{subfigure}
\begin{subfigure}{0.32\columnwidth}
\includegraphics[width=\linewidth]{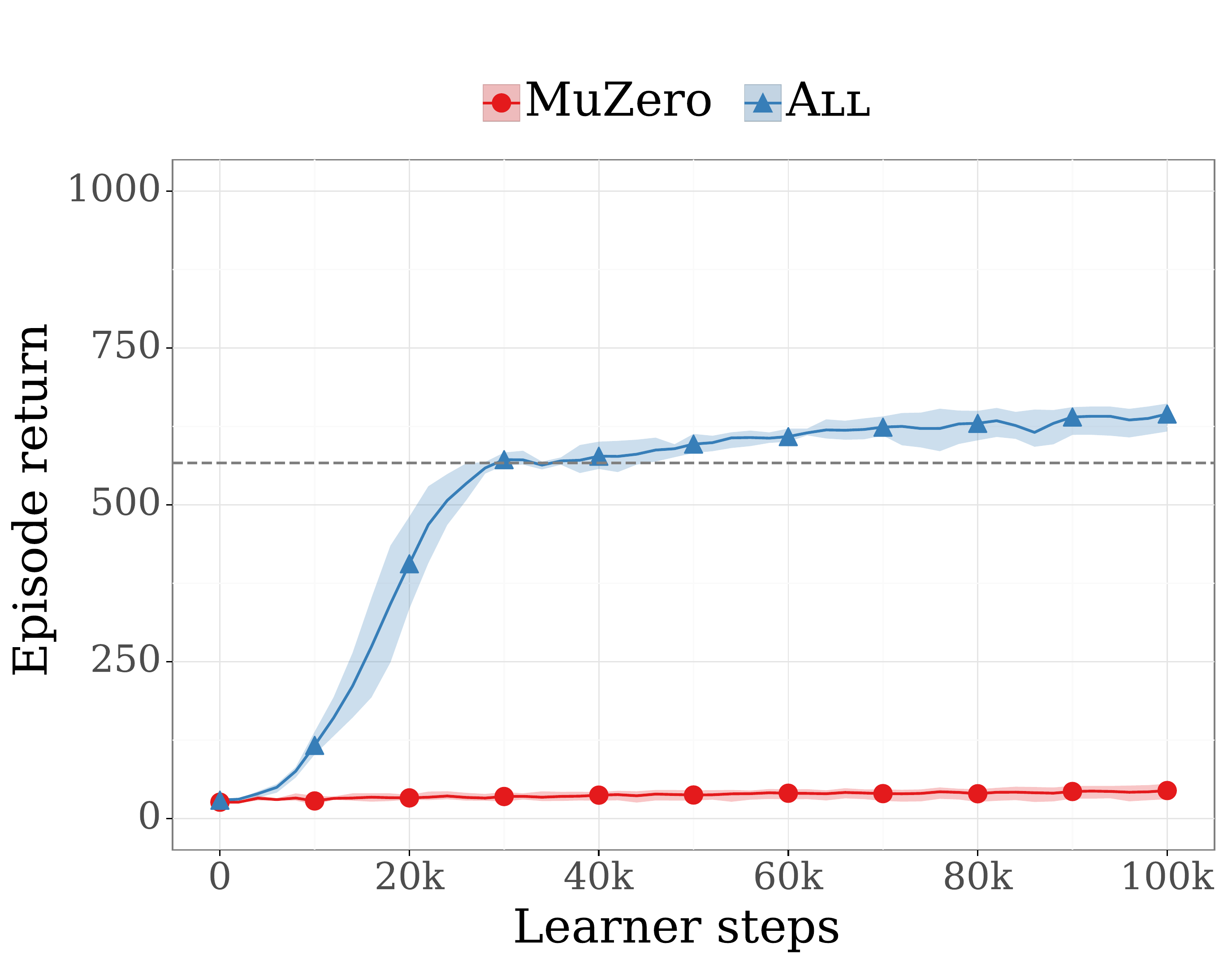}\caption{8 simulations}
\end{subfigure}

\begin{subfigure}{0.32\columnwidth}
\includegraphics[width=\linewidth]{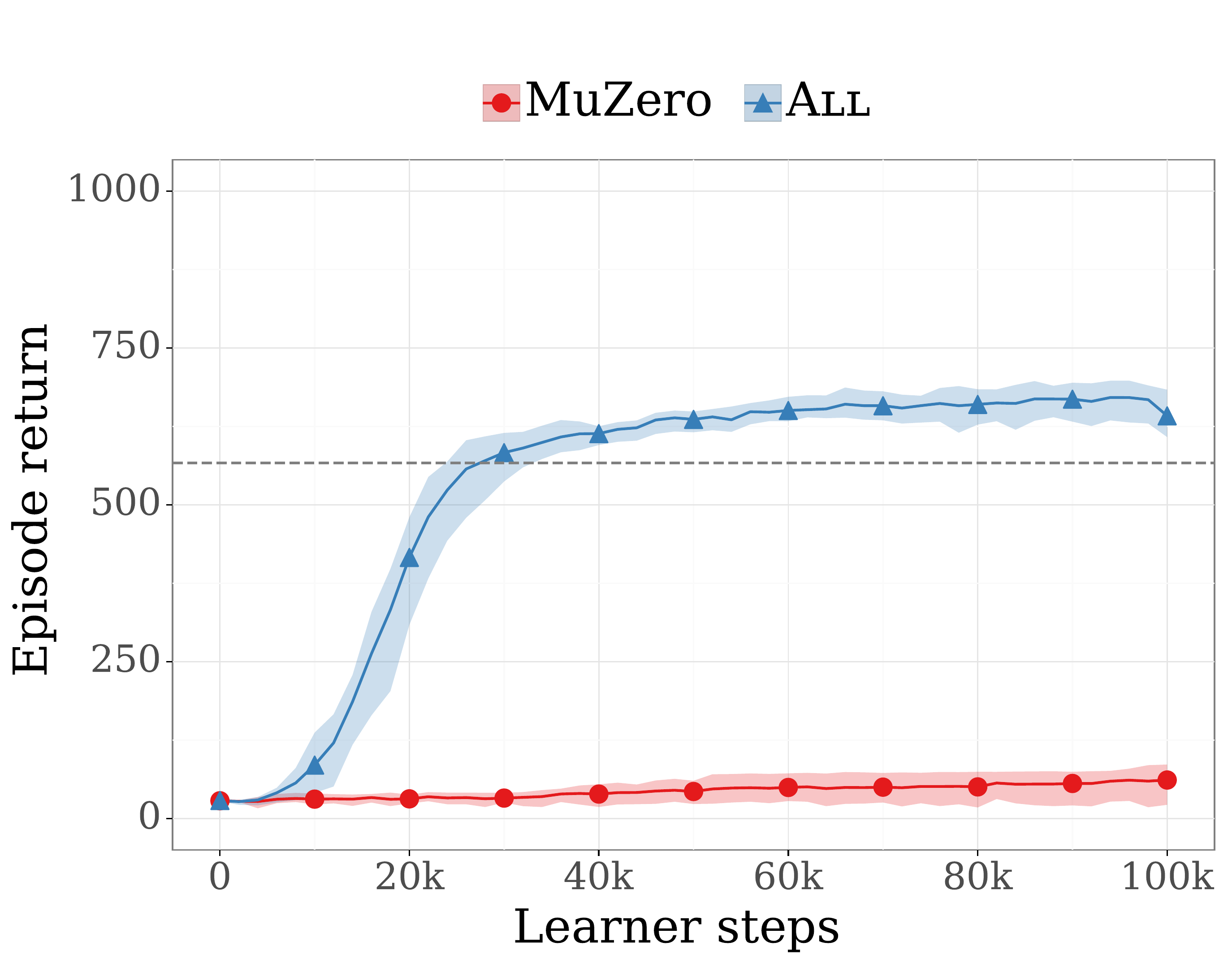}\caption{12 simulations}
\end{subfigure}
\begin{subfigure}{0.32\columnwidth}
\includegraphics[width=\linewidth]{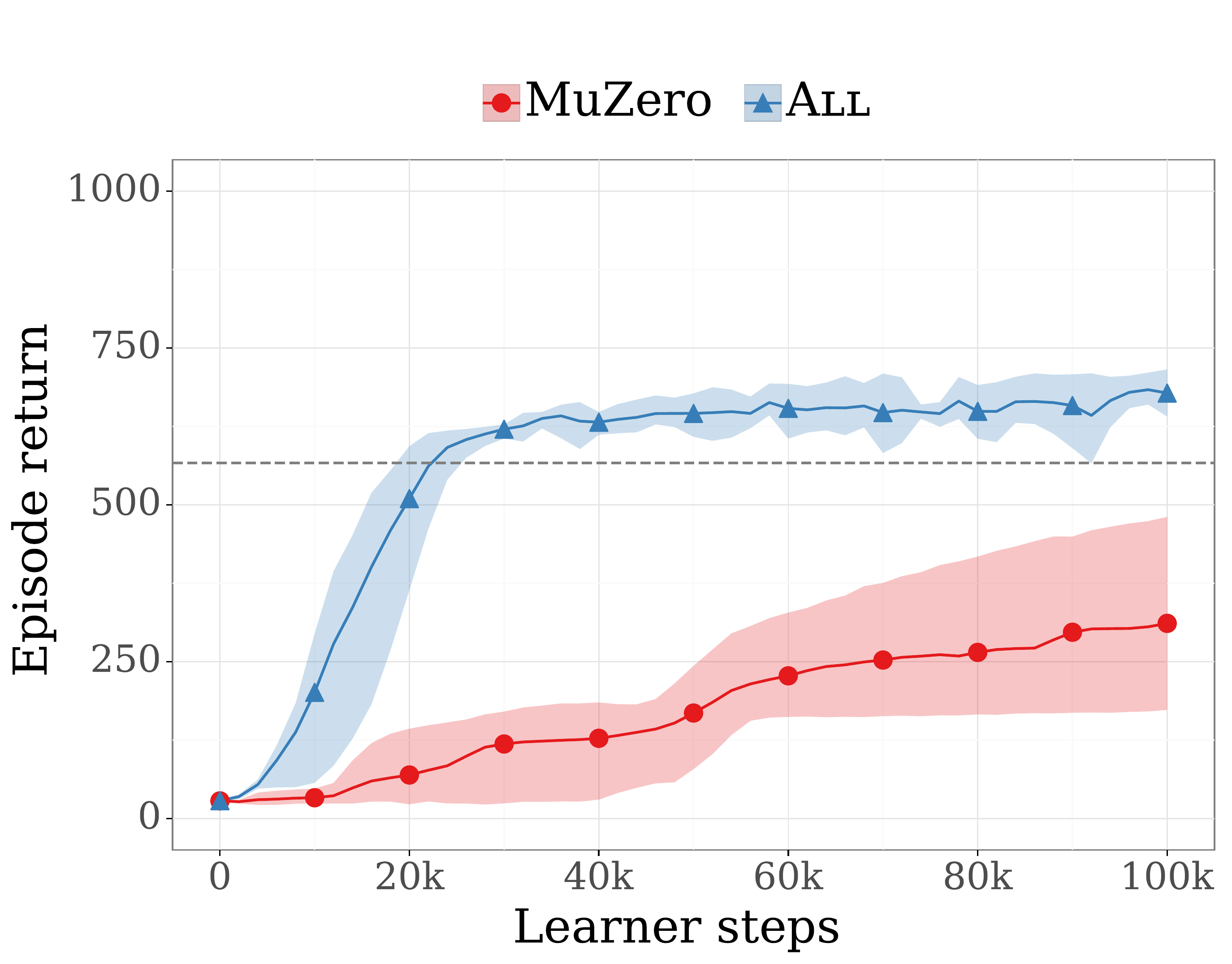}\caption{24 simulations}
\end{subfigure}
\begin{subfigure}{0.32\columnwidth}
\includegraphics[width=\linewidth]{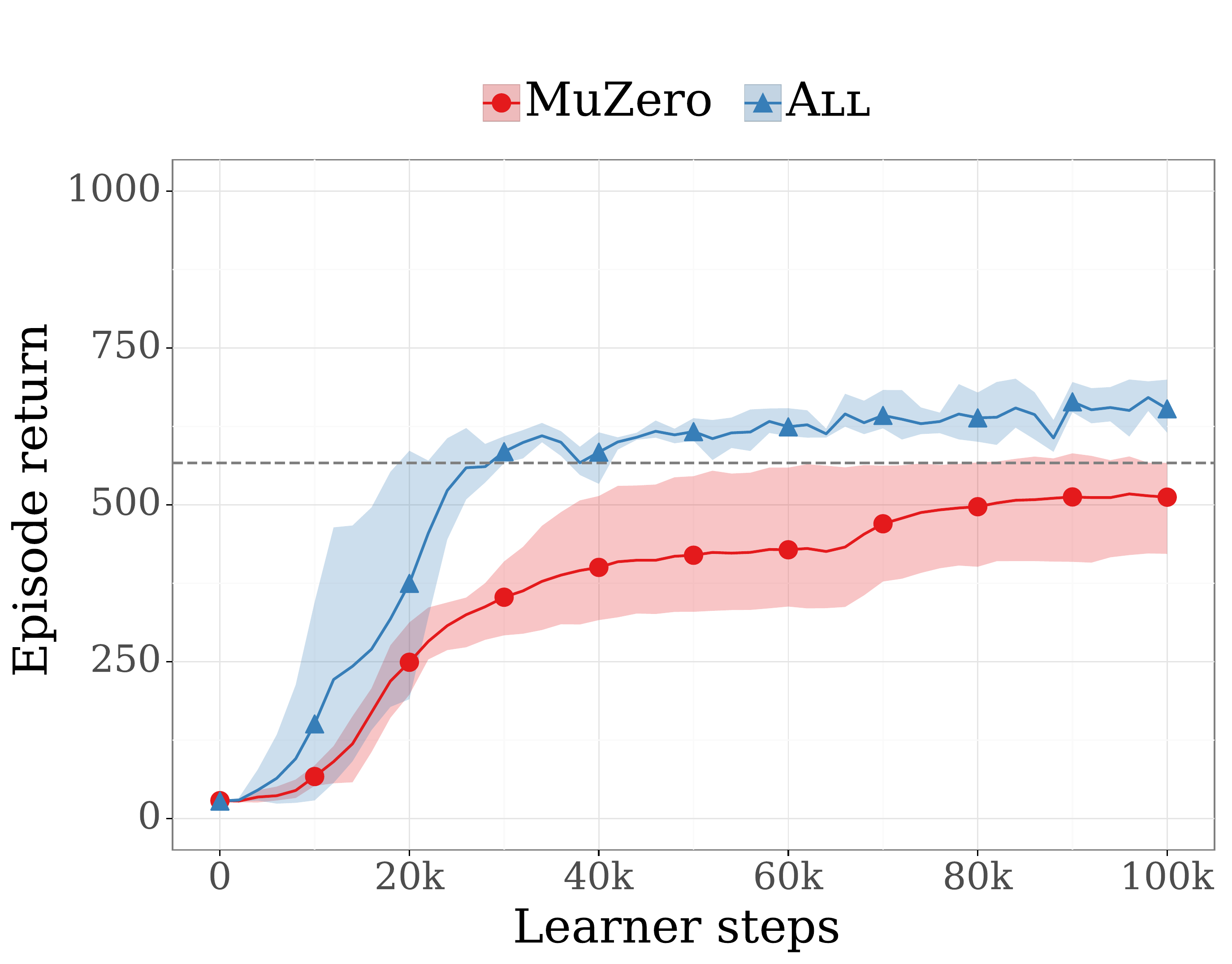}\caption{50 simulations}
\end{subfigure}
\caption{\label{fig:mz_vs_pz_cc_wr}Comparison of MuZero and \algoall{} on Walker Run.}
\end{figure}

\Cref{fig:mz_vs_pz_cc_cheetah,fig:mz_vs_pz_cc_ws,fig:mz_vs_pz_cc_ww,fig:mz_vs_pz_cc_wr} present a comparison of MuZero and \algoall{} on a subset of 4 of the medium-difficulty~\citep{van2020q} DeepMind Control Suite~\citep{tassa2018control} tasks chosen for their relatively high-dimensional action space among these medium-difficulty problems ($n_\text{dim} = 6$). \Cref{fig:mz_vs_pz_with_nsim} compare the score of MuZero and \algoall{} after 100k learner steps on these four medium difficulty Control problems. These continuous control problems are cast to a discrete action formulation using the method presented in \Cref{app:continuous_action_discretization}; note that these experiments only use pixel renderings and not the underlying scalar states.

These curves present the median (solid line) and min-max interval (shaded area) computed over 3 seeds in the same settings as described in \Cref{app:mz_vs_pz_ms_pacman}. The hyper-parameters are the same as in the other experiments; no specific tuning was performed for the continuous control domain. The horizontal dashed line corresponds to the performance of the D4PG algorithm when trained on pixel observations only~\citep{barth-maron2018distributional}, as reported by~\citep{tassa2018control}.

\subsection{Complemantary experiments on comparison with PPO}
Since we interpret the MCTS-based algorithms as regularized policy optimization algorithms, as a sanity check for the proposal's performance gains, we compare it with state-of-the-art proximal policy optimization (PPO) ~\citep{schulman2017proximal}. Since PPO is a near on-policy optimization algorithm, whose gradient updates are purely based on on-policy data, we adopt a lighter network architecture to ensure its stability. Please refer to the public code base ~\citep{dhariwal2017openai} for a review of the neural network architecture and algorithmic details.

To assess the performance of PPO, we train with both state-based inputs and image-based inputs. State-based inputs are low-dimensional sensor data of the environment, which renders the input sequence strongly Markovian ~\citep{tassa2018control}. For image-based training, we adopt the same inputs as in the main paper. The performance is reported in \Cref{table:ppo-comparison} where each score is the evaluation performance of PPO after the convergence takes place. We observe that state-based PPO performs significantly better than image-based PPO, while in some cases it matches the performance of \algoall. In general, image-based PPO significantly underperforms \algoall.

\begin{table}[h]
    \vskip 0.1in
    \begin{center}
    \footnotesize
    \begin{sc}
    \begin{tabular}{l r r r r}\toprule[1.5pt]
        \bf Benchmarks & \bf PPO (state) & \bf PPO (image) & \bf MuZero (image) & \bf \algoall (image) \\\midrule
Walker-Walk  & 406 & 270 & 925 & 941 \\ 
Walker-Stand & 937 & 357 & 959 & 951 \\ 
Walker-Run   & 340 & 71  & 533 & 644 \\ 
Cheetah-Run  & 538 & 285 & 887 & 882 \\
         \bottomrule[1.46pt]
    \end{tabular} \par
    \end{sc}
    \caption{Comparison to the performance of PPO baselines on benchmark tasks. The inputs to PPO are either state-based or image-based. The performance is computed as the evaluated returns after the training is completed, averaged across 3 random seeds.}
    \label{table:ppo-comparison}
    \end{center}
\end{table}

\clearpage
\section{Derivations for \Cref{sec:mcts_as_po}}

\subsection{Proof of \Cref{prop:action_selection_diff}, \Cref{eq:action_selection_pibar} and \Cref{prop:pihat_leq_pibar}.}
\label{sec:proof_action_selection_pibar}

We start with a definition of the $f$-divergence \cite{csiszar1964informationstheoretische}. 
\begin{definition}{$f$-divergence}
For any probability distributions $p$ and $q$ on $\mathcal{A}$ and function $f:\Real\rightarrow\Real$ such that $f$ is a convex function on $\mathbb{R}$ and $f(1) = 0$, the $f$-divergence $D_f$ between $p$ and $q$ is defined as
\begin{align}
    D_f(p,q) = \sum_{b\in\mathcal{A}} q(b) f\prth{\frac{p(b)}{q(b)}}
\end{align}
\end{definition}

\begin{remark}{}
Let $D_f$ be a $f$-divergence, 

\hspace{2em}(1) $\forall x,y\; D(x,y) \ge 0$
\hspace{2em}(2) $D(x,y) = 0 \iff x=y$
\hspace{2em}(3) $D(x,y)$ is jointly convex in $x$ and $y$.
\end{remark}

We states four lemmas that we formally prove in \Cref{sec:proofs_lemma}. 
\begin{lemma}{}
\label{lemma:general_f_divergence}
\begin{align}
    \nabla_{\pi} \prth{\mathbf{q}^T \pi - \lambda \cdot D_f\brkt{\pi, \mathbf{\pi_\theta}}} 
    = \mathbf{q} - \lambda\cdot f'\prth{\frac{\pi}{\pi_\theta}}
\end{align}
\end{lemma}

Where $\pi_\theta$ is assumed to be non zero. We now restate the definition of $\pihat[a] \eqdef \frac{n_a + 1}{N + |\mathcal{A}|}$. 
\begin{lemma}{}
\label{lemma:part_f_divergence}
\begin{align}
    \argmax_a \brkt{\frac{\partial}{\partial n_a} \prth{\mathbf{q}^T \pi - \lambda \cdot D_f\brkt{\pihat, \mathbf{\pi_\theta}}}}
    = \argmax_a \brkt{\mathbf{q_a} - \lambda\cdot f'\prth{\frac{\pihat(a)}{\pi_\theta(a)}}}
\end{align}
\end{lemma}


Now we consider a more general definition of $\pibar$ using any $f$-divergence for some $\lambda_{f} > 0$ and assume $\pi_\theta > 0$, 
\begin{align}
\label{eq:general_pibar}
\pibar_f \eqdef \argmin_{\mathbf{y} \in \mathcal{S}}  \mathbf{q}^T \mathbf{y} - \lambda_{f} D_f\prth{\pi_\theta, \mathbf{y}}.
\end{align}
We also consider the following action selection formula based on $f$-divergence $D_f$. 
\begin{align}
    a^\ast_f \eqdef \argmax_a \brkt{q_a - \lambda_{f}\cdot f'\prth{\frac{\pihat}{\pi_\theta}}},\quad\text{with}.
\end{align}

\begin{lemma}
\label{lemma:argmax}
\begin{align}
    \argmax_a \brkt{q_a - \lambda_{f}\cdot f'\prth{\frac{\pihat(a)}{\pi_\theta(a)}}} = \argmax_a \brkt{f'\prth{\frac{\pibar_f(a)}{\pi_\theta(a)}} - f'\prth{\frac{\pihat(a)}{\pi_\theta(a)}}}.
\end{align}
\end{lemma}

\begin{lemma}
\label{lemma:pihat_lower_pibar}
\begin{align}
    \pihat(a^\ast_f) \le \pibar(a^\ast_f).
\end{align}
\end{lemma}

Applying \Cref{lemma:part_f_divergence,lemma:pihat_lower_pibar} with the appropriate function $f$ directly leads to \Cref{prop:action_selection_diff}, \Cref{prop:pihat_leq_pibar}, and
\Cref{prop:uct}. In particular, we use
\begin{align}
    &\text{For AlphaZero:} \hspace{-7em} & f(x) = -\log(x) \\
    &\text{For UCT:} & f(x) = 2 - 2\sqrt{x}
\end{align}




\begin{tabular}{l l l l l}\toprule
    Algorithm & Function $f(x)$ & Derivative $f'(x)$ & Associated $f$-divergence & Associated action selection formula \\
     \midrule
    --- & $x\cdot\log\prth{x} $ & $ \log\prth{x} + 1 $ & $ D_f(p,q) = \text{KL}(p,q) $ & $ \argmax_a  q_a + \frac{c}{\sqrt{N}}\cdot\log\prth{\frac{\pi_\theta(a)}{n_a+1}} $ \\
    UCT & $2 - 2\sqrt{x} $ & $ -\frac{1}{\sqrt{x}} $ & $ D_f(p,q) = 2 - 2\sum_{b\in\mathcal{A}} \sqrt{p_a\cdot q_a} $ & $ \argmax_a  q_a + c \cdot \sqrt{\frac{\pi_\theta}{n_a+1}}$ \\
    AlphaZero & $-\log\prth{x} $ & $ -\frac{1}{x} $ & $ D_f(p,q) = \text{KL}(q,p) $ & $ \argmax_a q_a + c \cdot\pi_\theta\cdot \frac{\sqrt{N}}{n_a+1}$ \\
    \bottomrule
\end{tabular}

\subsection{Proofs of \Cref{lemma:general_f_divergence,lemma:part_f_divergence,lemma:argmax,lemma:pihat_lower_pibar}}
\paragraph{Proof of \Cref{lemma:general_f_divergence}}
\begin{proof}
For any action $a\in\mathcal{A}$ using basic differentiation rules we have
\begin{align}
    \prth{\nabla_{\pi} \prth{\mathbf{q}^T \pi - \lambda_f \cdot D_f\brkt{\pi, \mathbf{\pi_\theta}}}}[a]
    &= \frac{\partial}{\partial \pi_a} \brkt{\,\sum_{b\in\mathcal{A}} \prth{q_b\cdot\pi_b} - \lambda_f \sum_{b\in\mathcal{A}} \pi_\theta(b) f\prth{\frac{\pi_b}{\pi_\theta(b)}}} \\
    &= \frac{\partial}{\partial \pi_a} \brkt{\,\sum_{b\in\mathcal{A}} \prth{q_b\cdot\pi_b}} - \lambda_f\cdot\frac{\partial}{\partial \pi_a} \sum_{b\in\mathcal{A}} \pi_\theta(b) f\prth{\frac{\pi_b}{\pi_\theta(b)}} \\
    &= q_a - \lambda_f\cdot\pi_\theta(a)\cdot\frac{\partial}{\partial \pi_a}f\prth{\frac{\pi_a}{\pi_\theta(a)}} \\
    &= q_a - \lambda_f\cdot f'\prth{\frac{\pi_a}{\pi_\theta(a)}}
    = \prth{\mathbf{q} - \lambda_f\cdot f'\prth{\frac{\pi}{\pi_\theta}}}[a]
\end{align}
\end{proof}

\paragraph{Proof of \Cref{lemma:part_f_divergence}}
\label{sec:proofs_lemma}
\begin{proof}
\begin{align}
    \frac{\partial}{\partial n_a} \prth{\mathbf{q}^T \pi - \lambda_f \cdot D_f\brkt{\pihat, \mathbf{\pi_\theta}}} 
    &= \frac{\partial}{\partial n_a} \brkt{\,\sum_{b\in\mathcal{A}} \prth{q_b\cdot\frac{n_b + 1}{|\mathcal{A}| + \sum_{c\in\mathcal{A}} n_c}} - \lambda_f \sum_{b\in\mathcal{A}} \pi_\theta(b) \cdot f\prth{\frac{n_b + 1}{\prth{|\mathcal{A}| + \sum_{c\in\mathcal{A}} n_c}\cdot \pi_\theta(b)}}} \\
    &=\beta + \frac{q_a}{|\mathcal{A}| + \sum_{c\in\mathcal{A}} n_c} - \frac{\lambda_f}{|\mathcal{A}| + \sum_{c\in\mathcal{A}} n_c} f'\prth{\frac{n_a + 1}{\prth{|\mathcal{A}| + \sum_{c\in\mathcal{A}} n_c}\cdot \pi_\theta(b)}}\\
    &=\beta + \frac{1}{|\mathcal{A}| + \sum_{c\in\mathcal{A}} n_c}\prth{ q_a - \lambda_f f'\prth{\frac{\pihat(a)}{\pi_\theta(b)}}},
\end{align}
where $\beta = -\cfrac{\sum_{b\in\mathcal{A}} \brkt{q_b - \lambda_f f'\prth{\frac{\pihat(b)}{\pi_\theta(b)}}}}{\prth{|\mathcal{A}| + \sum_{c\in\mathcal{A}} n_c}^2}$ is independent of $a$. Also because $\frac{1}{|\mathcal{A}| + \sum_{c\in\mathcal{A}} n_c} > 0$ then 
\begin{align}
    \argmax_a \frac{\partial}{\partial n_a} \prth{\mathbf{q}^T \pi - \lambda_f \cdot D_f\brkt{\pihat, \mathbf{\pi_\theta}}} 
    = \argmax_a \brkt{\mathbf{q}_a - \lambda_f \cdot f'\prth{\frac{\pihat(a)}{\pi_\theta(a)}}} \\
\end{align}
\end{proof}


\paragraph{Proof of \Cref{lemma:argmax}}
\begin{proof}
The \Cref{eq:general_pibar} is a differentiable strictly convex optimization problem, its unique solution satisfies the KKT condition requires (see Section~5.5.3 of \citealp{boyd2004convex}) therefore there exists $\alpha\in\mathbb{R}$ such that for all actions $a,$
\begin{align}
\nabla_{\pibar} \prth{\mathbf{q}^T \pibar - \lambda_f D_f\prth{\mathbf{\pi_\theta}, \mathbf{\pibar}}} = \alpha \mathbbm{1}
\end{align}
where $\mathbbm{1}$ is the vector constant equal one: $\forall a\,\mathbbm{1}_a = 1$. Using \Cref{lemma:general_f_divergence} setting $\pi$ to $\pibar$ we get
\begin{align}
\exists \alpha \quad q - \lambda_f \cdot f'\prth{\frac{\pibar}{\pi_\theta}} = \alpha \mathbbm{1}.
\label{eq:kkt}
\end{align}

\begin{align}
    q - \lambda_{f}\cdot f'\prth{\frac{\pihat}{\pi_\theta}} &= q_a - \lambda_{f}\cdot \prth{f'\prth{\frac{\pihat}{\pi_\theta}} + f'\prth{\frac{\pibar}{\pi_\theta}} - f'\prth{\frac{\pibar}{\pi_\theta}}} \\
    & = \alpha\mathbbm{1} + \lambda_{f}\cdot\prth{f'\prth{\frac{\pibar}{\pi_\theta}} - f'\prth{\frac{\pihat}{\pi_\theta}}}\\
    \argmax\brkt{q - \lambda_{f}\cdot f'\prth{\frac{\pihat}{\pi_\theta}}} &= \argmax\brkt{\alpha\mathbbm{1} + \lambda_{f}\cdot\prth{f'\prth{\frac{\pibar}{\pi_\theta}} - f'\prth{\frac{\pihat}{\pi_\theta}}}}\\
    &= \argmax\brkt{f'\prth{\frac{\pibar}{\pi_\theta}} - f'\prth{\frac{\pihat}{\pi_\theta}}} \hspace{2em}\text{(because $\lambda_f > 0$)}\\
\end{align}

\end{proof}

\paragraph{Proof of \Cref{lemma:pihat_lower_pibar}}
\begin{proof}
Since $\sum_a \pihat(a|x) = \sum_a \pibar(a|x) = 1$, there exists at least an action $a_0$ for which $0 \leq \pihat(a_0|x) \leq \pibar(a_0|x)$ then $0 \leq \frac{\pihat(a_0|x)}{\pi_\theta(a|x)} \leq \frac{\pibar(a_0|x)}{\pi_\theta(a|x)}$ as $\pi_\theta(a|x) > 0$. Because $f$ is convex then $f'$ is increasing and therefore 
\begin{align}
\label{eq:lemma4_1}
    f'\prth{\frac{\pibar(a_0|x)}{\pi_\theta(a_0|x)}} - f'\prth{\frac{\pihat(a_0|x)}{\pi_\theta(a_0|x)}} \ge 0.
\end{align}
Now using \Cref{lemma:argmax}
\begin{align}
\label{eq:lemma4_2}
    f'\prth{\frac{\pibar(a^\ast_f|x)}{\pi_\theta(a^\ast_f|x)}} - f'\prth{\frac{\pihat(a^\ast_f|x)}{\pi_\theta(a^\ast_f|x)}} \ge f'\prth{\frac{\pibar(a_0|x)}{\pi_\theta(a_0|x)}} - f'\prth{\frac{\pihat(a_0|x)}{\pi_\theta(a_0|x)}}
\end{align}
We put \Cref{eq:lemma4_1,eq:lemma4_2} together 
\begin{align}
    f'\prth{\frac{\pibar(a^\ast_f|x)}{\pi_\theta(a^\ast_f|x)}} - f'\prth{\frac{\pihat(a^\ast_f|x)}{\pi_\theta(a^\ast_f|x)}} \ge 0 
\end{align}
Finally we use again that $f'$ is increasing and $\pi_\theta > 0$ to conclude the proof
\begin{align}
    \pihat(a^\ast_f|x) \le \pibar(a^\ast_f|x) 
\end{align}
\end{proof}

\subsection{Tracking property in the constant $\pibar$ case} 

\label{sec:one_over_N_bound}
Let $\pi$ be some target distribution independent of the round $t \ge 0$. At each round $t,$ starting from $t=1,$ an action $a_t\in\mathcal{A}$ is selected and for any $t\ge 0$, we define
\begin{align*}
    p_t(a) \triangleq \frac{n_t(a_t) + 1}{|\mathcal{A}| + \sum_b n_t(b)}\CommaBin
\end{align*}
where for any action $a\in\mathcal{A}$, $n_t(a)$ is the number of rounds the action $a$ has been selected,
\begin{align*}
    \forall a\in\mathcal{A}\quad n_t(a) \triangleq \sum_{i\le t} \delta\prth{a_t = a} \quad\text{ and } \delta\prth{a_t = a} \triangleq 1 \;\text { if and only if }\; a_t = a.
\end{align*}
\begin{proposition}{}
Assume that for all rounds  $t\ge 1,$ and for the chosen action $a_t\in\mathcal{A}$  we have
\begin{align}
    p_t(a_t) \le \pi(a_t).
    \label{eq:ass_ppi}
\end{align}
Then, we have that
\begin{align*}
     \forall a\in\mathcal{A}, t\ge 1\quad \left|\pi(a) - p_t(a)\right| \le \frac{|\mathcal{A}| - 1}{|\mathcal{A}| + t}\cdot
\end{align*}
\end{proposition}

Before proving the proposition above, note that $\mathcal{O}(1/t)$ is the best approximation w.r.t.\,$t,$ since for any integer $k\ge 0$, taking $\pi(a) = (\frac{1}{2} + k)/(|\mathcal{A}| + t),$ we have 
that for all $n\ge 0,$
\begin{align*}
     \left|\pi(a) - \frac{n+1}{|\mathcal{A}| + t}\right| \ge \frac{1}{2}\frac{1}{|\mathcal{A}| + t}\CommaBin 
\end{align*}
which follows from the fact that $\forall k,n\in\mathbb{N},$ $\left|\frac{1}{2} + k - (n+1)\right| = \left|k - n - \frac{1}{2}\right| \ge \frac{1}{2}\cdot$ 

\begin{proof}
By induction on the round $t,$ we prove that 
\begin{align}
    \forall t\ge1,a\in\mathcal{A}\quad p_t(a) \le \pi(a) + \frac{1}{|\mathcal{A}| + t}\cdot
    \label{eq:induction_ppi}
\end{align}
At round $t=1$, \Cref{eq:induction_ppi} holds as for any action $a$, $n_t(a) \ge 0$ therefore $p_t(a) \le 1$. Now, let us assume that \Cref{eq:induction_ppi} holds for some $t\ge 1$. We have that for all $a,$
\begin{align*}
    \quad\frac{1 + n_t(a)}{|\mathcal{A}| + \sum_b n_t(b)} \le \pi(a) + \frac{1}{|\mathcal{A}| + t}\cdot
\end{align*}

Note that at each round, there is exactly one action chosen and therefore, $\sum_b n_t(b) = t$. Furthermore, for $a'\neq a_{t+1},$ we have that $n_{t+1}(a') = n_t(a'),$ since $a'$ has not been chosen at round $t+1$. Therefore, for $a'\neq a_{t+1},$
\begin{align*}
    p_{t+1}(a') = \frac{n_{t+1}(a') + 1}{|\mathcal{A}| + t + 1} = \frac{n_{t}(a') + 1}{|\mathcal{A}| + t + 1} \le \frac{|\mathcal{A}| + t}{|\mathcal{A}| + t+1} p_{t}(a') \le \frac{|\mathcal{A}| + t}{|\mathcal{A}| + t+1} \left( \pi(a') + \frac{1}{|\mathcal{A}| +t }\right) \le \pi(a') + \frac{1}{|\mathcal{A}| + t + 1}\cdot
\end{align*}

Now, for the chosen action, $n_{t+1}(a_{t+1}) = n_t(a_{t+1}) + 1$. Using our assumption stated in \Cref{eq:ass_ppi}, we have that
\begin{align*}
     p_{t+1}(a_{t+1}) = \frac{n_{t+1}(a_{t+1})+1}{|\mathcal{A}| + t+1} = \frac{n_{t}(a_{t+1}) + 1+1}{|\mathcal{A}| + t+1} \le \frac{n_{t}(a_{t+1})+1}{|\mathcal{A}| + t+1} + \frac{1}{|\mathcal{A}| + t + 1} \le \pi(a_{t+1}) + \frac{1}{|\mathcal{A}|+t+1}\CommaBin
\end{align*}
which concludes the induction. Next, we compute a lower bound. For any action $a\in\mathcal{A}$ and round $t\ge 1,$
\begin{align*}
    p_t(a) = 1 - \sum_{b\neq a} p_t(b) \ge 1 - \sum_{b\neq a} \prth{\pi(b) + \frac{1}{|\mathcal{A}| + t}} = \prth{1 - \sum_{b\neq a} \pi(b)} - \sum_{b\neq a} \frac{1}{|\mathcal{A}| + t} = \pi(a) - \frac{|\mathcal{A}| - 1}{|\mathcal{A}| + t}\cdot
\end{align*}
We have for any action $a\in\mathcal{A},$ \[\pi(a) - \frac{|\mathcal{A}| - 1}{|\mathcal{A}| + t} \le p_t(a) \le \pi(a) + \frac{1}{|\mathcal{A}| + t}\cdot\] 
Since when $|\mathcal{A}| = 1,$ then by definition, $p_t(a) = \pi(a) = 1$ and for all rounds $t\ge 1,$ we get
\begin{align*}
    ||\pi - p_t||_{\infty} \le \frac{|\mathcal{A}| - 1}{|\mathcal{A}| + t}\cdot
\end{align*}

\end{proof}






\end{document}